\documentclass[10pt]{article} 
\usepackage[accepted]{tmlr}

\usepackage{amsmath,amsfonts,bm}

\def\eqref#1{equation~\ref{#1}}

\def\1{\bm{1}}

\DeclareMathAlphabet{\mathsfit}{\encodingdefault}{\sfdefault}{m}{sl}
\SetMathAlphabet{\mathsfit}{bold}{\encodingdefault}{\sfdefault}{bx}{n}

\usepackage{hyperref}
\usepackage{multirow}
\usepackage{graphicx}
\usepackage{xcolor}
\usepackage{subcaption}
\usepackage{url}
\usepackage{mathtools}
\usepackage{booktabs}
\usepackage[normalem]{ulem}

\title{WaveletDiff: Multilevel Wavelet Diffusion For Time Series Generation}

\author{\name Yu-Hsiang Wang \email yw121@illinois.edu \\
      \addr Department of Electrical and Computer Engineering\\
      University of Illinois Urbana-Champaign
      \AND
      \name Olgica Milenkovic \email milenkov@illinois.edu \\
      \addr Department of Electrical and Computer Engineering\\
      University of Illinois Urbana-Champaign}

\def\month{09}  
\def\year{2026} 
\def\openreview{\url{https://openreview.net/forum?id=YSTJ71fDBm}} 

\begin{document}

\maketitle

\begin{abstract}
Time series are ubiquitous in many applications that involve forecasting, classification and causal inference tasks, such as healthcare, finance, audio signal processing and climate sciences. Still, large, high-quality time series datasets remain scarce. Synthetic generation can address this limitation; however, current models confined either to the time or frequency domains struggle to reproduce the inherently multi-scaled structure of real-world time series.
We introduce WaveletDiff, a new framework that trains diffusion models \emph{directly on wavelet coefficients} to exploit the inherent multi-resolution structure of time series data. The model combines dedicated transformers for each decomposition level with cross-level attention mechanisms that enable selective information exchange between temporal and frequency scales through adaptive gating. It is also informed by level-specific energy constraints based on Parseval's theorem which preserve time-frequency properties throughout the diffusion process. Comprehensive tests across six real-world datasets from energy, finance, and neuroscience domains demonstrate that WaveletDiff outperforms the diffusion baselines FourierDiffusion, Diffusion-TS, and SigDiffusions on the majority of metrics, with the smallest margin over FourierDiffusion, while still achieving roughly $3\times$ lower discriminative and Context-FID scores. Against the VAE/transformer-based MSDformer, the results are mostly comparable, with WaveletDiff using fewer parameters and less training time on most datasets. The most revealing finding is the significant performance gap on fMRI data (in favor of MSDformer) and EEG (in favor of WaveletDiff). This finding is explained via a careful testing/examination of the properties of wavelet coefficients for generative, as opposed to analyses/decomposition tasks. Our code is available at \href{https://github.com/GarlicWang/WaveletDiff}{https://github.com/GarlicWang/WaveletDiff}.
\end{abstract}

\section{Introduction}
Time series data arises in diverse practical settings, including healthcare~\citep{lee2019dynamic, vanderSchaar2019survival}, finance~\citep{sezer2020financial, ozbayoglu2020deep}, climate sciences~\citep{dinku2019challenges}, audio processing~\citep{mitra2025music} and engineering~\citep{susto2020predictive, lei2020survey}. Due to various constraints, acquiring sufficiently high-quality labeled time-series datasets remains a challenge~\citep{wang2021deep, desai2025timeseries}. The problem may be mitigated through synthetic time series generation, which also offers promising solutions for data augmentation~\citep{wen2021time, forestier2017generating, ryu2023simpsi}, privacy preservation~\citep{wang2020partgan, jordon2022synthetic, nosowsky2006hipaa}, forecasting~\citep{taga2025timepfn} and simulations~\citep{nikolenko2021synthetic, elEmam2020synthetic}.

Current time series generation methods predominantly operate either directly in the time domain or frequency domain, and come with different advantages and  limitations. Time-domain approaches, including those based on GANs~\citep{yoon2019time, pei2021generatingrealworldtimeseries, eskandarinasab2024seriesgantimeseriesgeneration}, autoregressive~\citep{salinas2020deepar} and diffusion models~\citep{lim2023regulartimeseriesgenerationusing, 10.5555/3692070.3693584, sikder2024transfusiongeneratinglonghigh} are well-suited for modeling local temporal patterns, but struggle with long-term dependencies and preservation of important spectral characteristics. To address time-domain induced limitations, recent approaches have increasingly leveraged frequency-domain analysis, often along with temporal modeling~\citep{tian2020tfgantimefrequencydomain, chi2024rfdiffusionradiosignalgeneration, crabbe2024time, huang2024generative}. These methods are of relevance since many real-world time series tend to exhibit higher localization in the frequency rather than the time domain. Representative methods include FourierFlow~\citep{alaa2021generative}, which applies normalizing flows to Fourier representations, Diffusion-TS~\citep{yuan2024diffusionts}, which combines Fourier decompositions with diffusion models, and various frequency-enhanced transformers~\citep{zhou2022film, zhou2022fedformer, xu2023fits, yi2023frets} that use both spectral and temporal analyses. However, these approaches typically process time and frequency domain information either in a separate manner or impose trade-offs between temporal resolution and spectral coherence. They are also not able to simultaneously capture both local and global time and spectral patterns, which is crucial for synthesizing realistic time series.

Wavelet transforms represent a natural approach to address the above issue by creating a multi-resolution representation that simultaneously captures both temporal and spectral information~\citep{mallat1989theory, cohen2001uncertainty}. Unlike the Fourier transform, which captures global frequency properties, wavelets maintain temporal localization while also providing useful decompositions into multiple frequency bands~\citep{rioul1992timefrequency, daubechies1988timefrequency}. This results in a highly versatile time-frequency hierarchical representation~\citep{mallat1989theory}. As a result, wavelet-based analyses have been used with success for various signal processing applications including speech recognition, financial trends analysis, image processing, and biomedical signal analysis~\citep{daubechies1992ten, vetterli1995wavelets, burrus1998introduction}. Despite these results, a handful of known wavelet-based approaches for time series generation have failed to provide improvements over Fourier-based methods~\citep{takahashi2024generation, kazemi2022timeseriessynthesismultiscale, 8595010}. This may be attributed to the fact that, almost exclusively, the methods treat wavelet coefficients as image structures and then follow up by applying standard image generation techniques such as convolutional neural networks or image-based diffusion models. While potentially useful for data-poor applications and highly specialized time series, indirect time series $\to$ wavelet $\to$ image conversion methods in general suffer from pattern distortions caused by noninvertible image features. 

A more adequate approach based on wavelet decompositions is to run diffusion models directly in the wavelet domain, which is a new direction proposed in this work. For diffusion, methods such as denoising diffusion probabilistic models (DDPMs)~\citep{ho2020denoising} which have demonstrated remarkable success in image~\citep{dhariwal2021diffusion}, audio~\citep{kong2021diffwave}, and text generation~\citep{austin2021structured}, are considered state-of-the-art for time-series generation. However, these diffusion models are tailor-made for highly specific time-series formats (e.g., audio or financial data), and may not be suitable for other modalities. This motivates us to implement a new wavelet-space diffusion model, termed WaveletDiff, which is universally applicable as it inherently respects different multi-level structures. Unlike frequency-domain approaches, WaveletDiff also captures temporal patterns at different scales simultaneously. Our key innovations lie in running forward diffusion processes for each wavelet level individually and in parallel, following fine-tuned exponential noise mechanisms and using dedicated level-transformer denoising networks combined with a cross-level attention mechanism that enables information exchange between different decomposition levels. This design preserves the hierarchical nature of wavelet representations while allowing the model to learn complex inter-scale dependencies crucial for realistic time series generation. 

Switching from time to frequency domain diffusion models has led to strong empirical gains and established transform-domain data generation as a compelling direction. WaveletDiff offers significant improvements over the only other direct-transform approach (Fourier), and these gains come from design choices that exploit the multiresolutional structure of wavelet decompositions — per-level processing, time–frequency coupling and cross-level attention. As a result, wavelet-based diffusion represents a new generative modeling method with potentially many additional applications.

Our technical contributions include: \\
\textbf{1.} A diffusion framework that operates directly in the wavelet domain and tunes the noise addition process to the approximation and detail levels, identifies the most suitable choice of mother wavelet for different time series and uses level-specific loss functions and transformers.\\
\textbf{2.} A cross-scale attention mechanism that enables information flow between different temporal scales while preserving their individual properties.\\
\textbf{3.} A wavelet-aware loss weighting mechanism that prevents some levels from dominating the training objective through level-specific balancing strategies.\\
\textbf{4.} A new evaluation metric based on the Dynamic Time-Warping distance and extensive comparative analysis of both short and long time series datasets, including ETTh1, ETTh2, Stocks, Exchange Rate, fMRI, EEG~\citep{haoyietal-informer-2021, lai2018modelinglongshorttermtemporal, eeg_eye_state_264}. The results show significant performance gains of WaveletDiff compared to Fourier and time-based methods, which are roughly three-fold on average for discriminative and Context-FID scores. \\ 
\textbf{5.} Our results also reveal for the first time that signal features amenable for wavelet analysis (e.g., as is the case for fRMI data) may hurt generative models that operate in the wavelet domain. \\
\textbf{6.} The first empirical evaluation of reproducibility of diffusion models for time series, akin to recent efforts reported for images~\citep{pmlr-v235-zhang24cn, li2024understanding, kadkhodaie2024generalization}, as well as an added verification of WaveletDiff being robust in the context of irregularly sampled time series generation.

\section{Related Work}
\subsection{Time Series Generation with Diffusion Models}
Early generative AI methods for time series focused on conditional generation tasks such as forecasting and imputation~\citep{rasul2021autoregressive,tashiro2021csdi,li2022csdi,NEURIPS2024_5e364212,pei-etal-2025-leaf,hou2025stagediffstagewiselongtermtime,shen2024multiresolution,fadlon2025diffusionmodelregulartime}, while recent approaches target unconditional time series generation~\citep{shen2023timediff,barancikova2025sigdiffusions,Li_2025,suh2025timeautodiffunifiedframeworkgeneration,naiman2024generative}. The above methods employ various architectural choices including RNNs, transformers, and specialized denoising networks capable of handling sequential temporal data~\citep{kong2021diffwave,naiman2024utilizingimagetransformsdiffusion}, and almost exclusively operate in the time domain. This limits their ability to capture \emph{both} global and local, potentially varying, spectral properties.

\subsection{Frequency and Other Transform Domain Approaches for Time Series}
Recent works have demonstrated that real-world time series are more localized in the frequency domain, making spectral diffusion more effective than time diffusion~\citep{crabbe2024time}. Various frequency-based approaches include lightweight models using complex-valued operations~\citep{xu2023fits}, frequency-enhanced transformers combining discrete Fourier transform (DFT) with attention mechanisms~\citep{zhou2022fedformer,zhou2022film}, and specialized MLP architectures for frequency learning~\citep{yi2023frets}. Additional methods incorporate spectral filtering~\citep{zhang2024atfnet}, multi-resolution frequency analysis~\citep{wang2024spectral}, and normalizing flows within the Fourier domain~\citep{alaa2021generative}. Additional unconditional generation approaches include interpretable diffusion models that combine \emph{trend and seasonality} components~\citep{yuan2024diffusionts} and latent diffusion models that operate in compressed latent spaces for more efficient generation~\citep{qian2024timeldmlatentdiffusionmodel}. These methods often require separate processing pipelines for temporal and frequency components, limiting their ability to simultaneously capture multi-scale temporal-spectral relationships.

Several other lines of work perform diffusion in transformed domains to expose spectral or multi-scale structure that is difficult to model in the time domain alone. Frequency-domain approaches leverage Fourier or frequency-aware representations to better control global spectral properties and periodic patterns~\citep{crabbe2024time, luo2025waveletfourierdiffuserfrequencyaware,naiman2024utilizingimagetransformsdiffusion}. While transform-domain diffusion for time series remains relatively rare, diffusion models have been explored in other transformed domains, primarily for audio and vision. Spectrogram-domain diffusion leverages short-time Fourier transform (STFT) representations to capture localized time–frequency structure for audio and music generation~\citep{zhu2023edmsoundspectrogrambaseddiffusion, vanukuri2025wavesimaginationunconditionalspectrogram, hawthorne2022multiinstrumentmusicsynthesisspectrogram}. Scattering-transform–based approaches use wavelet scattering features as conditioning to guide diffusion, benefiting from their stability to deformations and multi-scale sensitivity~\citep{10.1145/3746027.3762245, 10890904, gama2018diffusionscatteringtransformsgraphs}. Discrete cosine transform (DCT)–based diffusion performs generation directly in the DCT space, exploiting compact and decorrelated representations for efficiency and downstream tasks~\citep{ning2024dctdiff, 10975753}.

\subsection{Wavelet-Based Time Series Modeling}
Wavelet transforms provide multi-resolution time-frequency representation capabilities~\citep{mallat1989theory,daubechies1992ten,addison2017wavelet} and have been extensively used in time series analysis~\citep{percival2000wavelet, sang2013review, waveletadvantages2015, han2019twavenet}.
For generative modeling, wavelets have seen more limited applications, despite showing promise in terms of direct coefficient processing~\citep{phung2022wavelet,hu2023neuralwaveletdomaindiffusion3d,NEURIPS2022_03474669}. For time series forecasting, wavelets have been employed to enhance traditional forecasting models through their multi-resolution time-frequency analysis capabilities~\citep{zhou2025multiorderwaveletderivativetransform,Sasal_2022,arabi2024wavemaskmixexploringwaveletbasedaugmentations,Schlter2010UsingWF, 11197480, zhang2017financial}. For generation tasks, existing methods predominantly convert wavelet coefficients to image representations for processing with standard computer vision techniques~\citep{takahashi2024generation, kazemi2022timeseriessynthesismultiscale}. However, it is not clear that these indirect approaches fully exploit the hierarchical, global/local multi-scale structure of wavelet decompositions, due to the use of patches and/or potential incompatibilities between the image representations and actual level-dependent temporal and spectral characteristics. As a result, their  generative capabilities are modest for long time series with long-range contexts.

\section{Methodology}
\subsection{Wavelet Representations of Time Series}
\label{subsec:wavelet_representation}

\begin{figure}
    \centering
    \includegraphics[width=0.7\linewidth]{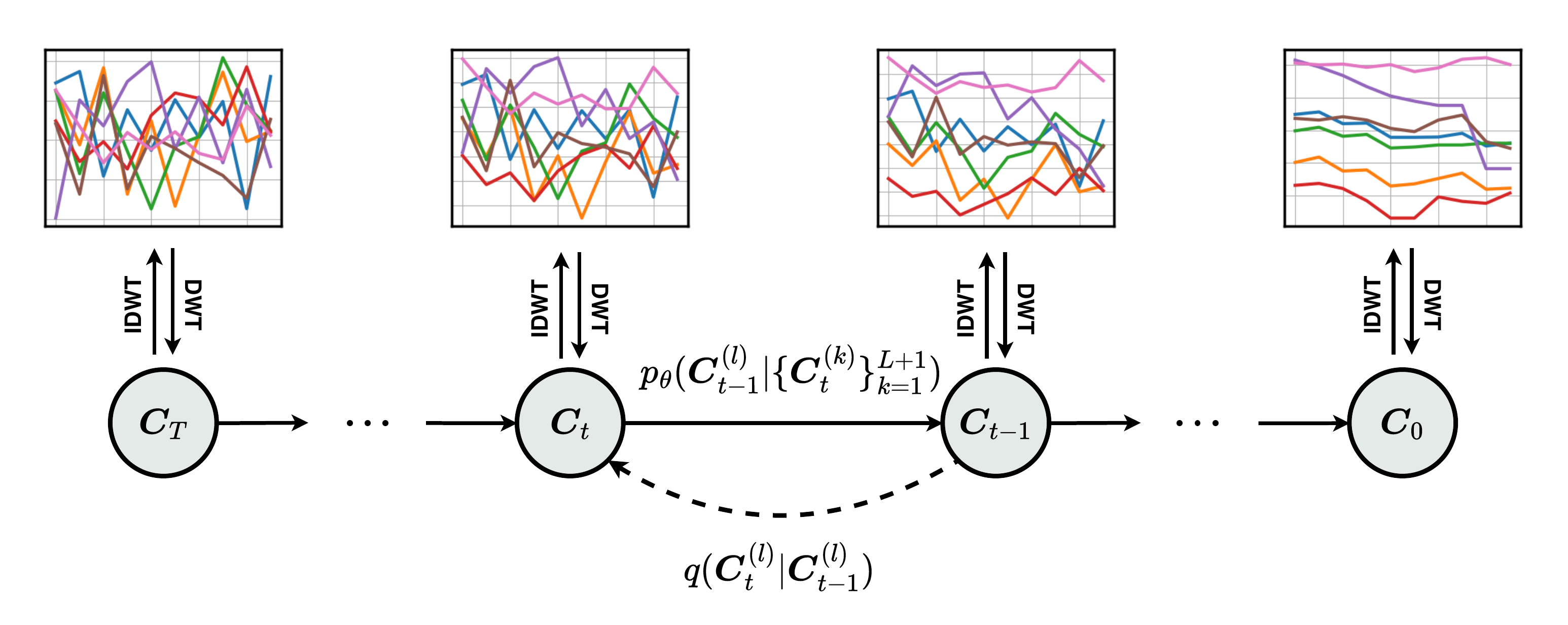}
    \caption{Direct wavelet coefficient diffusion, where the forward process proceeds independently at each decomposition level, while the reverse process integrates information across all levels for joint denoising.}
    \label{fig:diffusion}
\end{figure}

A multivariate time series dataset $\mathbf{X}\in \mathbb{R}^{N \times T \times D}$ with $N$ samples, $T$ timesteps and $D$ features (e.g., opening price, closing price, high/low, volume for financial data) comprises time series of the form $\mathbf{x}^{(i)}=[\mathbf{x}_0^{(i)}, \mathbf{x}_1^{(i)},\dots,\mathbf{x}_{T-1}^{(i)}]\in\mathbb{R}^{T \times D}$, where $i\in[1,N]$. The Discrete Wavelet Transform (DWT) decomposes each time series through a cascade of high-pass and low-pass filtering operations followed by downsampling. The decomposition utilizes a scaling function $\phi(t)$ and its associated \emph{mother wavelet $\psi(t)$}. The mother wavelet is characterized by its order $p$, which ensures that the wavelet is orthogonal to all polynomials of degree less than $p$ (hence, $p$ determines the number of vanishing moments). Higher-order wavelets provide better frequency localization but require longer filters. The filter length $F$ represents the number of nonzero coefficients in the discrete filters, which depends on the wavelet family and order $p$ (e.g., $F = 2p$ for Daubechies wavelets). More details are available in Appendix~\ref{appendix:wavelet}. These functions satisfy the two-scale relations:
\begin{equation}
\label{eq:wavelet_scale}
\psi(t) = \sqrt{2} \sum_{k=0}^{F-1} g_k \, \phi(2t - k), \qquad
\phi(t) = \sqrt{2} \sum_{k=0}^{F-1} h_k \, \phi(2t - k).
\end{equation}
where $\{g_k\}_{k=0}^{F-1}$ and $\{h_k\}_{k=0}^{F-1}$ are the high-pass and low-pass filter coefficients, respectively, with the relationship $g_k = (-1)^k h_{F-1-k}$ ensuring orthogonality. The DWT performs recursive decomposition over $L$ levels. Starting with the approximation coefficients $\boldsymbol{A}^{(0)}=\mathbf{X}$, at each level $l \in [1,L]$, we apply high-pass and low-pass filters followed by temporal-dimension downsampling:
\begin{equation}
\begin{aligned}
\boldsymbol{C}^{(l)}_{:,m,:} = \sum_{k} g_k \, \boldsymbol{A}^{(l-1)}_{:,2m-k,:}\ \text{(detail coeff.)}, \\
\boldsymbol{A}^{(l)}_{:,m,:} = \sum_{k} h_k \, \boldsymbol{A}^{(l-1)}_{:,2m-k,:}\ \text{(approximate coeff.)},
\end{aligned}
\end{equation}
where $m$ indexes the downsampled time dimension and the operation is applied independently across all $N$ samples and $D$ features.
Boundary effects are handled using symmetric extension, where the signal is mirrored at the tails to ensure sufficient coefficients for filtering operations. This decomposition yields the wavelet coefficient representation:
\begin{equation}
    \text{DWT}(\mathbf{X}) = \{\boldsymbol{C}^{(1)}, \ldots, \boldsymbol{C}^{(L)}, \boldsymbol{A}^{(L)}\},
\end{equation}
where $\boldsymbol{C}^{(l)} \in \mathbb{R}^{N \times d_l \times D},$ for $l\in[1,L],$ are the detail coefficients and $\boldsymbol{A}^{(L)}\in\mathbb{R}^{N \times d_{L} \times D}$ are the approximation coefficients. For consistency with diffusion notation, we write $\boldsymbol{C}^{(L+1)}=\boldsymbol{A}^{(L)}$.

The wavelet order $p$ is chosen based on the sequence length to ensure sufficient coefficients at each level, with longer sequences accommodating higher-order wavelets for better frequency localization. The coefficient dimension at each level $l$ is calculated recursively as:
\begin{equation}
d_{l}=\lfloor\frac{d_{l-1}+F-1}{2}\rfloor,\quad l=1,\dots,L,
\end{equation}
where $d_0=T$ is the original sequence length. This formula accounts for the filter overlap (requiring $F-1$ additional boundary coefficients) and dyadic downsampling (division by $2$) inherent to the wavelet decomposition process. The number of decomposition levels $L$ is determined based on the sequence length $T$ to ensure sufficient coefficients are available at each level while maintaining meaningful frequency separation, and is set to $L = \max\left(3, \min\left(7, \left\lfloor \log_2\left(\frac{T}{F - 1}\right) \right\rfloor\right)\right)$ in practice.

To reconstruct time series from diffusion-generated wavelet coefficients $\{\hat{\boldsymbol{C}}^{(1)},\dots,\hat{\boldsymbol{C}}^{(L)}, \hat{\boldsymbol{C}}^{(L+1)}\}$, we apply the Inverse Discrete Wavelet Transform (IDWT). The reconstruction proceeds from the coarsest level to the finest level. Starting with $\hat{\boldsymbol{A}}^{(L)}=\hat{\boldsymbol{C}}^{(L+1)}$, for each $l=L,\dots,1,$ we compute:
\begin{equation}
    \hat{\boldsymbol{A}}^{(l-1)}_{:,m,:} = \sum_{k} \tilde{h}_{m-2k} \hat{\boldsymbol{A}}^{(l)}_{:,k,:} + \sum_{k} \tilde{g}_{m-2k} \hat{\boldsymbol{C}}^{(l)}_{:,k,:},
\end{equation}
where $\tilde{h}$ and $\tilde{g}$ are the synthesis filters used for reconstruction. For orthogonal wavelets, these take the form $\tilde{h}_k = h_{-k}$ and $\tilde{g}_k = g_{-k}$, while for biorthogonal wavelets, they are independently designed dual filters that ensure perfect reconstruction. The reconstruction combines the current approximation $\hat{\boldsymbol{A}}^{(l)}$ with the detail coefficients $\hat{\boldsymbol{C}}^{(l)}$ through upsampling and filtering. The inverse transform can hence be written as:
\begin{equation}
    \hat{\mathbf{X}} = \text{IDWT}(\{\hat{\boldsymbol{C}}^{(1)}, \ldots, \hat{\boldsymbol{C}}^{(L)}, \hat{\boldsymbol{A}}^{(L)}\}),
\end{equation}
where $\hat{\mathbf{X}}=\hat{\boldsymbol{A}}^{(0)}\in\mathbb{R}^{N\times T\times D}$ is the reconstructed time series.

\subsection{Wavelet-Space Diffusion Framework}
\label{subsec:diffusion_framework}

\begin{figure}[t]
    \centering
    \begin{subfigure}[b]{0.2\textwidth}
        \centering
        \includegraphics[width=\textwidth]{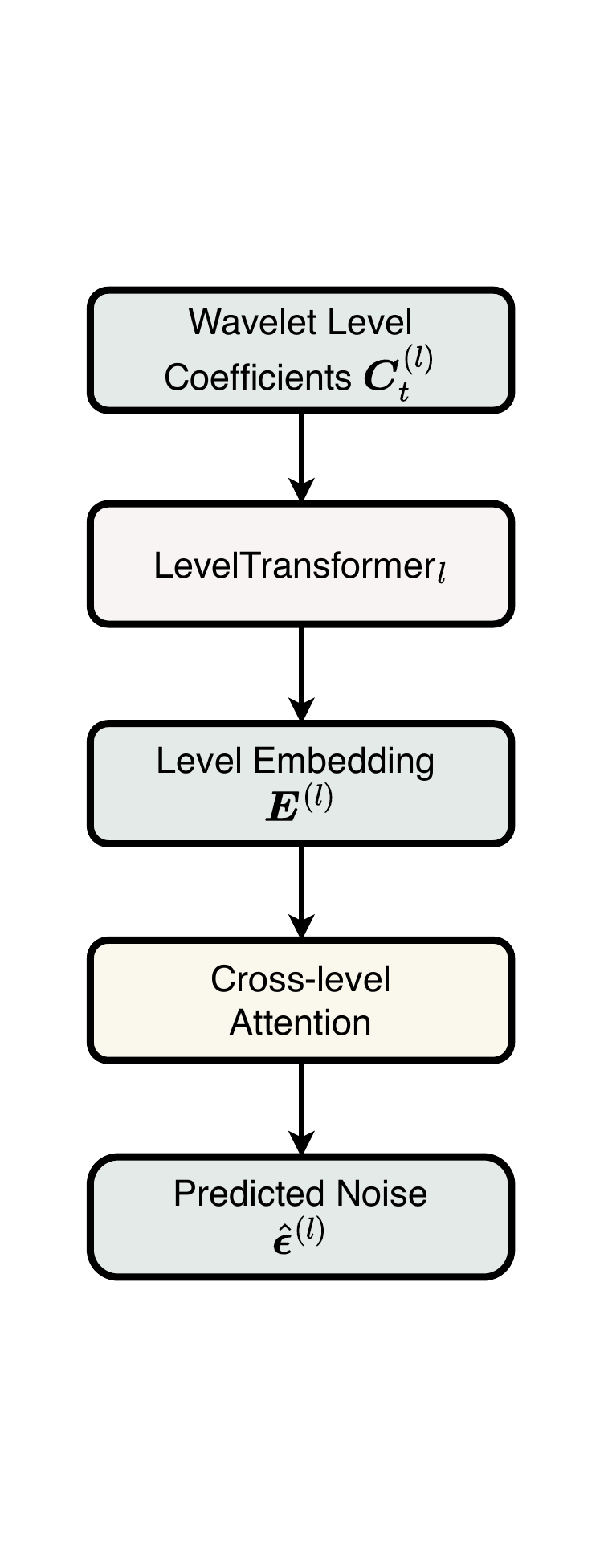}
        \caption{Denoising Network}
        \label{fig:subfig1}
    \end{subfigure}
    \hfill
    \begin{subfigure}[b]{0.22\textwidth}
        \centering       \includegraphics[width=\textwidth]{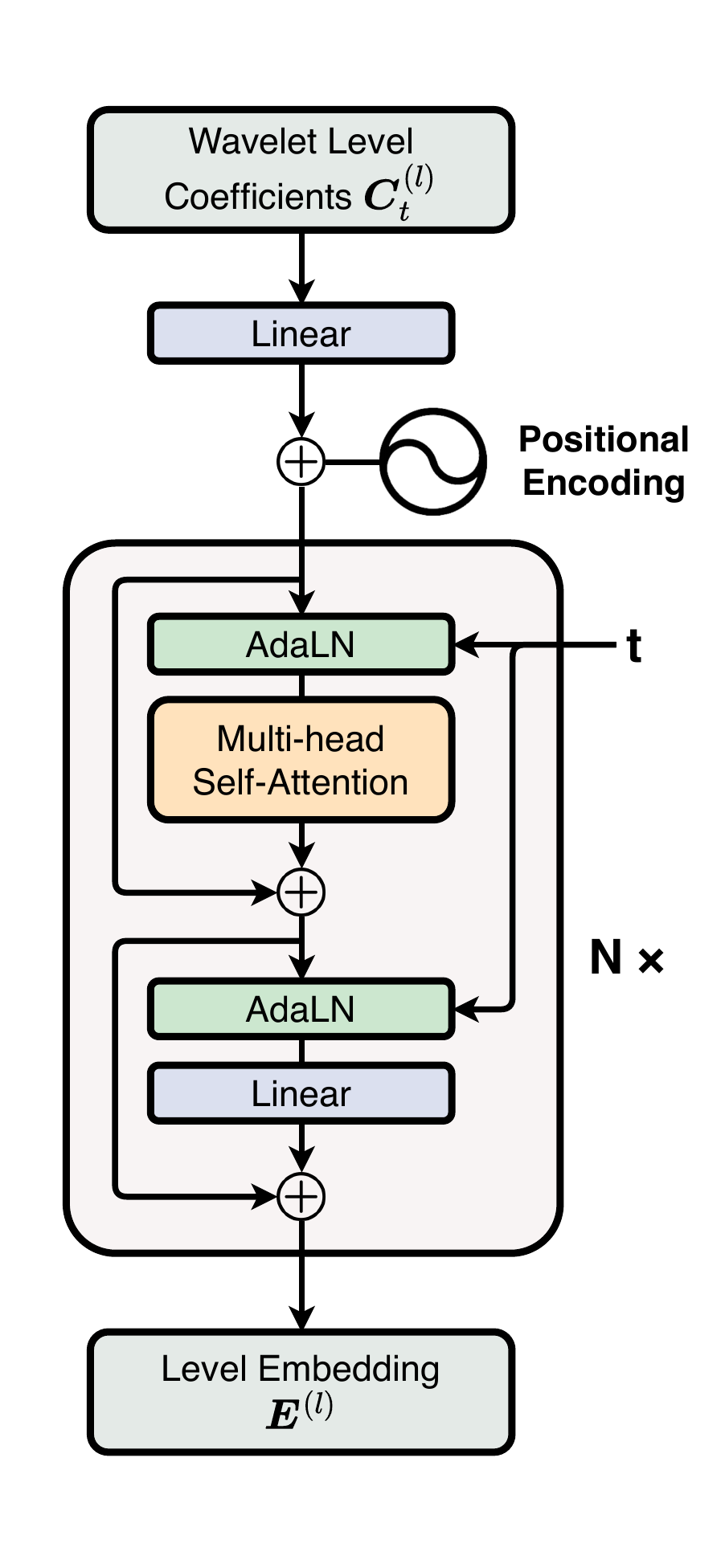}
        \caption{LevelTransformer}
        \label{fig:subfig2}
    \end{subfigure}
    \hfill
    \begin{subfigure}[b]{0.3\textwidth}
        \centering     \includegraphics[width=\textwidth]{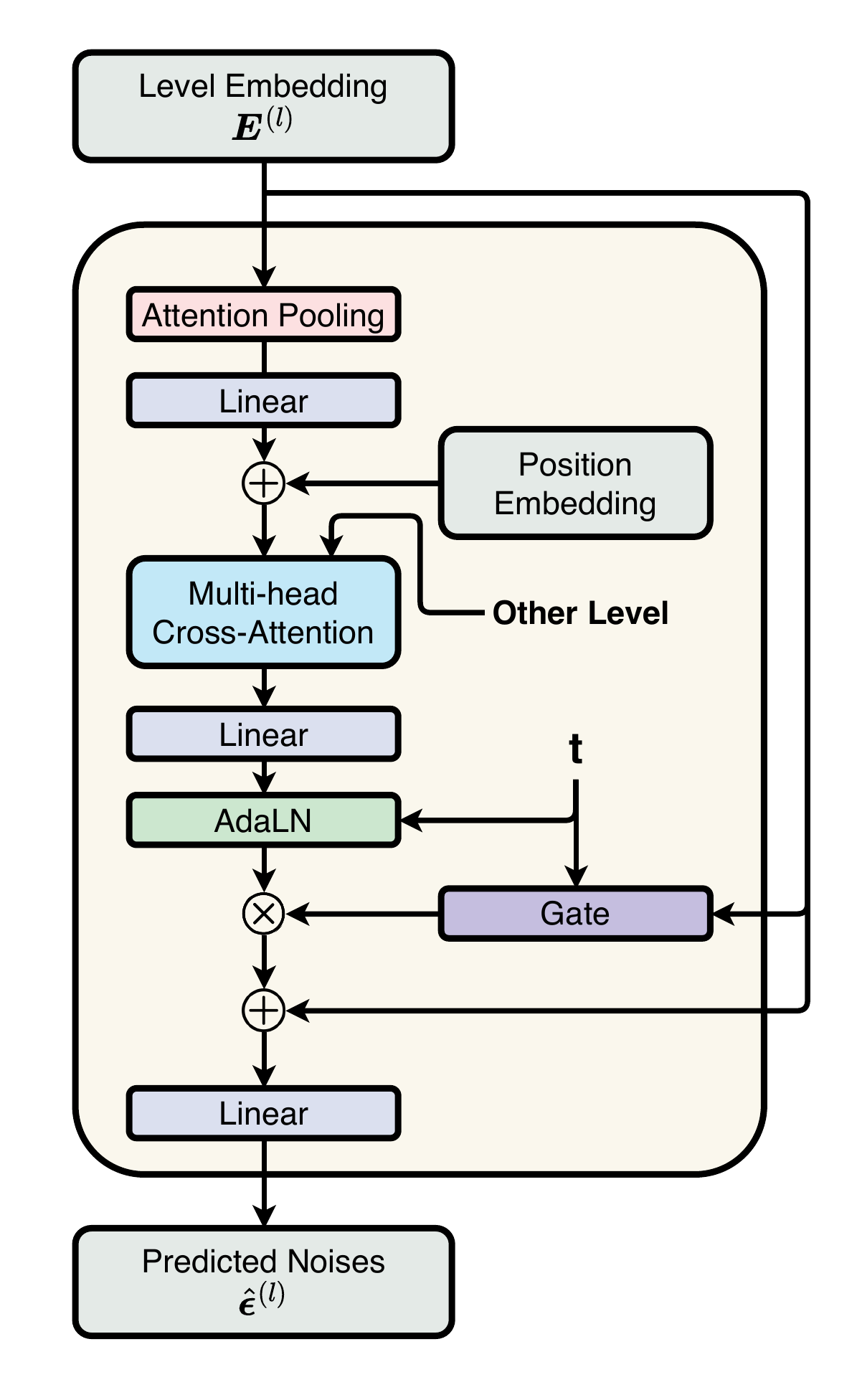}
        \caption{Cross-level Attention}
        \label{fig:subfig3}
    \end{subfigure}
    \caption{The wavelet coefficients are independently processed by LevelTransformers to obtain level-specific embeddings. These embeddings are obtained through interaction across levels via a cross-level attention module based on  adaptive gating mechanisms.}
    \label{fig:main}
\end{figure}

We propose to run the diffusion process on wavelet coefficients using Denoising Diffusion Probabilistic Models (DDPM), as shown in Figure~\ref{fig:diffusion}. The forward diffusion process in the wavelet domain gradually adds Gaussian noise to coefficients at all levels independently:
\begin{equation}
q(\boldsymbol{C}^{(l)}_t | \boldsymbol{C}^{(l)}_0) = \mathcal{N}(\boldsymbol{C}^{(l)}_t; \sqrt{\bar{\alpha}_t}\boldsymbol{C}^{(l)}_0, (1-\bar{\alpha}_t)\mathbf{I}),
\label{eq:forward_diffusion}
\end{equation}
where $l=1,\dots,L+1$, $\alpha_t = 1 - \beta_t$ and $\bar{\alpha}_t = \prod_{s=1}^t \alpha_s$ follow standard DDPM schedules. Specifically, we adopt an exponential noise schedule $\beta_t = \beta_{start} + (\beta_{end} - \beta_{start}) \cdot \left(1 - e^{-\gamma \cdot t}\right),$ where $\gamma$ is the exponential decay rate, $t \in [0,1]$ is the normalized timestep, and $\beta_{start}\text{ and }\beta_{end}$ are tuneable hyperparameters. Exponential schedules are better suited for wavelet-based time series generation than cosine schedules. This may be because cosine schedules start slowly, peak mid-epoch, and then decrease, with this smooth behavior stabilizing high-dimensional data like images. In contrast, time series and their wavelet decompositions have significantly lower dimensions, benefiting from more aggressive noise injection when coupled with transformer-based denoising models used at early backwards steps. 

We parameterize the reverse process using a cross-level transformer network that employs cross-attention to enable communication across levels:
\begin{equation}
p_\theta(\boldsymbol{C}^{(l)}_{t-1} | \{\boldsymbol{C}^{(k)}_t\}_{k=1}^{L+1}) = \mathcal{N}(\boldsymbol{C}^{(l)}_{t-1}; \boldsymbol{\mu}_\theta(\{\boldsymbol{C}^{(k)}_t\}_{k=1}^{L+1}, t), \boldsymbol{\Sigma}_\theta),
\end{equation}
where $\boldsymbol{\mu}_\theta(\{\boldsymbol{C}^{(k)}_t\}_{k=1}^{L+1}, t)$ represents the predicted mean of the reverse diffusion process, parameterized by the neural network $\theta$ and conditioned on all wavelet levels and the diffusion timestep $t$, and $\boldsymbol{\Sigma}_\theta$ is the predicted covariance matrix. Following the DDPM framework, we fix $\boldsymbol{\Sigma}_\theta=\beta_t\mathbf{I}$ and train the denoising network $f_\theta$ to predict the added noise $\boldsymbol{\epsilon}$ using the mean square error (MSE) as the loss:
\begin{equation}
\hat{\boldsymbol{\epsilon}}^{(l)} = f_\theta(\{\boldsymbol{C}^{(k)}_t\}_{k=1}^{L+1}, t),\quad l=1,\dots,L+1
\end{equation}
\begin{equation}
\mathcal{L}_{\text{recon}} = \mathbb{E}_{\boldsymbol{C},t,\mathbf{\epsilon}}\left[ \sum_{l=1}^{L+1} w_l \cdot \|\boldsymbol{\epsilon}^{(l)} - \hat{\boldsymbol{\epsilon}}^{(l)}\|^2 \right]
\label{eq:recon_loss}
\end{equation}
where $\boldsymbol{\epsilon}^{(l)}$ and $\hat{\boldsymbol{\epsilon}}^{(l)}$ are the true and predicted noise at level $l$, and $w_l$ are level-specific weights ensuring balanced contribution across scales. 

To preserve data spectra for long sequence generation, we optionally introduce an energy conservation penalty based on Parseval's theorem. For orthogonal wavelet bases, we let
\begin{equation}
\mathcal{L}_{\text{energy}} = \mathbb{E}_{\boldsymbol{C},t,\mathbf{\epsilon}} \left[ \sum_{l=1}^{L+1} \left|\mathcal{E}^{(l)} - \hat{\mathcal{E}}^{(l)}\right| \right],
\end{equation}
where $\mathcal{E}^{(l)} = \sum_{n=1}^{N} \sum_{j=1}^{d_l} \sum_{k=1}^{D} (\boldsymbol{C}^{(l)}_{n,j,k})^2$ represents the true energy at wavelet level $l$, and $\hat{\mathcal{E}}^{(l)} = \sum_{n=1}^{N} \sum_{j=1}^{d_l} \sum_{k=1}^{D} (\hat{\boldsymbol{C}}^{(l)}_{n,j,k})^2$ represents the predicted energy at wavelet level $l$. By enforcing energy preservation at each decomposition level, the constraints stabilize training and preserve the natural energy distribution across frequency scales. For biorthogonal and reverse biorthogonal wavelet bases, the corresponding energy regularization formulas are provided in Appendix~\ref{appendix:biorthogonal_parseval}.

The overall training objective combines both the reconstruction and energy terms $\mathcal{L} = \mathcal{L}_{\text{recon}} + \lambda_{\text{energy}} \mathcal{L}_{\text{energy}},$ where $\lambda_{\text{energy}}$ denotes the weight of the energy loss term. For short sequences, the base reconstruction loss is typically sufficient since the spectral energy drift is minimal over limited temporal horizons. The energy preservation term mostly benefits datasets with strong low-frequency trends and smooth spectral characteristics (e.g., ETTh1, Exchange Rate), while high-volatility datasets with abrupt changes (e.g., Stocks) are better reproduced through the reconstruction loss alone.

The denoising network $f_\theta$ uses dedicated transformers for each wavelet level. We adopt Adaptive Layer Normalization (AdaLN)~\citep{Peebles2022DiT} as the normalization layer. For level $l$, the coefficients $\boldsymbol{C}^{(l)}_t$ are processed through a specialized transformer,
\begin{equation}
\mathbf{E}^{(l)} = \text{LevelTransformer}_l(\boldsymbol{C}^{(l)}_t, \mathbf{t}),\quad l=1,\dots,L+1,
\end{equation}
where $\mathbf{t}$ denotes the diffusion time embedding and $\mathbf{E}^{(l)} \in \mathbb{R}^{N \times h_l \times D}$ represents the output level embeddings for level $l$, with $h_l$ denoting the embedding dimension at that level. 
The embeddings of each level are aggregated through attention-based pooling. Cross-level attention operates on these aggregated representations, allowing each level to adaptively incorporate contextual information from other scales through learned gating mechanisms (Figure~\ref{fig:main}).

\paragraph{Discussion: Choice of the Wavelet Domain.}
While diffusion models have been explored in several transform domains, including Fourier, spectrogram, scattering, and DCT representations, only Fourier-domain diffusion has been directly adopted for generic time series generation. Fourier-domain approaches provide a global frequency decomposition that is well suited for modeling stationary signals, but they lack temporal localization. As a result, Fourier coefficients entangle long- and short-term dynamics, making it difficult for diffusion models to capture transient events and regime changes that are ubiquitous in real-world time series.

On the other hand, the wavelet domain provides a multiresolution decomposition that for many types of time series separates global low-frequency structure from localized high-frequency details, naturally supporting level-wise modeling and denoising in diffusion models. Nevertheless, as will be discussed in the results section, one has to be careful to distinguish the utility of different wavelet families for signal analysis versus amenability to diffusion in the wavelet domain - in some cases, diffusion models may be outperformed by other generative modalities. The latter hinges on the statistical properties of the coefficients which in turn depend on the type of time-series to be generated.

Hence, rather than relying on qualitative arguments alone, we explicitly evaluate the effect of the transform-domain choice by comparing wavelet-domain diffusion with Fourier-domain diffusion baselines~\citep{crabbe2024time} in Section~\ref{sec:experiments}, with an additional focus on the choice of the wavelet family.

\section{Experiments}
\label{sec:experiments}
\subsection{Experimental Settings}\label{sec:exp_settings}
\paragraph{Benchmarks} 
We compare WaveletDiff to several state-of-the-art time series generation methods, including FourierDiffusion~\citep{crabbe2024time}, Diffusion-TS~\citep{yuan2024diffusionts}, TimeGAN~\citep{yoon2019time}, SigDiffusions~\citep{barancikova2025sigdiffusions}, KoVAE~\citep{naiman2024generative}, as well as MSDformer~\citep{feng2025msdformer}, which is an improved version of SDformer~\citep{chen2024sdformer}. In addition, we make use of two state-of-the-art forecasting time series foundation models (TSFMs), TimesFM~\citep{das2024decoderonlyfoundationmodeltimeseries} and Chronos~\citep{ansari2024chronoslearninglanguagetime}, each comprising $\sim$200M parameters, to perform unconditional generation. Specifically, we sampled length-64 windows as context and autoregressively forecasted the next $24$ steps to produce newly generated sequences. Although not fine-tuned for generation, these models leverage real-world inputs during generation and have large capacity. We therefore believe that considering this approach also provides a meaningful comparison.

\paragraph{Datasets}
We use six real-world datasets to evaluate our method, covering energy, finance, and neuroscience domains. \textbf{ETTh1} and \textbf{ETTh2}~\citep{haoyietal-informer-2021} are electricity transformer datasets containing oil temperature and six power load features recorded hourly from 2016 to 2018. \textbf{Stocks} is a multivariate financial time series dataset containing historical Google stock market data with price and volume features from 2004 to 2019. \textbf{Exchange Rate}~\citep{lai2018modelinglongshorttermtemporal} contains daily exchange rates of eight countries from 1990 to 2016. \textbf{fMRI} is the NetSim dataset containing simulated BOLD time series data for evaluating network modeling methods in functional magnetic resonance imaging. \textbf{EEG}~\citep{eeg_eye_state_264} contains multichannel electroencephalogram recordings that measure brain electrical activity over time, offering information about neural dynamics and cognitive processes. For more details, refer to Appendix~\ref{appendix:datasets}.

\paragraph{Metrics}
We evaluate generation quality using five complementary metrics. The \textbf{discriminative score} measures similarity between real and generated samples by training a binary classifier to distinguish them~\citep{yoon2019time}. The \textbf{predictive score} assesses the utility of synthetic data for forecasting real sequences using mean absolute error. \textbf{Context-Fréchet inception distance (Context-FID)}~\citep{paul2022psaganprogressiveselfattention} quantifies distributional distance using TS2Vec~\citep{yue2022ts2vecuniversalrepresentationtime} embeddings following~\citep{yuan2024diffusionts}. The \textbf{correlational score} evaluates temporal dependencies by comparing cross-correlation matrices. 

Although Context-FID quantifies distributional distance, it measures distance in the embedding space of a trained TS2Vec encoder, and therefore inherits whatever structural biases that encoder may have. In particular, it is unclear whether the encoder adequately captures temporal alignment, phase shifts, or nonlinear time distortions. Dynamic Time Warping (DTW), on the other hand, is a model-free, alignment-aware similarity measure that explicitly accounts for temporal warping by finding the optimal monotone alignment between two sequences. This makes DTW sensitive to shape-based similarity in a way that embedding-based distances may not be. We therefore propose the \textbf{Dynamic Time Warping Jensen--Shannon Distance (DTW-JS distance)}, which combines DTW with Jensen--Shannon divergence to assess the differences in the distributions of DTW-based discrepancies between real and generated data. DTW aims to capture optimal temporal alignments between two time series sequences $x$ and $y$ by minimizing
\begin{equation}
\text{DTW}(x, y) = \min_{\pi} \sum_{(i,j) \in \pi} |x_i - y_j|,
\end{equation}
where $\pi$ is a warping path allowing flexible temporal matching. We first create a reference set $\mathcal{M}$ by randomly sampling sequences from the union of the real $\mathcal{R}$ and generated $\mathcal{G}$ datasets, which are matched in size. For each sequence $s$ in the real dataset, we compute its mean DTW distance to all sequences in the reference set: $d_{\mathcal{R}}(s)=\frac{1}{|\mathcal{M}|}\sum_{r\in\mathcal{M}}\text{DTW}(s,r)$. We perform the same calculation for each generated sequence to obtain $d_\mathcal{G}(s)$. This creates two collections of mean distances across all choices of $s$, which we convert into empirical distributions $D_{\mathcal{R}}$ and $D_{\mathcal{G}}$. We then apply Jensen-Shannon divergence to compare these distance distributions:
\begin{equation}
\text{DTW-JS}(D_{\mathcal{R}}, D_{\mathcal{G}}) = \frac{1}{2}[\text{KL}(D_{\mathcal{R}} || D_M) + \text{KL}(D_{\mathcal{G}} || D_M)]
\end{equation}
where $D_M = \frac{1}{2}(D_{\mathcal{R}} + D_{\mathcal{G}})$ is the mixture of the two distance distributions. Small DTW-JS values indicate that the real and generated samples are ``distributionally'' similar in terms of their temporal patterns. More details can be found in Appendix~\ref{appendix:metrics}.

\subsection{Short Sequence Time Series Unconditional Generation}
We follow the evaluation setup of TimeGAN~\citep{yoon2019time} and Diffusion-TS~\citep{yuan2024diffusionts} to assess generation quality against baseline models. All datasets are segmented into sequences of length $24$ using a sliding window with stride $1$. For evaluation, we generate samples matching the size of the original training data for each dataset to ensure fair evaluation. Training configurations and times, as well as model complexity, are discussed in Appendices~\ref{appendix:training_configuration} and~\ref{appendix:computation}.

As shown in Table~\ref{tab:short_generation_results}, our method consistently outperforms all diffusion-based baseline methods across most datasets and metrics. While FourierDiffusion achieves competitive performance on certain datasets, WaveletDiff demonstrates superior and more consistent results across all evaluation scenarios. In our evaluations, we also tested different wavelet families and selected Symlets wavelets for Stocks, Coiflets wavelets for fMRI, and Daubechies wavelets for other datasets. In general, even when universally adopting Daubechies wavelets, WaveletDiff outperforms other diffusion paradigms. More details regarding the influence of the wavelet basis function on generative performance are available in Appendix~\ref{appendix:wavelet_selection}. To highlight our model’s ability to capture real data distributions, we present t-SNE embeddings and probability density plots for all six datasets in Appendix~\ref{appendix:tsne_pdf_short}. 

When compared against MSDformer~\citep{feng2025msdformer}, a transformer-based model operating on VQ-VAE latent representations, results are more mixed: MSDformer performs better on ETTh1, ETTh2, and especially fMRI, while WaveletDiff performs better on Exchange Rate and especially EEG. To better understand the significant performance gaps of WaveletDiff and MSDformer on fMRI and EEG, we investigate the underlying dataset characteristics in the wavelet domain that distinguish their generative performance.

\begin{table}[htbp]
\centering
\caption{Time series generation performance comparison on short sequences (length $24$) with training time (on one NVIDIA A100 GPU) and model parameter comparison between WaveletDiff and MSDformer.}
\label{tab:short_generation_results}
\resizebox{0.9\textwidth}{!}{%
\begin{tabular}{c|c|c|c|c|c|c|c}
\hline
\textbf{Metric} & \textbf{Methods} & \textbf{ETTh1} & \textbf{ETTh2} & \textbf{Stocks} & \textbf{Exchange Rate} & \textbf{fMRI} & \textbf{EEG} \\
\hline
\multirow{9}{*}{\begin{tabular}[c]{@{}c@{}}Discriminative\\ Score \\ (Lower the Better)\end{tabular}}
& WaveletDiff & \underline{0.005±.005} & \underline{0.008±.007} & \textbf{0.005±.004} & \textbf{0.004±.001} & \underline{0.087±.077} & \textbf{0.006±.008} \\
& MSDformer & \textbf{0.004$\pm$.003} & \textbf{0.006$\pm$.004} & \underline{0.011$\pm$.009} & \underline{0.008$\pm$.006} & \textbf{0.008$\pm$.004} & 0.053$\pm$.042 \\
& FourierDiffusion & 0.019±.007 & 0.016±.006 & 0.024±.003 & 0.015±.009 & 0.196±.013 & 0.016±.007 \\
& Diffusion-TS & 0.071±.002 & 0.038±.008 & 0.087±.008 & 0.032±.002 & 0.188±.018 & 0.304±.177 \\
& TimeGAN & 0.127±.047 & 0.106±.035 & 0.091±.047 & 0.257±.070 & 0.499±.001 & 0.161±.063 \\
& SigDiffusions & 0.353±.023 & 0.381±.048 & 0.371±.027 & 0.324±.055 & 0.482±.018 & 0.500±.000 \\
& KoVAE & 0.189±.020 & 0.049±.015 & 0.040±.023 & 0.122±.010 & 0.479±.025 & 0.244±.090 \\
& TimesFM & 0.066±.017 & 0.024±.012 & 0.059±.015 & \underline{0.008±.006} & 0.385±.188 & \underline{0.009±.004} \\
& Chronos & 0.110±.017 & 0.046±.043 & 0.073±.034 & 0.010±.005 & 0.458±.079 & 0.049±.009 \\
\hline
\multirow{9}{*}{\begin{tabular}[c]{@{}c@{}}Predictive\\ Score \\ (Lower the Better)\end{tabular}}
& WaveletDiff & \textbf{0.119±.002} & \textbf{0.106±.004} & \textbf{0.037±.000} & \underline{0.037±.002} & \underline{0.100±.000} & \textbf{0.000±.000} \\
& MSDformer & 0.121$\pm$.000 & \textbf{0.106$\pm$.005} & \textbf{0.037$\pm$.000} & \textbf{0.036$\pm$.003} & \textbf{0.093$\pm$.002} & \textbf{0.000$\pm$.000} \\
& FourierDiffusion & \underline{0.120±.005} & 0.111±.003 & \textbf{0.037±.000} & 0.040±.001 & \underline{0.100±.000} & \textbf{0.000±.000} \\
& Diffusion-TS & \underline{0.120±.004} & \underline{0.107±.003} & \textbf{0.037±.000} & \underline{0.037±.002} & \underline{0.100±.000} & \underline{0.001±.000} \\
& TimeGAN & 0.152±.015 & 0.128±.005 & \underline{0.038±.000} & 0.064±.005 & 0.124±.002 & \textbf{0.000±.000} \\
& SigDiffusions & 0.131±.002 & 0.125±.003 & 0.040±.001 & 0.089±.006 & 0.105±.000 & \textbf{0.000±.000} \\
& KoVAE & 0.126±.001 & 0.112±.003 & \textbf{0.037±.000} & 0.040±.004 & 0.225±.019 & \textbf{0.000±.000} \\
& TimesFM & 0.121±.003 & 0.108±.003 & \textbf{0.037±.000} & 0.038±.001 & 0.103±.000 & \textbf{0.000±.000} \\
& Chronos & 0.121±.003 & \underline{0.107±.004} & \textbf{0.037±.000} & 0.038±.001 & 0.103±.000 & \textbf{0.000±.000} \\
\hline
\multirow{9}{*}{\begin{tabular}[c]{@{}c@{}}Context-FID\\ Score \\ (Lower the Better)\end{tabular}}
& WaveletDiff & \underline{0.020±.001} & \underline{0.023±.002} & \underline{0.018±.002} & \textbf{0.006±.000} & \underline{0.104±.006} & \textbf{0.006±.000} \\
& MSDformer & \textbf{0.003$\pm$.000} & \textbf{0.004$\pm$.000} & \textbf{0.002$\pm$.000} & \underline{0.007$\pm$.000} & \textbf{0.010$\pm$.000} & 0.017$\pm$.003 \\
& FourierDiffusion & 0.031±.002 & 0.024±.003 & 0.093±.010 & 0.054±.013 & 0.169±.005 & \underline{0.012±.001} \\
& Diffusion-TS & 0.151±.007 & 0.054±.002 & 0.187±.016 & 0.056±.007 & 0.106±.003 & 0.017±.001 \\
& TimeGAN & 0.661±.041 & 0.157±.011 & 0.110±.012 & 0.660±.042 & 1.404±.114 & 0.018±.001 \\
& SigDiffusions & 2.413±.179 & 1.053±.099 & 3.494±.383 & 1.691±.157 & 6.576±.210 & 0.022±.001 \\
& KoVAE & 1.108±.079 & 0.211±.016 & 0.067±.005 & 0.177±.013 & 1.608±.050 & 0.018±.004 \\
& TimesFM & 0.389±.035 & 0.080±.013 & 0.149±.024 & 0.014±.000 & 12.262±.126 & 0.016±.002 \\
& Chronos & 1.080±.398 & 0.431±.129 & 0.325±.039 & 0.948±.297 & 8.910±.384 & 0.014±.001 \\
\hline
\multirow{9}{*}{\begin{tabular}[c]{@{}c@{}}Correlational\\ Score \\ (Lower the Better)\end{tabular}}
& WaveletDiff & \underline{0.043±.008} & \underline{0.083±.016} & \underline{0.005±.003} & \textbf{0.060±.020} & \underline{1.177±.031} & \textbf{1.811±.963} \\
& MSDformer & \textbf{0.038$\pm$.005} & \textbf{0.079$\pm$.016} & \textbf{0.004$\pm$.003} & \underline{0.066$\pm$.012} & \textbf{0.764$\pm$.049} & 4.058$\pm$.258 \\
& FourierDiffusion & 0.046±.009 & 0.095±.016 & 0.013±.003 & 0.072±.019 & 1.184±.023 & 3.544±.626 \\
& Diffusion-TS & 0.051±.007 & 0.089±.022 & 0.009±.007 & 0.115±.016 & 1.382±.036 & 4.764±.107 \\
& TimeGAN & 0.202±.010 & 0.185±.015 & 0.053±.003 & 0.416±.018 & 29.562±.067 & 8.820±.121 \\
& SigDiffusions & 0.210±.010 & 0.430±.025 & 0.070±.005 & 0.943±.024 & 15.389±.064 & 4.389±.257 \\
& KoVAE & 0.163±.019 & 0.243±.049 & 0.061±.004 & 0.150±.026 & 7.394±.033 & 6.918±.157 \\
& TimesFM & 0.102±.018 & 0.094±.016 & 0.028±.007 & 0.075±.016 & 6.450±.127 & 3.633±.495 \\
& Chronos & 0.118±.028 & 0.131±.016 & 0.121±.019 & 0.185±.056 & 4.873±.043 & \underline{3.039±.468} \\
\hline
\multirow{9}{*}{\begin{tabular}[c]{@{}c@{}}DTW-JS distance \\ (Lower the Better)\end{tabular}}
& WaveletDiff & \underline{0.101±.016} & \textbf{0.064±.014} & \underline{0.106±.027} & \textbf{0.121±.029} & \underline{0.191±.011} & \textbf{0.055±.011} \\
& MSDformer & \textbf{0.094$\pm$.023} & 0.077$\pm$.010 & \textbf{0.105$\pm$.008} & 0.127$\pm$.021 & \textbf{0.115$\pm$.024} & 0.605$\pm$.062 \\
& FourierDiffusion & 0.105±.022 & \underline{0.073±.014} & 0.138±.024 & 0.130±.021 & 0.286±.042 & \underline{0.067±.017} \\
& Diffusion-TS & 0.111±.020 & 0.087±.012 & 0.153±.019 & 0.139±.031 & 0.237±.042 & 0.220±.013 \\
& TimeGAN & 0.155±.030 & 0.097±.042 & 0.142±.028 & 0.231±.025 & 0.215±.037 & 0.632±.049 \\
& SigDiffusions & 0.259±.024 & 0.273±.034 & 0.377±.068 & 0.376±.036 & 0.693±.000 & 0.293±.125 \\
& KoVAE & 0.104±.018 & 0.075±.016 & 0.122±.025 & 0.155±.016 & 0.693±.000 & 0.595±.063 \\
& TimesFM & 0.148±.025 & 0.082±.013 & 0.118±.006 & 0.129±.030 & 0.693±.000 & 0.144±.021 \\
& Chronos & 0.127±.035 & 0.077±.018 & 0.115±.017 & \underline{0.123±.026} & 0.693±.000 & 0.086±.018 \\
\hline
\hline
\multirow{2}{*}{Training Time (h:m:s)} & WaveletDiff & \textbf{9:11:06} & \textbf{9:00:45} & \textbf{4:19:59} & \textbf{5:44:06} & \textbf{7:52:16} & \textbf{8:22:59} \\
& MSDformer & 9:36:02 & 9:27:33 & 5:23:36 & 11:47:29 & 25:53:28 & 9:48:10 \\
\hline
\multirow{2}{*}{Model Parameter (M)} & WaveletDiff & \textbf{63.1} & \textbf{63.1} & 63.1 & \textbf{63.1} & \textbf{63.2} & \textbf{63.1} \\
& MSDformer & 72 & 72 & \textbf{55.9} & 72 & 83.9 & 72 \\
\hline
\end{tabular}%
}
\end{table}

\begin{figure}[h]
    \centering
    \includegraphics[width=\linewidth]{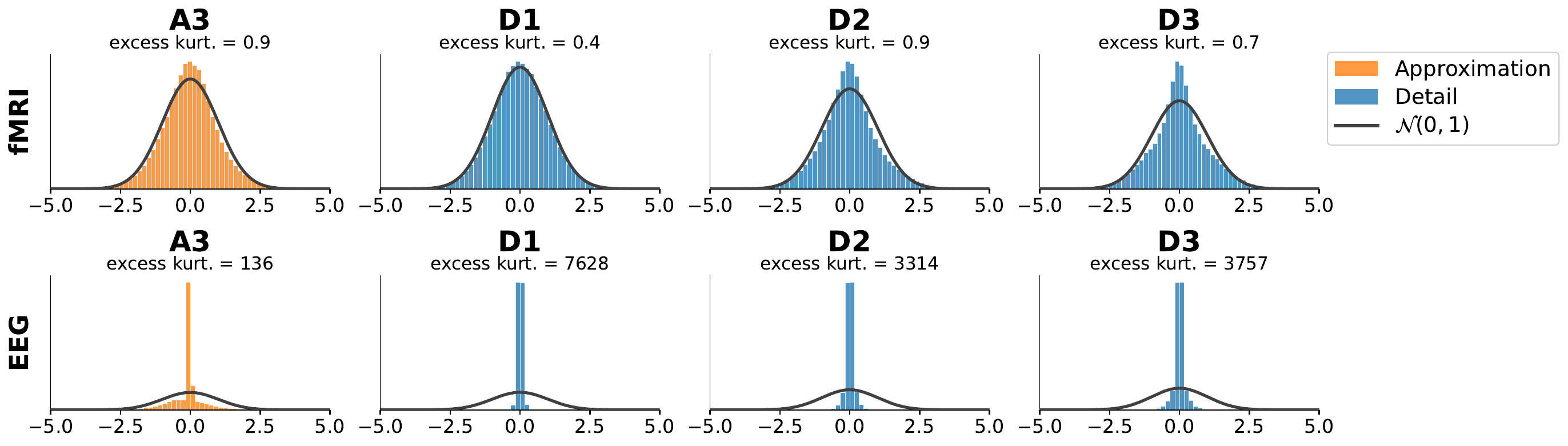}
    \caption{Standardized wavelet coefficient density of fMRI and EEG datasets. fMRI demonstrates near-Gaussian distributions while EEG exhibits highly non-Gaussian distributions. Energy distributions for coefficient at different levels are depicted in Figure~\ref{fig:energy_fraction_by_band} in the Appendix.}
    \label{fig:coefficient_distributions}
\end{figure}

\vspace{-2mm}
\paragraph{For Which Time Series Does Wavelet-Domain Diffusion Work the Best?} Note that WaveletDiff outperforms pure time and frequency domain diffusion models, and consequently, the issues to be discussed in what follows similarly pertain to other diffusion models that operate in transform domains. Among the six datasets, fMRI and EEG represent two extremes in terms of the kurtosis of their wavelet coefficients across decomposition levels, shown in Figure~\ref{fig:coefficient_distributions}. EEG exhibits highly non-Gaussian coefficient distributions, whereas the wavelet coefficients of fMRI are very close to Gaussian. This observation is consistent with prior findings reported in the literature~\citep{BULLMORE2004S234, maxim2005fractional} that pointed out that fMRI signals approximately follow a $1/f$ process, resulting in near-Gaussian wavelet coefficients for each level. The latter also explains why WaveletDiff performs relatively poorly on fMRI compared to MSDformer: The near-Gaussian coefficient distributions contain less higher-order statistical structure for our transformer to exploit. If levels were exactly Gaussian, the optimal denoisers would reduce to a $t$-dependent affine map which has a closed form. It would render our level-specific transformer design redundant. Since the coefficients are only near-Gaussian, residual patterns remain for the model to learn, which may explain why WaveletDiff still outperforms the other diffusion baselines on fMRI even though it underperforms MSDformer. On the other hand, for EEG signals, the wavelet coefficients exhibit distributions that exhibit more structure and are individually easier to learn, which makes them highly amenable for wavelet-domain diffusion.

To verify these findings even further, we added results on EEG-related time series recordings that exhibit similar wavelet-domain structure: Intracranial EEG (iEEG)~\citep{nejedly2020multicenter} and scalp-EEG sleep recordings from Sleep-EDF~\citep{kemp2000sleepedf, goldberger2000physionet}. As may be seen from  Table~\ref{tab:ieeg_sleepedf} in the Appendix, WaveletDiff outperforms MSDformer on these types of series as expected. A further ablation study in Table~\ref{tab:ablation} based on replacing level-specific transformers with one single shared transformer further supports our conclusions. Interestingly, the generative performance on fMRI improves on multiple metrics with a shared transformer, whereas the performance of other datasets generally degrades. This supports our hypothesis that level-specific modeling is less beneficial for near-Gaussian wavelet coefficients but clearly more effective for complex coefficient distributions.

\vspace{-2mm}
\paragraph{A Closer Look at the Context-FID Metric.} 
Although WaveletDiff performs comparably to MSDformer on almost all datasets tested except fMRI and EEG, it shows a pronounced gap with respect to Context-FID metric. This can be attributed to the fundamental architectural difference between the two generation paradigms. WaveletDiff is a diffusion model that directly learns the mapping from a Gaussian prior to the data distribution, whereas MSDformer generates samples using a discrete-token transformer operating on the latent manifold learned by a VQ-VAE. Since Context-FID is computed as the FID between TS2Vec embeddings of real and generated samples, MSDformer benefits from generation within a learned latent manifold, naturally leading to lower Context-FID scores. In contrast, diffusion models generate samples directly from Gaussian noise without such manifold constraints. Therefore, Context-FID may not be fully adequate for comparing these fundamentally different generation paradigms. This interpretation is further supported by the fact that, although WaveletDiff underperforms MSDformer on Context-FID, it consistently outperforms \emph{all other diffusion-based models on the same metric.} It also suggests a new direction for further improving WaveletDiff through specialized vector quantization methods and manifold constraints.



\subsection{Long Sequence Time Series Unconditional Generation}
We assess the performance of WaveletDiff for long time series generation by segmenting datasets into sequences of length $32$, $64$, and $128$, again using a sliding window with stride $1$. The same evaluation protocol is applied, where we generate samples matching the size of the original training data for each dataset. As seen in Table~\ref{tab:long_generation_results}, WaveletDiff offers consistent performance improvements across most settings. Here we used the spectral energy preservation term based on Parseval's theorem with a loss weight $\lambda_\text{energy}=0.3$ on ETTh1 and Exchange Rate, but not on Stocks due to its high volatility.

\begin{table}[ht]
\centering
\caption{Time series generation performance comparison on long sequences.}
\label{tab:long_generation_results}
\resizebox{0.9\textwidth}{!}{%
\begin{tabular}{c|c|c|c|c|c|c|c}
\hline
\textbf{Dataset} & \textbf{Metric} & \textbf{Length} & \textbf{WaveletDiff} & \textbf{FourierDiffusion} & \textbf{Diffusion-TS} & \textbf{TimeGAN} & \textbf{SigDiffusions} \\
\hline
\multirow{15}{*}{ETTh1}
& \multirow{3}{*}{\begin{tabular}[c]{@{}c@{}}Discriminative\\ Score\\ (Lower the Better)\end{tabular}}
& 32 & \textbf{0.016±.001} & 0.030±.004 & 0.078±.003 & 0.128±.036 & 0.346±.033 \\
& & 64 & \textbf{0.028±.009} & 0.048±.004 & 0.079±.010 & 0.116±.088 & 0.294±.156 \\
& & 128 & \textbf{0.034±.037} & 0.113±.006 & 0.159±.006 & 0.299±.148 & 0.462±.035 \\
\cline{2-8}
& \multirow{3}{*}{\begin{tabular}[c]{@{}c@{}}Predictive\\ Score\\ (Lower the Better)\end{tabular}}
& 32 & \textbf{0.119±.001} & \textbf{0.119±.005} & \textbf{0.119±.003} & 0.126±.009 & 0.129±.000 \\
& & 64 & \textbf{0.114±.007} & \textbf{0.114±.004} & 0.120±.004 & 0.125±.004 & 0.129±.002 \\
& & 128 & 0.113±.005 & \textbf{0.112±.007} & 0.116±.005 & 0.177±.015 & 0.129±.002 \\
\cline{2-8}
& \multirow{3}{*}{\begin{tabular}[c]{@{}c@{}}Context-FID\\ Score\\ (Lower the Better)\end{tabular}} 
& 32 & \textbf{0.038±.005} & 0.048±.003 & 0.204±.011 & 0.599±.044 & 2.875±.027 \\
& & 64 & \textbf{0.088±.005} & 0.135±.010 & 0.265±.012 & 0.978±.114 & 6.622±.354 \\
& & 128 & \textbf{0.256±.014} & 0.356±.021 & 0.805±.094 & 11.813±.851 & 11.596±.800 \\
\cline{2-8}
& \multirow{3}{*}{\begin{tabular}[c]{@{}c@{}}Correlational\\ Score\\ (Lower the Better)\end{tabular}} 
& 32 & \textbf{0.050±.004} & 0.056±.019 & 0.064±.014 & 0.118±.013 & 0.180±.013 \\
& & 64 & 0.054±.009 & \textbf{0.052±.006} & 0.059±.010 & 0.307±.015 & 0.200±.023 \\
& & 128 & \textbf{0.059±.021} & 0.072±.010 & 0.083±.004 & 1.098±.005 & 0.235±.015 \\
\cline{2-8}
& \multirow{3}{*}{\begin{tabular}[c]{@{}c@{}}DTW-JS\\ Distance\\ (Lower the Better)\end{tabular}}
& 32 & \textbf{0.095±.022} & 0.099±.035 & 0.113±.022 & 0.226±.019 & 0.235±.017 \\
& & 64 & \textbf{0.095±.028} & 0.105±.031 & 0.123±.034 & 0.208±.017 & 0.199±.031 \\
& & 128 & \textbf{0.105±.035} & 0.134±.022 & 0.122±.017 & 0.262±.051 & 0.122±.021 \\
\hline

\multirow{15}{*}{Stocks}
& \multirow{3}{*}{\begin{tabular}[c]{@{}c@{}}Discriminative\\ Score\\ (Lower the Better)\end{tabular}}
& 32 & \textbf{0.006±.004} & 0.022±.012 & 0.099±.012 & 0.197±.025 & 0.357±.027 \\
& & 64 & \textbf{0.007±.003} & 0.032±.018 & 0.099±.008 & 0.152±.020 & 0.324±.044 \\
& & 128 & \textbf{0.015±.008} & 0.086±.036 & 0.141±.011 & 0.270±.124 & 0.339±.007 \\
\cline{2-8}
& \multirow{3}{*}{\begin{tabular}[c]{@{}c@{}}Predictive\\ Score\\ (Lower the Better)\end{tabular}}
& 32 & \textbf{0.037±.000} & \textbf{0.037±.000} & 0.038±.000 & \textbf{0.037±.000} & 0.040±.001 \\
& & 64 & \textbf{0.036±.000} & \textbf{0.036±.000} & 0.037±.000 & 0.038±.000 & 0.039±.000 \\
& & 128 & \textbf{0.036±.000} & 0.038±.000 & 0.037±.000 & 0.070±.007 & 0.040±.000 \\
\cline{2-8}
& \multirow{3}{*}{\begin{tabular}[c]{@{}c@{}}Context-FID\\ Score\\ (Lower the Better)\end{tabular}} 
& 32 & \textbf{0.026±.006} & 0.087±.007 & 0.256±.029 & 0.449±.042 & 3.403±.373 \\
& & 64 & \textbf{0.047±.005} & 0.151±.026 & 0.369±.065 & 0.336±.046 & 4.229±.495 \\
& & 128 & \textbf{0.080±.012} & 0.379±.025 & 0.417±.077 & 3.231±.325 & 5.472±.004 \\
\cline{2-8}
& \multirow{3}{*}{\begin{tabular}[c]{@{}c@{}}Correlational\\ Score\\ (Lower the Better)\end{tabular}} 
& 32 & \textbf{0.002±.002} & 0.011±.001 & 0.017±.007 & 0.094±.006 & 0.075±.004 \\
& & 64 & \textbf{0.003±.001} & 0.013±.005 & 0.020±.002 & 0.098±.003 & 0.052±.004 \\
& & 128 & \textbf{0.004±.002} & 0.162±.011 & 0.021±.006 & 0.621±.006 & 0.091±.004 \\
\cline{2-8}
& \multirow{3}{*}{\begin{tabular}[c]{@{}c@{}}DTW-JS\\ Distance\\ (Lower the Better)\end{tabular}}
& 32 & \textbf{0.112±.025} & 0.118±.021 & 0.137±.026 & 0.182±.026 & 0.301±.060 \\
& & 64 & \textbf{0.136±.021} & 0.139±.008 & \textbf{0.136±.018} & 0.155±.031 & 0.261±.013 \\
& & 128 & \textbf{0.112±.013} & 0.127±.020 & 0.116±.004 & 0.420±.015 & 0.281±.058 \\
\hline

\multirow{15}{*}{Exchange Rate}
& \multirow{3}{*}{\begin{tabular}[c]{@{}c@{}}Discriminative\\ Score\\ (Lower the Better)\end{tabular}}
& 32 & \textbf{0.011±.005} & 0.018±.013 & 0.031±.006 & 0.254±.064 & 0.314±.024 \\
& & 64 & \textbf{0.020±.005} & 0.038±.015 & 0.028±.005 & 0.277±.046 & 0.300±.007 \\
& & 128 & \textbf{0.026±.008} & 0.092±.032 & 0.046±.007 & 0.106±.064 & 0.276±.015 \\
\cline{2-8}
& \multirow{3}{*}{\begin{tabular}[c]{@{}c@{}}Predictive\\ Score\\ (Lower the Better)\end{tabular}}
& 32 & \textbf{0.035±.002} & 0.040±.002 & 0.036±.002 & 0.069±.006 & 0.085±.007 \\
& & 64 & \textbf{0.035±.001} & 0.041±.001 & \textbf{0.035±.002} & 0.056±.005 & 0.078±.008 \\
& & 128 & \textbf{0.034±.003} & 0.044±.002 & \textbf{0.034±.002} & 0.048±.003 & 0.074±.005 \\
\cline{2-8}
& \multirow{3}{*}{\begin{tabular}[c]{@{}c@{}}Context-FID\\ Score\\ (Lower the Better)\end{tabular}} 
& 32 & \textbf{0.013±.001} & 4.057±.648 & 0.037±.003 & 1.038±.144 & 1.853±.164 \\
& & 64 & \textbf{0.022±.003} & 0.129±.065 & 0.056±.005 & 1.136±.114 & 1.834±.235 \\
& & 128 & \textbf{0.052±.003} & 0.264±.008 & 0.063±.004 & 0.849±.087 & 2.079±.168 \\
\cline{2-8}
& \multirow{3}{*}{\begin{tabular}[c]{@{}c@{}}Correlational\\ Score\\ (Lower the Better)\end{tabular}} 
& 32 & \textbf{0.064±.010} & 0.109±.037 & 0.091±.044 & 0.456±.016 & 1.065±.033 \\
& & 64 & \textbf{0.066±.024} & 0.096±.026 & 0.097±.012 & 0.421±.038 & 1.042±.027 \\
& & 128 & \textbf{0.065±.026} & 0.173±.011 & 0.101±.019 & 0.237±.035 & 1.001±.046 \\
\cline{2-8}
& \multirow{3}{*}{\begin{tabular}[c]{@{}c@{}}DTW-JS\\ Distance\\ (Lower the Better)\end{tabular}}
& 32 & \textbf{0.108±.037} & 0.116±.028 & 0.129±.023 & 0.182±.023 & 0.375±.045 \\
& & 64 & \textbf{0.132±.015} & 0.142±.028 & 0.136±.010 & 0.195±.036 & 0.305±.024 \\
& & 128 & \textbf{0.124±.018} & 0.146±.029 & 0.145±.028 & 0.224±.020 & 0.306±.022 \\
\hline
\end{tabular}%
}
\end{table}

\subsection{Reproducibility and Robustness to Irregular Sampling}
Diffusion model analysis has been extended in many directions, including conditional generation, ambient diffusion~\citep{daras2023ambient}, reproducibility and others. Many of these analyses have also been adapted to time series generation. In this work, we present the first study of reproducibility for time series diffusion models. Following recent diffusion model reproducibility studies~\citep{pmlr-v235-zhang24cn, li2024understanding, kadkhodaie2024generalization}, we investigate whether this phenomenon extends to the time series domain. Specifically, we train pairs of models with architectural variations and generate samples from identical Gaussian noise using deterministic DDIM sampling. Our results show that time series diffusion models exhibit strong reproducibility across all representation domains (see Appendix~\ref{appendix:reproducibility}). We further extend WaveletDiff to irregular sampled time series by following the preprocessing protocol of KoVAE and GT-GAN, where missing observations are first imputed using cubic spline interpolation before training. Experimental results demonstrate that WaveletDiff maintains consistently strong performance across different missing rates, indicating that the proposed wavelet-domain diffusion framework is robust to irregular sampling (see Appendix~\ref{appendix:irregular}).

\subsection{Ablation Study}
\paragraph{Wavelet selection.}
We analyzed the performance of the methods with respect to different wavelet design choices, including the wavelet basis, order, and level. We examined five different wavelet families and three different settings for the wavelet order and decomposition level. The ablation studies pertaining to the order and level  are performed exclusively on Daubechies wavelets for all datasets, in order to ensure consistent comparisons. As shown in Table~\ref{tab:wavelet} in the Appendix, across the five wavelet families, performance varies with the type of datasets but consistently outperforms all baselines. We believe this to be the case because different wavelet families possess distinct properties, such as orthogonality, compact support, and smoothness, which directly shape coefficient sparsity and localization in the wavelet domain, making them better suited to different dataset characteristics. This ability to tailor the wavelet basis to the characteristics of a dataset represents a key advantage of WaveletDiff over frequency domain approaches which rely on fixed Fourier basis functions regardless of the underlying signal properties. More details can be found in Appendix~\ref{appendix:wavelet_selection}. In addition, as shown in Table~\ref{tab:ablation_order_level} in the Appendix, different choices of wavelet order and decomposition level lead to minor performance variations, while all settings consistently outperform baselines.

\paragraph{WaveletDiff architectural components.}
To validate the effectiveness of different architectural components of WaveletDiff, we conduct an ablation study comparing our full model against six variants: (1) \emph{Predicting coefficients rather than noise:} Instead of predicting noise as in standard DDPM, this variant directly predicts the wavelet coefficients themselves, following the approach in Diffusion-TS~\citep{yuan2024diffusionts} which suggests this method outperforms noise prediction. (2) \emph{Removing cross-attention:} This variant disables information exchange between different wavelet decomposition levels, but maintains level-specific transformer models. (3) \emph{Using Cosine instead of Exponential schedules:} This variant replaces exponential with cosine noise scheduling during diffusion. (4) \emph{Using DDIM sampling:} This variant uses deterministic DDIM rather than DDPM sampling during inference. (5) \emph{Shared transformer:} Replacing the level-transformers to a single shared transformer. (6) \emph{w/o larger approx-level capacity:} Reducing the number of layers and dimensions of the approximate level transformer to be the same as detail level transformer.
The results in Table~\ref{tab:ablation} in the Appendix reveal that \emph{cross-level attention} is the universally most critical architectural component, with its removal causing discriminative scores and Context-FID scores to degrade on average by approximately $4\times$ and $3.5\times$, respectively. While cosine noise scheduling and DDIM sampling show competitive performance on certain datasets, they exhibit instability in neuroscience domain datasets fMRI and EEG.
Additionally, the individual level-transformers structure and larger approximate level transformer also demonstrates critical improvement in most settings.

\paragraph{Energy loss weight}
As pointed out in Section~\ref{subsec:diffusion_framework}, the base reconstruction loss is typically sufficient for short sequence generation since the spectral energy drift is minimal over limited temporal horizons. We only adopted energy-preserving regularization for long sequence generation on ETTh1 and Exchange Rate datasets. We found that setting the weight to $0.3$ consistently offered good performance across different settings. As shown in Table~\ref{tab:ablation_energy_weight} in the Appendix, the energy regularization term improves performance across most metrics, except for Exchange Rate with generation sequence length equal to $32$, where the short-horizon, low-variance dynamics lead to negligible spectral energy drift, limiting the benefits of regularization.

\subsection{Computational Analysis}
WaveletDiff comprises approximately 63M trainable parameters and the training times across different datasets and sequence lengths are presented in Table~\ref{tab:WaveletDiff_training_times} in the Appendix. Our model has larger computational times compared to diffusion-based baselines; however, for many time series generation applications, one most often prioritizes overall output quality over generation time. Also, note that generative models like ours are typically used in offline settings (e.g., augmentation, simulation), in which case latency is less critical than for forecasting. Also, time series are “less complex” than images or videos so that training completes within hours, while sampling thousands of sequences takes seconds. Given that generation quality is the primary bottleneck, we believe that the increased cost is justified by the substantial gains in fidelity (for details, see Appendix~\ref{appendix:computation}). Furthermore, as shown in Table~\ref{tab:short_generation_results}, WaveletDiff actually has a slightly smaller model size and shorter training time compared to the Transformer-based baseline MSDformer.



\subsubsection*{Broader Impact Statement}
While WaveletDiff demonstrates strong empirical performance across diverse datasets and metrics, synthetic time series generation remains an imperfect process, and generated samples may not fully reproduce all characteristics of real-world data. Users should be cautious when deploying WaveletDiff-generated data on high-stakes downstream tasks such as clinical decision-making, financial risk assessment, or safety-critical engineering applications, where even small distributional mismatches between synthetic and real data can lead to unreliable and potentially dangerous outcomes. We encourage practitioners to evaluate generated data using domain-specific quality criteria in addition to the metrics considered in this work, and to view synthetic data as a complement to, rather than a substitute for, real-world data collection whenever feasible.



\bibliography{main}

@ARTICLE{lee2019dynamic,
  author={Lee, Changhee and Yoon, Jinsung and Schaar, Mihaela van der},
  journal={IEEE Transactions on Biomedical Engineering}, 
  title={Dynamic-DeepHit: A Deep Learning Approach for Dynamic Survival Analysis With Competing Risks Based on Longitudinal Data}, 
  year={2020},
  volume={67},
  number={1},
  pages={122-133},
  doi={10.1109/TBME.2019.2909027}}

@inproceedings{taga2025timepfn,
  title={TimePFN: Effective multivariate time series forecasting with synthetic data},
  author={Taga, Ege Onur and Ildiz, Muhammed Emrullah and Oymak, Samet},
  booktitle={Proceedings of the AAAI Conference on Artificial Intelligence},
  volume={39},
  number={19},
  pages={20761--20769},
  year={2025}
}

@ARTICLE{mitra2025music,
  author={Mitra, Rohan and Zualkernan, Imran},
  journal={IEEE Access}, 
  title={Music Generation Using Deep Learning and Generative AI: A Systematic Review}, 
  year={2025},
  volume={13},
  number={},
  pages={18079-18106},
  doi={10.1109/ACCESS.2025.3531798}}

@misc{vanderSchaar2019survival,
    title={Time series in healthcare: challenges and solutions},
    author={{van der Schaar Lab}},
    year={2019},
    howpublished={\url{https://www.vanderschaar-lab.com/time-series-in-healthcare/}},
    note={Accessed: April 12, 2023}
}

@misc{sezer2020financial,
      title={Financial Time Series Forecasting with Deep Learning : A Systematic Literature Review: 2005-2019}, 
      author={Omer Berat Sezer and Mehmet Ugur Gudelek and Ahmet Murat Ozbayoglu},
      year={2019},
      eprint={1911.13288},
      archivePrefix={arXiv},
      primaryClass={cs.LG},
      url={https://arxiv.org/abs/1911.13288}, 
}

@misc{ozbayoglu2020deep,
      title={Deep Learning for Financial Applications : A Survey}, 
      author={Ahmet Murat Ozbayoglu and Mehmet Ugur Gudelek and Omer Berat Sezer},
      year={2020},
      eprint={2002.05786},
      archivePrefix={arXiv},
      primaryClass={q-fin.ST},
      url={https://arxiv.org/abs/2002.05786}, 
}

@article{dinku2019challenges,
    title={Challenges with availability and quality of climate data in Africa},
    author={Dinku, Tufa},
    journal={Climate Services},
    volume={14},
    pages={6--15},
    year={2019},
    publisher={Elsevier}
}

@Article{susto2020predictive,
AUTHOR = {Çınar, Zeki Murat and Abdussalam Nuhu, Abubakar and Zeeshan, Qasim and Korhan, Orhan and Asmael, Mohammed and Safaei, Babak},
TITLE = {Machine Learning in Predictive Maintenance towards Sustainable Smart Manufacturing in Industry 4.0},
JOURNAL = {Sustainability},
VOLUME = {12},
YEAR = {2020},
NUMBER = {19},
ARTICLE-NUMBER = {8211},
URL = {https://www.mdpi.com/2071-1050/12/19/8211},
ISSN = {2071-1050},
DOI = {10.3390/su12198211}
}

@article{lei2020survey,
    title={A survey of predictive maintenance: Systems, purposes and approaches},
    author={Lei, Yaguo and Yang, Bo and Jiang, Xinwei and Jia, Feng and Li, Naipeng and Nandi, Asoke K},
    journal={IEEE Access},
    volume={8},
    pages={74305--74334},
    year={2020},
    publisher={IEEE}
}

@article{wang2021deep,
    title={Deep time series models for scarce data},
    author={Wang, Qiyao and Farahat, Ahmed and Gupta, Chetan and Zheng, Sheng},
    journal={Neurocomputing},
    volume={456},
    pages={504--518},
    year={2021},
    publisher={Elsevier}
}

@article{desai2025timeseries,
    title={Time series generation under data scarcity: A unified generative modeling approach},
    author={Desai, Abhyuday and Freeman, Cynthia and Wang, Zuhui and Beaver, Ian},
    journal={arXiv preprint arXiv:2505.12850},
    year={2025}
}

@article{wen2021time,
    title={Time series data augmentation for deep learning: A survey},
    author={Wen, Qingsong and Sun, Liang and Yang, Fan and Song, Xiuqiang and Gao, Jingkun and Wang, Xue and Xu, Huan},
    journal={Proceedings of the International Joint Conference on Artificial Intelligence},
    pages={4653--4660},
    year={2021}
}

@article{forestier2017generating,
    title={Data augmentation using synthetic data for time series classification with deep residual networks},
    author={Le Guennec, Arthur and Malinowski, Simon and Tavenard, Romain},
    journal={arXiv preprint arXiv:1808.02455},
    year={2018}
}

@article{ryu2023simpsi,
    title={{SimPSI}: A simple strategy to preserve spectral information in time series data augmentation},
    author={Ryu, Hyun and others},
    journal={arXiv preprint arXiv:2312.05790},
    year={2023}
}

@article{wang2020partgan,
    title={{Part-GAN}: Privacy-preserving time-series sharing},
    author={Wang, Shuo and Rudolph, Carsten and Nepal, Surya and Grobler, Marthie and Chen, Shangqi},
    booktitle={International Conference on Artificial Neural Networks},
    pages={578--593},
    year={2020},
    organization={Springer}
}

@article{jordon2022synthetic,
    title={Synthetic data for privacy-preserving clinical risk prediction},
    author={Jordon, James and Szpruch, Lukasz and Houssiau, Florimond and Bottarelli, Mirko and Cherubin, Giovanni and Maple, Carsten and Cohen, Samuel N and Weller, Adrian},
    journal={Scientific Reports},
    volume={14},
    number={1},
    pages={24119},
    year={2024},
    publisher={Nature Publishing Group}
}

@article{nosowsky2006hipaa,
    title={The {Health Insurance Portability and Accountability Act} of 1996 ({HIPAA}) privacy rule: implications for clinical research},
    author={Nosowsky, Rachel and Giordano, Thomas J},
    journal={Annual review of medicine},
    volume={57},
    pages={575--590},
    year={2006},
    publisher={Annual Reviews}
}

@article{nikolenko2021synthetic,
    title={Synthetic data generation for machine learning},
    author={Nikolenko, Sergey I},
    journal={arXiv preprint arXiv:1909.11512},
    year={2021}
}

@article{elEmam2020synthetic,
    title={Synthetic data generation for realtime data pipelines},
    author={{El Emam}, Khaled and {Mosquera}, Lucy and {Hoptroff}, Richard},
    journal={Online Scientific Research},
    year={2022}
}

@inproceedings{yoon2019time,
 author = {Yoon, Jinsung and Jarrett, Daniel and van der Schaar, Mihaela},
 booktitle = {Advances in Neural Information Processing Systems},
 pages = {},
 publisher = {Curran Associates, Inc.},
 title = {Time-series Generative Adversarial Networks},
 url = {https://proceedings.neurips.cc/paper_files/paper/2019/file/c9efe5f26cd17ba6216bbe2a7d26d490-Paper.pdf},
 volume = {32},
 year = {2019}
}

@inproceedings{haoyietal-informer-2021,
  author    = {Haoyi Zhou and
               Shanghang Zhang and
               Jieqi Peng and
               Shuai Zhang and
               Jianxin Li and
               Hui Xiong and
               Wancai Zhang},
  title     = {Informer: Beyond Efficient Transformer for Long Sequence Time-Series Forecasting},
  booktitle = {The Thirty-Fifth {AAAI} Conference on Artificial Intelligence, {AAAI} 2021, Virtual Conference},
  volume    = {35},
  number    = {12},
  pages     = {11106--11115},
  publisher = {{AAAI} Press},
  year      = {2021},
}

@misc{lai2018modelinglongshorttermtemporal,
      title={Modeling Long- and Short-Term Temporal Patterns with Deep Neural Networks}, 
      author={Guokun Lai and Wei-Cheng Chang and Yiming Yang and Hanxiao Liu},
      year={2018},
      eprint={1703.07015},
      archivePrefix={arXiv},
      primaryClass={cs.LG},
      url={https://arxiv.org/abs/1703.07015}, 
}

@inproceedings{yuan2024diffusionts,
  title={Diffusion-{TS}: Interpretable Diffusion for General Time Series Generation},
  author={Xinyu Yuan and Yan Qiao},
  booktitle={The Twelfth International Conference on Learning Representations},
  year={2024},
  url={https://openreview.net/forum?id=4h1apFjO99}
}

@misc{yue2022ts2vecuniversalrepresentationtime,
      title={TS2Vec: Towards Universal Representation of Time Series}, 
      author={Zhihan Yue and Yujing Wang and Juanyong Duan and Tianmeng Yang and Congrui Huang and Yunhai Tong and Bixiong Xu},
      year={2022},
      eprint={2106.10466},
      archivePrefix={arXiv},
      primaryClass={cs.LG},
      url={https://arxiv.org/abs/2106.10466}, 
}

@article{ho2020denoising,
  title={Denoising Diffusion Probabilistic Models},
  author={Jonathan Ho and Ajay Jain and Pieter Abbeel},
  year={2020},
  journal={arXiv preprint arxiv:2006.11239}
}

@inproceedings{dhariwal2021diffusion,
 author = {Dhariwal, Prafulla and Nichol, Alexander},
 booktitle = {Advances in Neural Information Processing Systems},
 editor = {M. Ranzato and A. Beygelzimer and Y. Dauphin and P.S. Liang and J. Wortman Vaughan},
 pages = {8780--8794},
 publisher = {Curran Associates, Inc.},
 title = {Diffusion Models Beat GANs on Image Synthesis},
 url = {https://proceedings.neurips.cc/paper_files/paper/2021/file/49ad23d1ec9fa4bd8d77d02681df5cfa-Paper.pdf},
 volume = {34},
 year = {2021}
}

@inproceedings{kong2021diffwave,
title={DiffWave: A Versatile Diffusion Model for Audio Synthesis},
author={Zhifeng Kong and Wei Ping and Jiaji Huang and Kexin Zhao and Bryan Catanzaro},
booktitle={International Conference on Learning Representations},
year={2021},
url={https://openreview.net/forum?id=a-xFK8Ymz5J}
}

@article{austin2021structured,
  title={Structured denoising diffusion models in discrete state-spaces},
  author={Austin, Jacob and Johnson, Daniel D and Ho, Jonathan and Tarlow, Daniel and van den Berg, Rianne},
  journal={Advances in Neural Information Processing Systems},
  volume={34},
  pages={17981--17993},
  year={2021}
}

@misc{salinas2020deepar,
      title={DeepAR: Probabilistic Forecasting with Autoregressive Recurrent Networks}, 
      author={David Salinas and Valentin Flunkert and Jan Gasthaus},
      year={2019},
      eprint={1704.04110},
      archivePrefix={arXiv},
      primaryClass={cs.AI},
      url={https://arxiv.org/abs/1704.04110}, 
}

@inproceedings{crabbe2024time,
  title={Time series diffusion in the frequency domain},
  author={Crabb{\'e}, Jonathan and Assaad, Serge and Thiran, Jean-Philippe and Sj{\"o}lund, Jens},
  booktitle={Proceedings of the 41st International Conference on Machine Learning},
  pages={9491--9510},
  year={2024}
}

@ARTICLE{mallat1989theory,
  author={Mallat, S.G.},
  journal={IEEE Transactions on Pattern Analysis and Machine Intelligence}, 
  title={A theory for multiresolution signal decomposition: the wavelet representation}, 
  year={1989},
  volume={11},
  number={7},
  pages={674-693},
  doi={10.1109/34.192463}
}

@book{daubechies1992ten,
  title={Ten lectures on wavelets},
  author={Daubechies, Ingrid},
  volume={61},
  year={1992},
  publisher={SIAM}
}

@article{takahashi2024generation,
  title={Generation of synthetic financial time series by diffusion models},
  author={Takahashi, Tomonori and Mizuno, Takayuki},
  journal={arXiv preprint arXiv:2401.03692},
  year={2024}
}

@article{rasul2021autoregressive,
  title={Autoregressive denoising diffusion models for multivariate probabilistic time series forecasting},
  author={Rasul, Kashif and Seward, Calvin and Schuster, Ingmar and Vollgraf, Roland},
  journal={International Conference on Machine Learning},
  pages={8857--8868},
  year={2021}
}

@InProceedings{shen2023timediff,
  title = 	 {Non-autoregressive Conditional Diffusion Models for Time Series Prediction},
  author =       {Shen, Lifeng and Kwok, James},
  booktitle = 	 {Proceedings of the 40th International Conference on Machine Learning},
  pages = 	 {31016--31029},
  year = 	 {2023},
  editor = 	 {Krause, Andreas and Brunskill, Emma and Cho, Kyunghyun and Engelhardt, Barbara and Sabato, Sivan and Scarlett, Jonathan},
  volume = 	 {202},
  series = 	 {Proceedings of Machine Learning Research},
  month = 	 {23--29 Jul},
  publisher =    {PMLR},
  url = 	 {https://proceedings.mlr.press/v202/shen23d.html}
}

@article{li2022csdi,
  title={Conditional score-based diffusion models for probabilistic time series imputation},
  author={Li, Yusuke and Lu, Xinyuan and Wang, Yiping and Dou, Dejing},
  journal={Advances in Neural Information Processing Systems},
  volume={35},
  pages={24804--24816},
  year={2022}
}

@inproceedings{alaa2021generative,
title={Generative Time-series Modeling with Fourier Flows},
author={Ahmed Alaa and Alex James Chan and Mihaela van der Schaar},
booktitle={International Conference on Learning Representations},
year={2021},
url={https://openreview.net/forum?id=PpshD0AXfA}
}

@article{zhang2017financial,
  title={Financial time series forecasting using a hybrid approach based on wavelet packet denoising and neural networks},
  author={Zhang, Qinghua and Wan, Liansheng},
  journal={Applied Soft Computing},
  volume={54},
  pages={243--255},
  year={2017}
}

@book{addison2017wavelet,
  title={The illustrated wavelet transform handbook: introductory theory and applications in science, engineering, medicine and finance},
  author={Addison, Paul S},
  year={2017},
  publisher={CRC press}
}

@inproceedings{han2019twavenet,
  title={TWaveNet: A tree-structured wavelet neural network for time series signal analysis},
  author={Han, Zhichao and Zhao, Jun and Leung, Henry and Ma, Kai Fung and Wang, Wei},
  booktitle={2019 International Joint Conference on Neural Networks (IJCNN)},
  pages={1--8},
  year={2019},
  organization={IEEE}
}

@inproceedings{phung2022wavelet,
  title={Wavelet diffusion models are fast and scalable image generators},
  author={Phung, Hao and Dao, Quan and Tran, Anh},
  booktitle={Proceedings of the IEEE/CVF Conference on Computer Vision and Pattern Recognition},
  pages={10199--10208},
  year={2022}
}

@InProceedings{zhou2022fedformer,
  title = 	 {{FED}former: Frequency Enhanced Decomposed Transformer for Long-term Series Forecasting},
  author =       {Zhou, Tian and Ma, Ziqing and Wen, Qingsong and Wang, Xue and Sun, Liang and Jin, Rong},
  booktitle = 	 {Proceedings of the 39th International Conference on Machine Learning},
  pages = 	 {27268--27286},
  year = 	 {2022},
  editor = 	 {Chaudhuri, Kamalika and Jegelka, Stefanie and Song, Le and Szepesvari, Csaba and Niu, Gang and Sabato, Sivan},
  volume = 	 {162},
  series = 	 {Proceedings of Machine Learning Research},
  month = 	 {17--23 Jul},
  publisher =    {PMLR},
  url = 	 {https://proceedings.mlr.press/v162/zhou22g.html}
}

@inproceedings{xu2023fits,
  title={FITS: Modeling Time Series with 10k Parameters},
  author={Xu, Zhijian and Sharma, Atul and Zhang, Yicheng and Arik, Sercan O. and Pfister, Tomas},
  booktitle={International Conference on Learning Representations},
  year={2024}
}

@article{zhou2022film,
  title={FiLM: Frequency improved Legendre Memory Model for Long-term Time Series Forecasting},
  author={Zhou, Tian and Ma, Ziqing and Wang, Xue and Wen, Qingsong and Sun, Liang and Jin, Rong},
  journal={Advances in Neural Information Processing Systems},
  volume={35},
  pages={12677--12690},
  year={2022}
}

@article{zhang2024atfnet,
  title={ATFNet: Adaptive Time-Frequency Ensembled Network for Long-term Time Series Forecasting},
  author={Zhang, Hengyi and Wang, Zheng and Su, Weiwei},
  journal={arXiv preprint arXiv:2404.05192},
  year={2024}
}

@article{tashiro2021csdi,
  title={CSDI: Conditional score-based diffusion models for probabilistic time series imputation},
  author={Tashiro, Yusuke and Song, Jiaming and Song, Yang and Ermon, Stefano},
  journal={Advances in Neural Information Processing Systems},
  volume={34},
  pages={24804--24816},
  year={2021}
}

@inproceedings{wang2024spectral,
  title={Exploring time series analysis in frequency domain with complex-valued spectral attention and bidirectional variable mamba},
  author={Wang, Jinliang and Zhang, Weifeng and Li, Cheng and Liu, Yanfeng},
  journal={The Journal of Supercomputing},
  pages={1--25},
  year={2024},
  publisher={Springer}
}

@inproceedings{yi2023frets,
  title={Frequency-domain MLPs are more effective learners in time series forecasting},
  author={Yi, Kun and Zhang, Qi and Fan, Wei and Wang, Shoujin and Wang, Pengyang and He, Hui and An, Ning and Lian, Defu and Cao, Longbing and Niu, Zhendong},
  booktitle={Advances in Neural Information Processing Systems},
  volume={36},
  pages={76656--76671},
  year={2023}
}

@inproceedings{barancikova2025sigdiffusions,
   title={SigDiffusions: Score-Based Diffusion Models for Time Series via Log-Signature Embeddings},
   author={Barbora Barancikova and Zhuoyue Huang and Cristopher Salvi},
   booktitle={The Thirteenth International Conference on Learning Representations},
   year={2025},
   url={https://openreview.net/forum?id=Y8KK9kjgIK}
}

@misc{eskandarinasab2024seriesgantimeseriesgeneration,
      title={SeriesGAN: Time Series Generation via Adversarial and Autoregressive Learning}, 
      author={MohammadReza EskandariNasab and Shah Muhammad Hamdi and Soukaina Filali Boubrahimi},
      year={2024},
      eprint={2410.21203},
      archivePrefix={arXiv},
      primaryClass={cs.LG},
      url={https://arxiv.org/abs/2410.21203}, 
}

@misc{pei2021generatingrealworldtimeseries,
      title={Towards Generating Real-World Time Series Data}, 
      author={Hengzhi Pei and Kan Ren and Yuqing Yang and Chang Liu and Tao Qin and Dongsheng Li},
      year={2021},
      eprint={2111.08386},
      archivePrefix={arXiv},
      primaryClass={cs.LG},
      url={https://arxiv.org/abs/2111.08386}, 
}

@misc{chi2024rfdiffusionradiosignalgeneration,
      title={RF-Diffusion: Radio Signal Generation via Time-Frequency Diffusion}, 
      author={Guoxuan Chi and Zheng Yang and Chenshu Wu and Jingao Xu and Yuchong Gao and Yunhao Liu and Tony Xiao Han},
      year={2024},
      eprint={2404.09140},
      archivePrefix={arXiv},
      primaryClass={cs.LG},
      url={https://arxiv.org/abs/2404.09140}, 
}

@misc{tian2020tfgantimefrequencydomain,
      title={TFGAN: Time and Frequency Domain Based Generative Adversarial Network for High-fidelity Speech Synthesis}, 
      author={Qiao Tian and Yi Chen and Zewang Zhang and Heng Lu and Linghui Chen and Lei Xie and Shan Liu},
      year={2020},
      eprint={2011.12206},
      archivePrefix={arXiv},
      primaryClass={eess.AS},
      url={https://arxiv.org/abs/2011.12206}, 
}

@inproceedings{
huang2024generative,
title={Generative Learning for Financial Time Series with Irregular and Scale-Invariant Patterns},
author={Hongbin Huang and Minghua Chen and Xiao Qiao},
booktitle={The Twelfth International Conference on Learning Representations},
year={2024},
url={https://openreview.net/forum?id=CdjnzWsQax}
}

@misc{lim2023regulartimeseriesgenerationusing,
      title={Regular Time-series Generation using SGM}, 
      author={Haksoo Lim and Minjung Kim and Sewon Park and Noseong Park},
      year={2023},
      eprint={2301.08518},
      archivePrefix={arXiv},
      primaryClass={cs.LG},
      url={https://arxiv.org/abs/2301.08518}, 
}

@inproceedings{10.5555/3692070.3693584,
author = {Narasimhan, Sai Shankar and Agarwal, Shubhankar and Akcin, Oguzhan and Sanghavi, Sujay and Chinchali, Sandeep},
title = {Time Weaver: a conditional time series generation model},
year = {2024},
publisher = {JMLR.org},
booktitle = {Proceedings of the 41st International Conference on Machine Learning},
articleno = {1514},
numpages = {28},
location = {Vienna, Austria},
series = {ICML'24}
}

@misc{sikder2024transfusiongeneratinglonghigh,
      title={TransFusion: Generating Long, High Fidelity Time Series using Diffusion Models with Transformers}, 
      author={Md Fahim Sikder and Resmi Ramachandranpillai and Fredrik Heintz},
      year={2024},
      eprint={2307.12667},
      archivePrefix={arXiv},
      primaryClass={cs.LG},
      url={https://arxiv.org/abs/2307.12667}, 
}

@article{cohen2001uncertainty,
    title={The Uncertainty Principle for the Short-Time {Fourier} Transform and Wavelet Transform},
    author={Cohen, Leon},
    booktitle={Wavelet Transforms and Time-Frequency Signal Analysis},
    pages={217--245},
    year={2001},
    publisher={Birkhäuser}
}

@inproceedings{daubechies1988timefrequency,
    title={Time-frequency localization operators: A geometric phase space approach},
    author={Daubechies, Ingrid},
    journal={IEEE Transactions on Information Theory},
    volume={34},
    number={4},
    pages={605--612},
    year={1988}
}

@article{rioul1992timefrequency,
    title={Time-scale energy distributions: a general class extending wavelet transforms},
    author={Rioul, Olivier and Flandrin, Patrick},
    journal={IEEE Transactions on Signal Processing},
    volume={40},
    number={7},
    pages={1746--1757},
    year={1992}
}

@book{vetterli1995wavelets,
    title={Wavelets and subband coding},
    author={Vetterli, Martin and Kova{\v{c}}evi{\'c}, Jelena},
    year={1995},
    publisher={Prentice Hall}
}

@book{burrus1998introduction,
    title={Introduction to wavelets and wavelet transforms: a primer},
    author={Burrus, C Sidney and Gopinath, Ramesh A and Guo, Haitao},
    year={1998},
    publisher={Prentice Hall}
}

@misc{kazemi2022timeseriessynthesismultiscale,
      title={Time Series Synthesis via Multi-scale Patch-based Generation of Wavelet Scalogram}, 
      author={Amir Kazemi and Hadi Meidani},
      year={2022},
      eprint={2211.02620},
      archivePrefix={arXiv},
      primaryClass={eess.SP},
      url={https://arxiv.org/abs/2211.02620}, 
}

@INPROCEEDINGS{8595010,
  author={Zhao, Yi and Shen, Yanyan and Zhu, Yanmin and Yao, Junjie},
  booktitle={2018 IEEE International Conference on Data Mining (ICDM)}, 
  title={Forecasting Wavelet Transformed Time Series with Attentive Neural Networks}, 
  year={2018},
  volume={},
  number={},
  pages={1452-1457},
  doi={10.1109/ICDM.2018.00201}}

@inproceedings{NEURIPS2024_5e364212,
 author = {Yang, Xinyu and Sun, Yu and Yuan, Xiaojie and Chen, Xinyang},
 booktitle = {Advances in Neural Information Processing Systems},
 editor = {A. Globerson and L. Mackey and D. Belgrave and A. Fan and U. Paquet and J. Tomczak and C. Zhang},
 pages = {52595--52623},
 publisher = {Curran Associates, Inc.},
 title = {Frequency-aware Generative Models for Multivariate Time Series Imputation},
 url = {https://proceedings.neurips.cc/paper_files/paper/2024/file/5e364212327fa3a59ae3595b025c469f-Paper-Conference.pdf},
 volume = {37},
 year = {2024}
}

@misc{qian2024timeldmlatentdiffusionmodel,
      title={TimeLDM: Latent Diffusion Model for Unconditional Time Series Generation}, 
      author={Jian Qian and Bingyu Xie and Biao Wan and Minhao Li and Miao Sun and Patrick Yin Chiang},
      year={2024},
      eprint={2407.04211},
      archivePrefix={arXiv},
      primaryClass={cs.LG},
      url={https://arxiv.org/abs/2407.04211}, 
}

@misc{zhou2025multiorderwaveletderivativetransform,
      title={Multi-Order Wavelet Derivative Transform for Deep Time Series Forecasting}, 
      author={Ziyu Zhou and Jiaxi Hu and Qingsong Wen and James T. Kwok and Yuxuan Liang},
      year={2025},
      eprint={2505.11781},
      archivePrefix={arXiv},
      primaryClass={cs.LG},
      url={https://arxiv.org/abs/2505.11781}, 
}

@inproceedings{Sasal_2022,
   title={W-Transformers: A Wavelet-based Transformer Framework for Univariate Time Series Forecasting},
   url={http://dx.doi.org/10.1109/ICMLA55696.2022.00111},
   DOI={10.1109/icmla55696.2022.00111},
   booktitle={2022 21st IEEE International Conference on Machine Learning and Applications (ICMLA)},
   publisher={IEEE},
   author={Sasal, Lena and Chakraborty, Tanujit and Hadid, Abdenour},
   year={2022},
   month=dec, pages={671–676} 
}

@misc{arabi2024wavemaskmixexploringwaveletbasedaugmentations,
      title={Wave-Mask/Mix: Exploring Wavelet-Based Augmentations for Time Series Forecasting}, 
      author={Dona Arabi and Jafar Bakhshaliyev and Ayse Coskuner and Kiran Madhusudhanan and Kami Serdar Uckardes},
      year={2024},
      eprint={2408.10951},
      archivePrefix={arXiv},
      primaryClass={cs.LG},
      url={https://arxiv.org/abs/2408.10951}, 
}

@inproceedings{Schlter2010UsingWF,
  title={Using wavelets for time series forecasting: Does it pay off?},
  author={Stephan Schl{\"u}ter and Carola Deuschle},
  year={2010},
  url={https://api.semanticscholar.org/CorpusID:122828772}
}

@misc{hu2023neuralwaveletdomaindiffusion3d,
      title={Neural Wavelet-domain Diffusion for 3D Shape Generation, Inversion, and Manipulation}, 
      author={Jingyu Hu and Ka-Hei Hui and Zhengzhe Liu and Ruihui Li and Chi-Wing Fu},
      year={2023},
      eprint={2302.00190},
      archivePrefix={arXiv},
      primaryClass={cs.CV},
      url={https://arxiv.org/abs/2302.00190}, 
}

@inproceedings{NEURIPS2022_03474669,
 author = {Guth, Florentin and Coste, Simon and De Bortoli, Valentin and Mallat, Stephane},
 booktitle = {Advances in Neural Information Processing Systems},
 editor = {S. Koyejo and S. Mohamed and A. Agarwal and D. Belgrave and K. Cho and A. Oh},
 pages = {478--491},
 publisher = {Curran Associates, Inc.},
 title = {Wavelet Score-Based Generative Modeling},
 url = {https://proceedings.neurips.cc/paper_files/paper/2022/file/03474669b759f6d38cdca6fb4eb905f4-Paper-Conference.pdf},
 volume = {35},
 year = {2022}
}

@book{percival2000wavelet,
  title={Wavelet methods for time series analysis},
  author={Percival, Donald B and Walden, Andrew T},
  volume={4},
  year={2000},
  publisher={Cambridge University Press},
  address={Cambridge, UK},
  series={Cambridge Series in Statistical and Probabilistic Mathematics},
  isbn={9780521685085}
}

@article{sang2013review,
  title={A review on the applications of wavelet transform in hydrology time series analysis},
  author={Sang, Yan-Fang},
  journal={Atmospheric Research},
  volume={122},
  pages={8--15},
  year={2013},
  publisher={Elsevier},
  doi={10.1016/j.atmosres.2012.11.003}
}

@article{waveletadvantages2015,
author = {Patrik, Sleziak and Kamila, Hlavčová and Szolgay, Jan},
year = {2015},
month = {06},
pages = {},
title = {Advantages Of A Time Series Analysis Using Wavelet Transform As Compared With A Fourier Analysis},
volume = {23},
journal = {Slovak Journal of Civil Engineering},
doi = {10.1515/sjce-2015-0010}
}

@misc{eeg_eye_state_264,
  author       = {Roesler, Oliver},
  title        = {{EEG Eye State}},
  year         = {2013},
  howpublished = {UCI Machine Learning Repository},
  note         = {{DOI}: https://doi.org/10.24432/C57G7J}
}

@misc{smith2018superconvergencefasttrainingneural,
      title={Super-Convergence: Very Fast Training of Neural Networks Using Large Learning Rates}, 
      author={Leslie N. Smith and Nicholay Topin},
      year={2018},
      eprint={1708.07120},
      archivePrefix={arXiv},
      primaryClass={cs.LG},
      url={https://arxiv.org/abs/1708.07120}, 
}

@article{Peebles2022DiT,
  title={Scalable Diffusion Models with Transformers},
  author={William Peebles and Saining Xie},
  year={2022},
  journal={arXiv preprint arXiv:2212.09748},
}

@InProceedings{pmlr-v235-zhang24cn,
  title = 	 {The Emergence of Reproducibility and Consistency in Diffusion Models},
  author =       {Zhang, Huijie and Zhou, Jinfan and Lu, Yifu and Guo, Minzhe and Wang, Peng and Shen, Liyue and Qu, Qing},
  booktitle = 	 {Proceedings of the 41st International Conference on Machine Learning},
  pages = 	 {60558--60590},
  year = 	 {2024},
  editor = 	 {Salakhutdinov, Ruslan and Kolter, Zico and Heller, Katherine and Weller, Adrian and Oliver, Nuria and Scarlett, Jonathan and Berkenkamp, Felix},
  volume = 	 {235},
  series = 	 {Proceedings of Machine Learning Research},
  month = 	 {21--27 Jul},
  publisher =    {PMLR},
  url = 	 {https://proceedings.mlr.press/v235/zhang24cn.html}
}

@inproceedings{
li2024understanding,
title={Understanding Generalizability of Diffusion Models Requires Rethinking the Hidden Gaussian Structure},
author={Xiang Li and Yixiang Dai and Qing Qu},
booktitle={The Thirty-eighth Annual Conference on Neural Information Processing Systems},
year={2024},
url={https://openreview.net/forum?id=Sk2duBGvrK}
}

@inproceedings{
kadkhodaie2024generalization,
title={Generalization in diffusion models arises from geometry-adaptive harmonic representations},
author={Zahra Kadkhodaie and Florentin Guth and Eero P Simoncelli and St{\'e}phane Mallat},
booktitle={The Twelfth International Conference on Learning Representations},
year={2024},
url={https://openreview.net/forum?id=ANvmVS2Yr0}
}

@article{lee2019pywavelets,
  title = {PyWavelets: A Python package for wavelet analysis},
  author = {Lee, Gregory R. and Gommers, Ralf and Wasilewski, Filip and Wohlfahrt, Kai and O’Leary, Aaron},
  journal = {Journal of Open Source Software},
  volume = {4},
  number = {36},
  pages = {1237},
  year = {2019},
  doi = {10.21105/joss.01237}
}

@misc{paul2022psaganprogressiveselfattention,
      title={PSA-GAN: Progressive Self Attention GANs for Synthetic Time Series}, 
      author={Jeha Paul and Bohlke-Schneider Michael and Mercado Pedro and Kapoor Shubham and Singh Nirwan Rajbir and Flunkert Valentin and Gasthaus Jan and Januschowski Tim},
      year={2022},
      eprint={2108.00981},
      archivePrefix={arXiv},
      primaryClass={cs.LG},
      url={https://arxiv.org/abs/2108.00981}, 
}

@inproceedings{pei-etal-2025-leaf,
    title = "{LEAF}: Large Language Diffusion Model for Time Series Forecasting",
    author = "Pei, Yuhang  and
      Ren, Tao  and
      Wang, Yifan  and
      Sun, Zhipeng  and
      Ju, Wei  and
      Chen, Chong  and
      Hua, Xian-Sheng  and
      Luo, Xiao",
    editor = "Christodoulopoulos, Christos  and
      Chakraborty, Tanmoy  and
      Rose, Carolyn  and
      Peng, Violet",
    booktitle = "Findings of the Association for Computational Linguistics: EMNLP 2025",
    month = nov,
    year = "2025",
    address = "Suzhou, China",
    publisher = "Association for Computational Linguistics",
    url = "https://aclanthology.org/2025.findings-emnlp.58/",
    doi = "10.18653/v1/2025.findings-emnlp.58",
    pages = "1076--1091",
    ISBN = "979-8-89176-335-7"
}

@inproceedings{
    shen2024multiresolution,
    title={Multi-Resolution Diffusion Models for Time Series Forecasting},
    author={Lifeng Shen and Weiyu Chen and James Kwok},
    booktitle={The Twelfth International Conference on Learning Representations},
    year={2024},
    url={https://openreview.net/forum?id=mmjnr0G8ZY}
}

@misc{hou2025stagediffstagewiselongtermtime,
      title={Stage-Diff: Stage-wise Long-Term Time Series Generation Based on Diffusion Models}, 
      author={Xuan Hou and Shuhan Liu and Zhaohui Peng and Yaohui Chu and Yue Zhang and Yining Wang},
      year={2025},
      eprint={2508.21330},
      archivePrefix={arXiv},
      primaryClass={cs.LG},
      url={https://arxiv.org/abs/2508.21330}, 
}

@misc{fadlon2025diffusionmodelregulartime,
      title={A Diffusion Model for Regular Time Series Generation from Irregular Data with Completion and Masking}, 
      author={Gal Fadlon and Idan Arbiv and Nimrod Berman and Omri Azencot},
      year={2025},
      eprint={2510.06699},
      archivePrefix={arXiv},
      primaryClass={cs.LG},
      url={https://arxiv.org/abs/2510.06699}, 
}

@article{Li_2025,
   title={Population Aware Diffusion for Time Series Generation},
   volume={39},
   ISSN={2159-5399},
   url={http://dx.doi.org/10.1609/aaai.v39i17.34038},
   DOI={10.1609/aaai.v39i17.34038},
   number={17},
   journal={Proceedings of the AAAI Conference on Artificial Intelligence},
   publisher={Association for the Advancement of Artificial Intelligence (AAAI)},
   author={Li, Yang and Meng, Han and Bi, Zhenyu and Urnes, Ingolv T. and Chen, Haipeng},
   year={2025},
   month=apr, pages={18520–18529} 
}

@misc{suh2025timeautodiffunifiedframeworkgeneration,
      title={TimeAutoDiff: A Unified Framework for Generation, Imputation, Forecasting, and Time-Varying Metadata Conditioning of Heterogeneous Time Series Tabular Data}, 
      author={Namjoon Suh and Yuning Yang and Din-Yin Hsieh and Qitong Luan and Shirong Xu and Shixiang Zhu and Guang Cheng},
      year={2025},
      eprint={2406.16028},
      archivePrefix={arXiv},
      primaryClass={cs.LG},
      url={https://arxiv.org/abs/2406.16028}, 
}

@misc{naiman2024utilizingimagetransformsdiffusion,
      title={Utilizing Image Transforms and Diffusion Models for Generative Modeling of Short and Long Time Series}, 
      author={Ilan Naiman and Nimrod Berman and Itai Pemper and Idan Arbiv and Gal Fadlon and Omri Azencot},
      year={2024},
      eprint={2410.19538},
      archivePrefix={arXiv},
      primaryClass={cs.LG},
      url={https://arxiv.org/abs/2410.19538}, 
}

@misc{luo2025waveletfourierdiffuserfrequencyaware,
      title={Wavelet Fourier Diffuser: Frequency-Aware Diffusion Model for Reinforcement Learning}, 
      author={Yifu Luo and Yongzhe Chang and Xueqian Wang},
      year={2025},
      eprint={2509.19305},
      archivePrefix={arXiv},
      primaryClass={cs.LG},
      url={https://arxiv.org/abs/2509.19305}, 
}

@ARTICLE{11197480,
  author={Li, Jiacheng and Chen, Wei and Liu, Yican and Yang, Junmei and Zhou, Zhiheng and Zeng, Delu},
  journal={IEEE Transactions on Instrumentation and Measurement}, 
  title={Generative Self-Supervised Time-Series Forecasting Leveraging Wavelet Diffusion}, 
  year={2025},
  volume={74},
  number={},
  pages={1-15},
  doi={10.1109/TIM.2025.3619658}}

@misc{zhu2023edmsoundspectrogrambaseddiffusion,
      title={EDMSound: Spectrogram Based Diffusion Models for Efficient and High-Quality Audio Synthesis}, 
      author={Ge Zhu and Yutong Wen and Marc-André Carbonneau and Zhiyao Duan},
      year={2023},
      eprint={2311.08667},
      archivePrefix={arXiv},
      primaryClass={cs.SD},
      url={https://arxiv.org/abs/2311.08667}, 
}

@misc{vanukuri2025wavesimaginationunconditionalspectrogram,
      title={Waves of Imagination: Unconditional Spectrogram Generation using Diffusion Architectures}, 
      author={Rahul Vanukuri and Shafi Ullah Khan and Talip Tolga Sarı and Gokhan Secinti and Diego Patiño and Debashri Roy},
      year={2025},
      eprint={2510.10044},
      archivePrefix={arXiv},
      primaryClass={cs.NI},
      url={https://arxiv.org/abs/2510.10044}, 
}

@misc{hawthorne2022multiinstrumentmusicsynthesisspectrogram,
      title={Multi-instrument Music Synthesis with Spectrogram Diffusion}, 
      author={Curtis Hawthorne and Ian Simon and Adam Roberts and Neil Zeghidour and Josh Gardner and Ethan Manilow and Jesse Engel},
      year={2022},
      eprint={2206.05408},
      archivePrefix={arXiv},
      primaryClass={cs.SD},
      url={https://arxiv.org/abs/2206.05408}, 
}

@inproceedings{10.1145/3746027.3762245,
author = {Mao, Qirong and Wu, Qiwei and Liu, Na and Ding, Yakui and Gao, Lijian},
title = {Scattering-Conditioned Diffusion Models for Multiple Appropriate Facial Reaction Generation},
year = {2025},
isbn = {9798400720352},
publisher = {Association for Computing Machinery},
address = {New York, NY, USA},
url = {https://doi.org/10.1145/3746027.3762245},
doi = {10.1145/3746027.3762245},
booktitle = {Proceedings of the 33rd ACM International Conference on Multimedia},
pages = {13985–13991},
numpages = {7},
location = {Dublin, Ireland},
series = {MM '25}
}

@ARTICLE{10890904,
  author={Dong, Sihao and Xiao, Xiayang and Pan, Mingjun},
  journal={IEEE Geoscience and Remote Sensing Letters}, 
  title={Scattering Guided Image Despeckling: A Score-Based Diffusion Approach in Frequency Domain}, 
  year={2025},
  volume={22},
  number={},
  pages={1-5},
  doi={10.1109/LGRS.2025.3542383}
}

@misc{gama2018diffusionscatteringtransformsgraphs,
      title={Diffusion Scattering Transforms on Graphs}, 
      author={Fernando Gama and Alejandro Ribeiro and Joan Bruna},
      year={2018},
      eprint={1806.08829},
      archivePrefix={arXiv},
      primaryClass={cs.LG},
      url={https://arxiv.org/abs/1806.08829}, 
}

@article{ning2024dctdiff,
  title={DCTdiff: Intriguing Properties of Image Generative Modeling in the DCT Space},
  author={Ning, Mang and Li, Mingxiao and Su, Jianlin and Jia, Haozhe and Liu, Lanmiao and Bene{\v{s}}, Martin and Salah, Albert Ali and Ertugrul, Itir Onal},
  journal={arXiv preprint arXiv:2412.15032},
  year={2024}
}

@ARTICLE{10975753,
  author={Zhong, Linzhan and Chen, Fangni and Zheng, Bolun and Feng, Rui and Zhang, Lei and Wan, Jian},
  journal={IEEE Access}, 
  title={DCT-DiffPose: A Lightweight Diffusion Model With Multi-Hypothesis for 3D Human Pose Estimation}, 
  year={2025},
  volume={13},
  number={},
  pages={73319-73331},
  doi={10.1109/ACCESS.2025.3563898}}

@inproceedings{
naiman2024generative,
title={Generative Modeling of Regular and Irregular Time Series Data via Koopman {VAE}s},
author={Ilan Naiman and N. Benjamin Erichson and Pu Ren and Michael W. Mahoney and Omri Azencot},
booktitle={The Twelfth International Conference on Learning Representations},
year={2024},
url={https://openreview.net/forum?id=eY7sLb0dVF}
}

@misc{das2024decoderonlyfoundationmodeltimeseries,
      title={A decoder-only foundation model for time-series forecasting}, 
      author={Abhimanyu Das and Weihao Kong and Rajat Sen and Yichen Zhou},
      year={2024},
      eprint={2310.10688},
      archivePrefix={arXiv},
      primaryClass={cs.CL},
      url={https://arxiv.org/abs/2310.10688}, 
}

@misc{ansari2024chronoslearninglanguagetime,
      title={Chronos: Learning the Language of Time Series}, 
      author={Abdul Fatir Ansari and Lorenzo Stella and Caner Turkmen and Xiyuan Zhang and Pedro Mercado and Huibin Shen and Oleksandr Shchur and Syama Sundar Rangapuram and Sebastian Pineda Arango and Shubham Kapoor and Jasper Zschiegner and Danielle C. Maddix and Hao Wang and Michael W. Mahoney and Kari Torkkola and Andrew Gordon Wilson and Michael Bohlke-Schneider and Yuyang Wang},
      year={2024},
      eprint={2403.07815},
      archivePrefix={arXiv},
      primaryClass={cs.LG},
      url={https://arxiv.org/abs/2403.07815}, 
}

@book{wojtaszczyk1997mathematical,
  author    = {Wojtaszczyk, Przemys{\l}aw},
  title     = {A Mathematical Introduction to Wavelets},
  series    = {London Mathematical Society Student Texts},
  number    = {37},
  publisher = {Cambridge University Press},
  address   = {Cambridge},
  year      = {1997},
  doi       = {10.1017/CBO9780511623790},
  isbn      = {9780521578943}
}

@book{mallat2008wavelet,
  author    = {Mallat, St{\'e}phane},
  title     = {A Wavelet Tour of Signal Processing: The Sparse Way},
  edition   = {3rd},
  publisher = {Academic Press},
  year      = {2008}
}

@article{chen2024sdformer,
  title={Sdformer: Similarity-driven discrete transformer for time series generation},
  author={Chen, Zhicheng and Feng, Shibo and Zhang, Zhong and Xiao, Xi and Gao, Xingyu and Zhao, Peilin},
  journal={Advances in Neural Information Processing Systems},
  volume={37},
  pages={132179--132207},
  year={2024}
}

@article{feng2025msdformer,
  title={MSDformer: Multi-scale Discrete Transformer For Time Series Generation},
  author={Feng, Shibo and Chen, Zhicheng and Xiao, Xi and Zhang, Zhong and Li, Qing and Gao, Xingyu and Zhao, Peilin},
  journal={arXiv preprint arXiv:2505.14202},
  year={2025}
}

@misc{jeon2022gtgangeneralpurposetime,
      title={GT-GAN: General Purpose Time Series Synthesis with Generative Adversarial Networks}, 
      author={Jinsung Jeon and Jeonghak Kim and Haryong Song and Seunghyeon Cho and Noseong Park},
      year={2022},
      eprint={2210.02040},
      archivePrefix={arXiv},
      primaryClass={cs.LG},
      url={https://arxiv.org/abs/2210.02040}, 
}

@article{BULLMORE2004S234,
title = {Wavelets and functional magnetic resonance imaging of the human brain},
journal = {NeuroImage},
volume = {23},
pages = {S234-S249},
year = {2004},
note = {Mathematics in Brain Imaging},
issn = {1053-8119},
doi = {https://doi.org/10.1016/j.neuroimage.2004.07.012},
url = {https://www.sciencedirect.com/science/article/pii/S1053811904003775},
author = {Ed Bullmore and Jalal Fadili and Voichita Maxim and Levent Şendur and Brandon Whitcher and John Suckling and Michael Brammer and Michael Breakspear}
}

@article{maxim2005fractional,
  title     = {Fractional Gaussian noise, functional MRI and Alzheimer's disease},
  author    = {Maxim, Voichi{\c{t}}a and {\c{S}}endur, Levent and Fadili, Jalal and Suckling, John and Gould, Rebecca and Howard, Robert and Bullmore, Ed},
  journal   = {NeuroImage},
  volume    = {25},
  number    = {1},
  pages     = {141--158},
  year      = {2005},
  publisher = {Elsevier},
  doi       = {10.1016/j.neuroimage.2004.10.044}
}

@article{nejedly2020multicenter,
  title={Multicenter intracranial EEG dataset for classification of graphoelements and artifactual signals},
  author={Nejedly, Petr and Kremen, Vaclav and Sladky, Vladimir and Cimbalnik, Jan and Klimes, Petr and Plesinger, Filip and Mivalt, Filip and Travnicek, Vojtech and Viscor, Ivo and Pail, Martin and others},
  journal={Scientific data},
  volume={7},
  number={1},
  pages={179},
  year={2020},
  publisher={Nature Publishing Group UK London}
}

@article{kemp2000sleepedf,
  author  = {Kemp, Bob and Zwinderman, Aeilko H. and Tuk, Bert and
             Kamphuisen, Hilbert A. C. and Ober{\'e}y{\'e}, Josefien J. L.},
  title   = {Analysis of a Sleep-Dependent Neuronal Feedback Loop:
             The Slow-Wave Microcontinuity of the {EEG}},
  journal = {IEEE Transactions on Biomedical Engineering},
  volume  = {47},
  number  = {9},
  pages   = {1185--1194},
  year    = {2000},
  doi     = {10.1109/10.867928}
}

@article{goldberger2000physionet,
  author  = {Goldberger, Ary L. and Amaral, Luis A. N. and Glass, Leon and
             Hausdorff, Jeffrey M. and Ivanov, Plamen Ch. and Mark, Roger G. and
             Mietus, Joseph E. and Moody, George B. and Peng, Chung-Kang and
             Stanley, H. Eugene},
  title   = {{PhysioBank}, {PhysioToolkit}, and {PhysioNet}: Components of a
             New Research Resource for Complex Physiologic Signals},
  journal = {Circulation},
  volume  = {101},
  number  = {23},
  pages   = {e215--e220},
  year    = {2000},
  doi     = {10.1161/01.CIR.101.23.e215}
}

@inproceedings{
daras2023ambient,
title={Ambient Diffusion: Learning Clean Distributions from Corrupted Data},
author={Giannis Daras and Kulin Shah and Yuval Dagan and Aravind Gollakota and Alex Dimakis and Adam Klivans},
booktitle={Thirty-seventh Conference on Neural Information Processing Systems},
year={2023},
url={https://openreview.net/forum?id=wBJBLy9kBY}
}
\bibliographystyle{tmlr}

\appendix
\section{Mother Wavelet Families}
\label{appendix:wavelet}
Different wavelet families provide distinct characteristics of multi-scale decompositions through their specific filter coefficients $\{h_k\}$ and $\{g_k\}$ in the two-scale relations (Equation~\ref{eq:wavelet_scale}). In our experiments, we used five representative wavelet families, each satisfying different relationships between their order $p$ and filter length $F$. The vanishing moment properties and definitions presented here follow the PyWavelets framework implementation~\citep{lee2019pywavelets}.

\subsection{Daubechies Wavelets (db)}

Daubechies wavelets of order $p$ (e.g., db-$p$) provide orthogonality, compact support, and exactly $p$ vanishing moments for the wavelet function $\psi(t)$, with filter length $F = 2p$. The scaling function $\phi(t)$ has zero vanishing moments for orthogonal wavelets. The filter coefficients $\{h_k\}_{k=0}^{F-1}$ are derived from polynomial factorization to maximize regularity and maintain compact support.

For db2 (Daubechies-2 with $p=2$, $F=4$), the low-pass filter coefficients equal:
\begin{align}
h_0 &= \frac{1+\sqrt{3}}{4\sqrt{2}}, \quad h_1 = \frac{3+\sqrt{3}}{4\sqrt{2}}, \\
h_2 &= \frac{3-\sqrt{3}}{4\sqrt{2}}, \quad h_3 = \frac{1-\sqrt{3}}{4\sqrt{2}}.
\end{align}
The high-pass coefficients satisfy $g_k = (-1)^k h_{F-1-k}$. The orthogonality and vanishing moment conditions ensure that
\begin{align}
&\sum_{k=0}^{F-1} h_k = \sqrt{2}, \\
&\sum_{k=0}^{F-1} h_k h_{k+2m} = \delta_{m,0}, \\
&\sum_{k=0}^{F-1} k^j h_k = 0 \quad \text{for } j = 1, \ldots, p-1,
\end{align}
where $\delta_{m,0}$ is the Kronecker delta function, defined as:
\begin{equation}
\delta_{m,0} = \begin{cases}
1 & \text{if } m = 0 \\
0 & \text{if } m \neq 0.
\end{cases}
\end{equation}

\subsection{Symlets (sym)}
Symlets are modified Daubechies wavelets designed to improve symmetry while maintaining orthogonality and compact support. With a filter length $F = 2p$, Symlets have the same vanishing moment properties as Daubechies wavelets, i.e., $p$ vanishing moments for the wavelet function $\psi(t)$ and zero vanishing moments for the scaling function $\phi(t)$. They minimize an  asymmetry measure $A$ that quantifies the deviation from perfect symmetry, namely
\begin{equation}
A = \sum_{k=0}^{F-1} k \cdot |h_k|^2 - \frac{F-1}{2}\sum_{k=0}^{F-1} |h_k|^2,
\end{equation}
where the first term represents the weighted center of mass of the filter coefficients, while the second term represents the theoretical center for a perfectly symmetric filter. A lower value of $A$ indicates better symmetry.

For sym2 ($p=2$, $F=4$), the coefficients are optimized versions of db2 coefficients, satisfying the same orthogonality conditions but with improved phase linearity and near-symmetric properties for better temporal localization.

\subsection{Coiflets (coif)}
Coiflets of order $p$ are designed with balanced vanishing moments for both the scaling $\phi(t)$ and wavelet function $\psi(t)$, with filter length $F = 6p$. The wavelet function $\psi(t)$ has $2p$ vanishing moments while the scaling function $\phi(t)$ has $2p-1$ vanishing moments, providing more balance for the moments when compared to Daubechies wavelets (in which case the scaling function has zero vanishing moments). The filter coefficients satisfy extended moment conditions of the form
\begin{align}
\sum_{k=0}^{F-1} h_k &= \sqrt{2}, \\
\sum_{k=0}^{F-1} k^j h_k &= 0 \quad \text{for } j = 1, \ldots, 2p-1, \\
\sum_{k=0}^{F-1} g_k &= 0, \\
\sum_{k=0}^{F-1} k^j g_k &= 0 \quad \text{for } j = 1, \ldots, 2p.
\end{align}
For coif1 ($p=1$, $F=6$), the wavelet function has two vanishing moments and the scaling function has one vanishing moment. The six filter coefficients provide enhanced moment balancing between analysis and synthesis operations, with the scaling function having non-zero vanishing moments unlike orthogonal Daubechies wavelets.

\subsection{Biorthogonal Wavelets (bior)}
Biorthogonal wavelets use different filters for decomposition and reconstruction, denoted as bior$p_r.p_d$ where $p_r$ and $p_d$ are the orders that determine the vanishing moment properties. For a general bior$p_r.p_d$ wavelet, we have the following:
\begin{itemize}
    \item \emph{Wavelet function} $\psi(t)$: $p_r$ vanishing moments;
    \item \emph{Scaling function} $\phi(t)$: $p_d$ vanishing moments;
    \item \emph{Filter lengths}: These depend on the specific bior$p_r.p_d$ configuration and are not given by simple formulas like those of other wavelet families.
\end{itemize}
For decomposition one uses a low-pass filter $\{h_k\}$ and a high-pass filter $\{g_k\}$, while for reconstruction one uses a low-pass filter $\{\tilde{h}_k\}$ (denoted with tilde) and a high-pass filter $\{\tilde{g}_k\}$. The tilde notation $\tilde{\ }$ indicates the dual (reconstruction) filters that are different from the primal (decomposition) filters.

For bior2.2 ($p_r=p_d=2$), both decomposition and reconstruction wavelet functions have two vanishing moments, and both scaling functions have two vanishing moments, which result in perfect symmetry. The filters also satisfy the perfect reconstruction condition,
\begin{equation}
\sum_{k} h_k \tilde{h}_{k+2m} + g_k \tilde{g}_{k+2m} = \delta_{m,0}
\end{equation}

In the z-domain, where $H(z)$, $G(z)$, $\tilde{H}(z)$, and $\tilde{G}(z)$ are the z-transforms of the respective filter sequences, the perfect reconstruction condition is succinctly summarized as
\begin{equation}
H(z)\tilde{H}(z^{-1}) + H(-z)\tilde{H}(-z^{-1}) = 2.
\end{equation}

\subsection{Reverse Biorthogonal Wavelets (rbio)}

Reverse biorthogonal wavelets (rbio$p_r.p_d$) interchange the decomposition and reconstruction filter roles compared to standard biorthogonal wavelets, according to:
\begin{align}
h^{\text{rbio}}_k &= \tilde{h}^{\text{bior}}_k \\
g^{\text{rbio}}_k &= \tilde{g}^{\text{bior}}_k
\end{align}
For reverse biorthogonal wavelets, the vanishing moment assignment follows the same pattern as biorthogonal wavelets:
\begin{itemize}
    \item \emph{Wavelet function} $\psi(t)$ has $p_r$ vanishing moments.
    \item \emph{Scaling function} $\phi(t)$ has $p_d$ vanishing moments.
\end{itemize}
For rbio2.2 ($p_r = p_d = 2$), both the wavelet function $\psi(t)$ and scaling function $\phi(t)$ have two vanishing moments each, maintaining symmetric properties.

The choice of wavelet family affects the sparsity and localization properties of the decomposition, with symmetric wavelets (bior, rbio) providing better phase preservation, while orthogonal wavelets (db, sym) ensure energy conservation through orthogonality.

\section{Energy Regularization for Biorthogonal Wavelets}
\label{appendix:biorthogonal_parseval}

For biorthogonal and reverse biorthogonal wavelets, the discrete wavelet transform is not orthonormal, and the squared $\ell_2$ norm of the wavelet coefficients does not in general equal the signal energy. To arrive at the correct energy equations, let $\{\psi_i\}$ denote the primal (synthesis) wavelet basis and $\{\tilde{\psi}_i\}$ denote the corresponding dual (analysis) wavelet basis, satisfying the biorthogonality condition
\begin{equation}
\langle \psi_i, \tilde{\psi}_j \rangle = \delta_{ij}.
\end{equation}
Given a signal $x$, define the analysis coefficients and the corresponding primal-basis projections as
\begin{equation}
c_i = \langle x, \tilde{\psi}_i \rangle, \qquad \tilde{c}_i = \langle x, \psi_i \rangle.
\end{equation}
The signal energy then satisfies the \emph{biorthogonal}  Parseval identity
\begin{equation}
\|x\|_2^2 = \sum_i c_i \, \tilde{c}_i.
\end{equation}
Applying this identity to wavelet coefficients at each decomposition level $l$, a theoretically consistent energy regularization term can be defined as
\begin{equation}
\mathcal{L}_{\text{bio-energy}} = \mathbb{E}\!\left[ \sum_{l=1}^{L+1} \left| \sum_{n,j,k} \boldsymbol{C}^{(l)}_{n,j,k}\, \tilde{\boldsymbol{C}}^{(l)}_{n,j,k} - \sum_{n,j,k} \hat{\boldsymbol{C}}^{(l)}_{n,j,k}\, \widehat{\tilde{\boldsymbol{C}}}^{(l)}_{n,j,k}\right|\right],
\end{equation}
where $\boldsymbol{C}^{(l)}$ denotes the standard analysis coefficients with respect to the dual (analysis) wavelet basis, and $\tilde{\boldsymbol{C}}^{(l)}$ denotes the corresponding projections onto the primal (synthesis) wavelet basis.

\section{Experimental Details}
\subsection{Datasets}\label{appendix:datasets}
Table~\ref{tab:dataset_details} lists detailed properties of the datasets used in our experiments, and their repository links.
\begin{table}[htbp]
    \caption{Summary of the dataset types and statistics.}
    \centering
    \begin{tabular}{c|c|c|c}
        \hline
        \textbf{Dataset} & \textbf{\# of Samples} & \textbf{Dim} & \textbf{Source} \\
        \hline
        ETTh1 & $17420$ & $7$ & https://github.com/zhouhaoyi/ETDataset \\
        ETTh2 & $17420$ & $7$ & https://github.com/zhouhaoyi/ETDataset \\
        Stocks & $3685$ & $6$ & https://finance.yahoo.com/quote/GOOG \\
        Exchange Rate & $7588$ & $8$ & https://github.com/laiguokun/multivariate-time-series-data \\
        fMRI & $10000$ & $50$ & https://www.fmrib.ox.ac.uk/datasets/netsim \\
        EEG & $14980$ & $14$ & https://archive.ics.uci.edu/dataset/264/eeg+eye+state \\
        \hline
    \end{tabular}
    \label{tab:dataset_details}
\end{table}

\subsection{Metrics}\label{appendix:metrics}
\textbf{Discriminative Score.} The discriminative score captures how difficult it is for a classifier to distinguish between real and generated samples. The score is measured by $|\text{acc}-0.5|$, where acc is the classification accuracy. A score close to $0$ indicates that real and generated samples are indistinguishable to the classifier, while a score close to $0.5$ indicates they are very different. We follow the setup of TimeGAN~\citep{yoon2019time} using a $2$-layer GRU-based neural network as the classifier, trained with binary cross-entropy loss to distinguish between real (label=1) and synthetic (label=0) sequences.

\textbf{Predictive Score.} The predictive score captures how useful generated samples are for the forecasting task on real data. The score is measured by the mean absolute error (MAE) between predicted values and ground-truth values on test data. We follow TimeGAN~\citep{yoon2019time} using a $2$-layer GRU-based sequence predictor trained on synthetic data to predict the next time step features, evaluated on real sequences. Lower MAE values indicate better predictive utility of the generated samples.

\textbf{Context-FID.}~\citep{paul2022psaganprogressiveselfattention} The Fr\'echet Inception Distance (FID) measures the distance between two multivariate Gaussian distributions, i.e.,
\begin{equation}
\text{FID}(X, Y) = ||\mu_X - \mu_Y||^2 + \text{Tr}(\Sigma_X + \Sigma_Y - 2(\Sigma_X \Sigma_Y)^{1/2}),
\end{equation}
where $\mu_X, \mu_Y$ are the means and $\Sigma_X, \Sigma_Y$ are the covariance matrices of the two distributions. Context-FID adapts this to time series by replacing the Inception-v3 features with time series features. We follow Diffusion-TS~\citep{yuan2024diffusionts} using TS2Vec~\citep{yue2022ts2vecuniversalrepresentationtime} representations as the features. We extract embeddings from both real and generated sequences using a trained TS2Vec encoder, then compute FID in the embedding space. Lower Context-FID values indicate better distributional similarity.
    
\textbf{Correlational Score.} This metric assesses temporal dependencies by comparing cross-correlation matrices between real and generated data. For sequences with $D$ features, we compute the sample covariance matrix for each dataset, convert them to correlation matrices, and then measure the average absolute difference across all feature pairs according to
\begin{equation}
\text{Correlational Score} = \frac{1}{10} \sum_{i=1}^{D} \sum_{j=1}^{D} |\rho^{real}_{i,j} - \rho^{generated}_{i,j}|,
\end{equation}
where $\rho^{real}_{i,j}$ and $\rho^{generated}_{i,j}$ are the correlation coefficients between features $i$ and $j$ for real and generated data, respectively. Note that we follow the Diffusion-TS~\citep{yuan2024diffusionts} setup using the factor $\frac{1}{10}$, although $\frac{1}{D^2}$ could provide better normalization across different feature dimensions. The former choice of normalization ensures direct comparability with prior work.

\textbf{DTW-JS Distance.} We propose a Dynamic Time Warping Jensen-Shannon Distance (DTW-JS distance) metric, which combines DTW's temporal alignment capabilities with Jensen-Shannon divergence for distributional comparison. DTW computes the optimal alignment distance between two time series sequences $x$ and $y$ by minimizing
\begin{equation}
\text{DTW}(x, y) = \min_{\pi} \sum_{(i,j) \in \pi} |x_i - y_j|
\end{equation}
where $x$ and $y$ are two time series sequences, $\pi$ represents a warping path consisting of index pairs $(i,j)$ that map elements from sequence $x$ to sequence $y$, and the path must satisfy DTW constraints: monotonicity (indices only increase), continuity (no skipping), and boundary conditions (path starts at $(1,1)$ and ends at $(|x|,|y|)$). The warping allows sequences to be stretched or compressed along the time axis to find the best alignment, enabling DTW to handle sequences of different lengths and account for temporal shifts or speed variations between similar patterns. 

For our metric, we create a reference set $\mathcal{M}$ by randomly sampling from both real samples $\mathcal{R}$ and generated samples $\mathcal{G}$ (i.e., by taking the union of samples of these two sets, and ensuring that both sets have the same number of elements). For each sample $s$ in the real set $\mathcal{R}$ and generated set $\mathcal{G}$, we compute its mean DTW distance to all samples in the reference set:
\begin{equation}
d(s) = \frac{1}{|\mathcal{M}|} \sum_{r \in \mathcal{M}} \text{DTW}(s, r)
\end{equation}
This creates two collections of mean DTW distances, which we histogram across different samples $s$ to form distance distributions $D_{\mathcal{R}}$ and $D_{\mathcal{G}}$ for real and generated samples, respectively. We then apply Jensen-Shannon divergence to compute the distance between the two distance distributions, i.e.,
\begin{equation}
\text{DTW-JS}(D_{\mathcal{R}}, D_{\mathcal{G}}) = \frac{1}{2}[\text{KL}(D_{\mathcal{R}} || D_M) + \text{KL}(D_{\mathcal{G}} || D_M)]
\end{equation}
where $D_M = \frac{1}{2}(D_{\mathcal{R}} + D_{\mathcal{G}})$ is the mixture of the two distance distributions. This approach measures distributional similarity between real and generated samples while accounting for temporal alignment flexibility, providing a robust evaluation metric that captures both temporal structure and statistical properties.

\subsubsection{Validation and Sensitivity Analysis of DTW-JS Distance}
\begin{figure}
    \centering
    \includegraphics[width=0.5\linewidth]{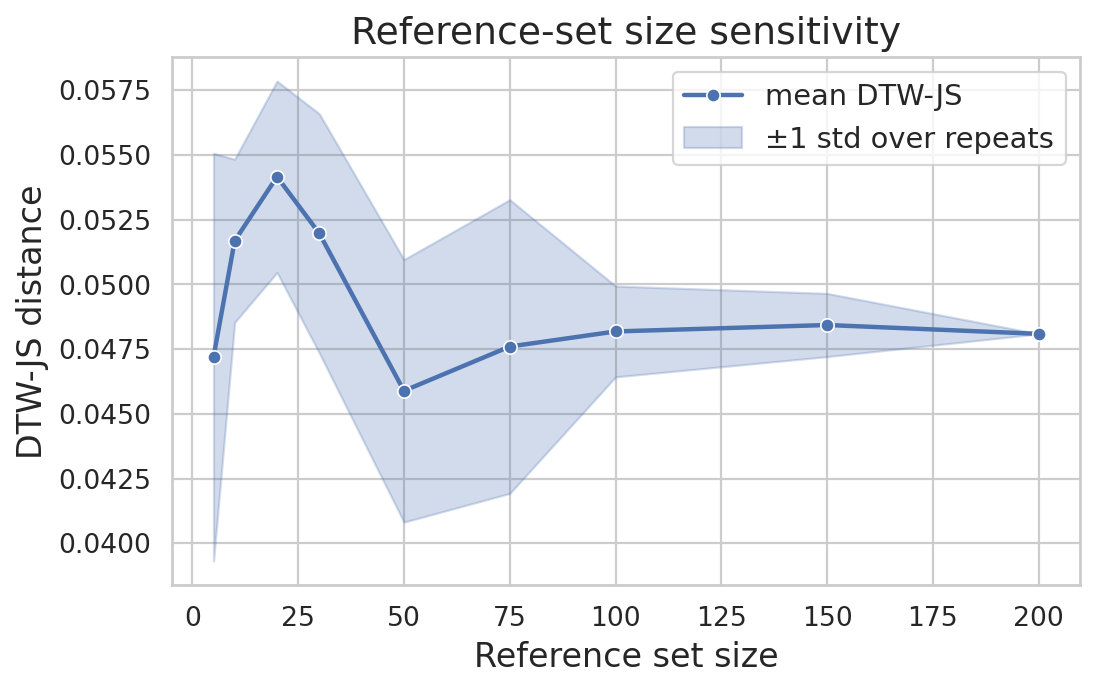}
    \caption{DTW-JS distance on EEG versus reference-set size $|\mathcal{M}|$. The metric is unstable for $|\mathcal{M}|<50$ and stabilizes for $|\mathcal{M}|>75$; we standardly use $|\mathcal{M}|=100$.}
    \label{fig:reference_set_size_sensitivity}
\end{figure}

\begin{figure}
    \centering
    \includegraphics[width=0.5\linewidth]{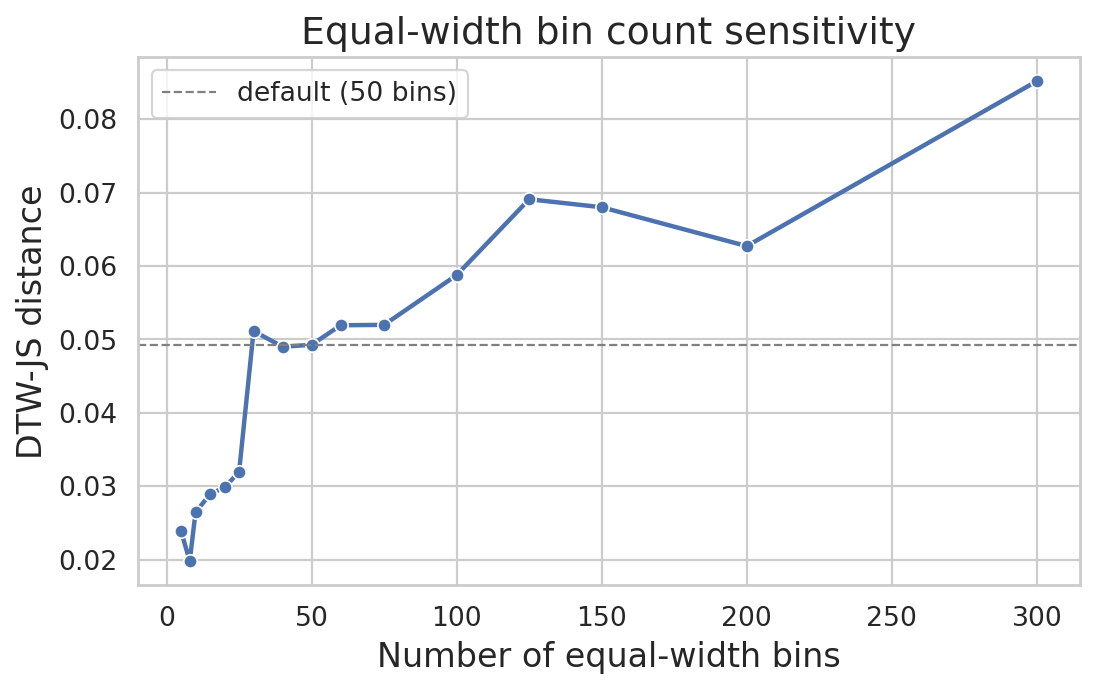}
    \caption{DTW-JS distance on EEG versus number of equal-width bins. Too few bins (<30) underestimate and too many (>100) inflate the distance; we standardly use 50 bins.}
    \label{fig:binning_sensitivity_equal_width}
\end{figure}

To validate the reliability of the proposed DTW-JS distance and characterize its behavior, we analyze its sensitivity to the two hyperparameters introduced in its definition, the reference-set size and the number of histogram bins; we also examine its computational cost. Unless otherwise stated, all analyses pertain to the EEG dataset as a representative example (on which WaveletDiff excels).
\paragraph{Sensitivity to reference-set size.}
We randomly sample $100$ sequences each from the generated and real datasets, and draw the reference set $M$ from their union. We vary the reference-set size from $5$ to $200$ samples. As shown in Figure~\ref{fig:reference_set_size_sensitivity}, DTW-JS is unstable for small reference sets ($n<50$) but converges and remains stable for larger sets ($n>75$). In practice we use a reference set of $100$ samples throughout our experiments.
\paragraph{Sensitivity to histogram/binning size choices.}
By fixing the reference set at $100$ samples, we analyze the sensitivity of DTW-JS to the number of bins used to approximate the distance distributions, with all bins of equal-width. As shown in Figure~\ref{fig:binning_sensitivity_equal_width}, too few bins ($<30$) make the distributions overly coarse and systematically underestimate the DTW-JS distance, whereas too many bins ($>100$) make them overly small and inflate the distance. We use $50$ bins throughout, which yields a stable estimate, and apply the same setting when evaluating all baseline models to ensure a fair comparison.
\paragraph{Computational cost}
Under the default setting, with the real, generated, and reference sets each containing $100$ samples, computing the DTW-JS distance once takes $5.77$ seconds on average. Following the same protocol as for the other four metrics, we repeat the computation five times and report the mean and standard deviation, giving a total cost of roughly $30$ seconds per evaluation.

\subsection{Wavelet Basis Function Analysis}
\label{appendix:wavelet_selection}

The choice of mother wavelet significantly influences the multi-scale decomposition characteristics and subsequent series generation quality. Different wavelet families exhibit distinct properties in terms of orthogonality, compact support, and smoothness, which directly affect the sparsity and localization of coefficients in the wavelet domain. This choice becomes especially critical when dealing with diverse dataset characteristics, as different signal types require wavelets that can optimally capture their specific temporal-spectral patterns. As shown in Table~\ref{tab:wavelet}, we systematically evaluated five representative wavelet families: Daubechies (db), Symlets (sym), Coiflets (coif), Biorthogonal (bior), and reverse Biorthogonal wavelets (rbio). 

\begin{table}[t]
    \caption{Mother wavelet selection.}
    \centering
    \resizebox{\textwidth}{!}{
    \begin{tabular}{c|c|c|c|c|c|c|c}
        \hline
        \textbf{Metrics} & \textbf{Wavelet} & \textbf{ETTh1} & \textbf{ETTh2} & \textbf{Stocks} & \textbf{Exchange Rate} & \textbf{fMRI} & \textbf{EEG} \\
        \hline
        \multirow{5}{*}{\begin{tabular}[c]{@{}c@{}}Discriminative\\ Score\\ (Lower the Better)\end{tabular}} 
        & db & \textbf{0.005±.005} & \textbf{0.008±.007} & 0.013±.007 & \textbf{0.004±.001} & 0.175±.071 & \textbf{0.006±.008} \\
        & sym & 0.023±.005 & 0.023±.005 & \textbf{0.005±.004} & 0.011±.009 & 0.196±.066 & 0.007±.003 \\
        & coif & 0.025±.010 & 0.031±.004 & 0.017±.009 & 0.073±.010 & \textbf{0.087±.077} & 0.014±.014 \\
        & bior & 0.022±.008 & 0.033±.002 & 0.012±.004 & 0.051±.012 & 0.273±.007 & 0.008±.004 \\
        & rbio & 0.057±.009 & 0.074±.008 & 0.010±.008 & 0.094±.010 & 0.129±.127 & 0.017±.009 \\
        \hline
        \multirow{5}{*}{\begin{tabular}[c]{@{}c@{}}Predictive\\ Score\\ (Lower the Better)\end{tabular}} 
        & db & 0.119±.002 & 0.106±.004 & \textbf{0.037±.000} & 0.037±.002 & \textbf{0.100±.000} & \textbf{0.000±.000} \\
        & sym & 0.117±.004 & 0.107±.003 & \textbf{0.037±.000} & \textbf{0.035±.003} & \textbf{0.100±.000} & \textbf{0.000±.000} \\
        & coif & \textbf{0.115±.005} & 0.106±.004 & \textbf{0.037±.000} & 0.036±.003 & \textbf{0.100±.000} & \textbf{0.000±.000} \\
        & bior & 0.122±.002 & 0.109±.004 & \textbf{0.037±.000} & 0.037±.001 & \textbf{0.100±.000} & \textbf{0.000±.000} \\
        & rbio & 0.121±.004 & \textbf{0.104±.001} & \textbf{0.037±.000} & 0.036±.001 & \textbf{0.100±.000} & \textbf{0.000±.000} \\
        \hline
        \multirow{5}{*}{\begin{tabular}[c]{@{}c@{}}Context-FID\\ Score\\ (Lower the Better)\end{tabular}} 
        & db & \textbf{0.020±.001} & \textbf{0.023±.002} & 0.024±.004 & \textbf{0.006±.000} & \textbf{0.104±.003} & \textbf{0.006±.000} \\
        & sym & 0.052±.004 & 0.051±.006 & 0.018±.002 & 0.009±.001 & 0.122±.007 & 0.011±.001 \\
        & coif & 0.079±.008 & 0.069±.009 & 0.018±.003 & 0.108±.008 & \textbf{0.104±.006} & \textbf{0.006±.000} \\
        & bior & 0.049±.003 & 0.156±.016 & \textbf{0.016±.001} & 0.088±.013 & 0.119±.004 & \textbf{0.006±.001} \\
        & rbio & 0.161±.005 & 0.225±.041 & \textbf{0.016±.002} & 0.175±.027 & 0.176±.008 & 0.007±.001 \\
        \hline
        \multirow{5}{*}{\begin{tabular}[c]{@{}c@{}}Correlational\\ Score\\ (Lower the Better)\end{tabular}} 
        & db & 0.043±.008 & 0.083±.016 & 0.006±.003 & \textbf{0.060±.020} & \textbf{1.073±.005} & \textbf{1.811±.963} \\
        & sym & 0.055±.008 & 0.073±.025 & 0.005±.003 & 0.066±.012 & 1.172±.048 & 2.164±.533 \\
        & coif & \textbf{0.036±.006} & \textbf{0.064±.011} & 0.007±.004 & 0.167±.032 & 1.177±.031 & 1.971±.969 \\
        & bior & 0.048±.011 & 0.094±.009 & 0.005±.003 & 0.137±.020 & 1.147±.033 & 3.034±.759 \\
        & rbio & 0.051±.008 & 0.099±.026 & \textbf{0.003±.004} & 0.161±.018 & 1.402±.034 & 1.959±.707 \\
        \hline
        \multirow{5}{*}{\begin{tabular}[c]{@{}c@{}}DTW-JS\\ Distance\\ (Lower the Better)\end{tabular}} 
        & db & \textbf{0.101±.016} & \textbf{0.064±.014} & 0.121±.013 & \textbf{0.121±.029} & 0.199±.043 & 0.055±.011 \\
        & sym & 0.123±.009 & 0.067±.023 & \textbf{0.106±.027} & 0.132±.017 & 0.283±.035 & \textbf{0.049±.015} \\
        & coif & 0.104±.015 & 0.086±.015 & 0.115±.016 & 0.157±.022 & \textbf{0.191±.011} & 0.062±.017 \\
        & bior & 0.117±.013 & 0.095±.019 & 0.109±.009 & 0.129±.033 & 0.280±.011 & 0.050±.009 \\
        & rbio & 0.110±.021 & 0.086±.033 & 0.119±.024 & 0.144±.037 & 0.464±.021 & 0.068±.022 \\
        \hline
    \end{tabular}}
    \label{tab:wavelet}
\end{table}

The results reveal dataset-specific preferred wavelets: Symlets work best for the Stocks dataset, likely due to their enhanced symmetry properties that better capture the near-symmetric fluctuations characteristic of financial time series. Coiflets demonstrate the best performance on the fMRI dataset, benefiting from balanced vanishing moments for both scaling and wavelet functions, which effectively capture the smooth yet complex spatiotemporal dynamics of brain activity. For the remaining datasets (ETTh1, ETTh2, Exchange Rate, EEG), Daubechies wavelets consistently provide the best overall performance. This ability to adapt the wavelet basis to match dataset characteristics represents an important advantage of WaveletDiff over frequency-domain approaches, which are constrained to use fixed Fourier basis functions regardless of the underlying signal properties, limiting their capacity to optimally represent diverse temporal patterns across different domains.

\begin{table}[htbp]
    \centering
    \caption{Ablation study results for the wavelet order and decomposition level.}
    \label{tab:ablation_order_level}
    \resizebox{\textwidth}{!}{
    \begin{tabular}{c|c|c|c|c|c|c|c|c}
        \hline
        \textbf{Metrics} & \textbf{Order} & \textbf{Level} & \textbf{ETTh1} & \textbf{ETTh2} & \textbf{Stocks} & \textbf{Exchange Rate} & \textbf{fMRI} & \textbf{EEG} \\
        \hline
        \multirow{3}{*}{\begin{tabular}[c]{@{}c@{}}Discriminative\\ Score\\ (Lower the Better)\end{tabular}} 
        & 2 & 3 & 0.005±.005 & \textbf{0.008±.007} & 0.013±.007 & \textbf{0.004±.001} & \textbf{0.175±.071} & \textbf{0.006±.008} \\
        & 2 & 4 & 0.011±.006 & 0.011±.004 & \textbf{0.011±.006} & 0.006±.002 & 0.198±.090 & 0.007±.003 \\
        & 1 & 3 & \textbf{0.004±.002} & 0.009±.001 & 0.016±.012 & \textbf{0.004±.003} & 0.298±.039 & 0.009±.003 \\
        \hline
        \multirow{3}{*}{\begin{tabular}[c]{@{}c@{}}Predictive\\ Score\\ (Lower the Better)\end{tabular}} 
        & 2 & 3 & 0.119±.002 & 0.106±.004 & \textbf{0.037±.000} & 0.037±.002 & \textbf{0.100±.000} & \textbf{0.000±.000} \\
        & 2 & 4 & 0.120±.002 & 0.107±.003 & \textbf{0.037±.000} & \textbf{0.036±.003} & \textbf{0.100±.000} & \textbf{0.000±.000} \\
        & 1 & 3 & \textbf{0.118±.004} & \textbf{0.104±.002} & \textbf{0.037±.000} & 0.037±.002 & 0.101±.000 & \textbf{0.000±.000} \\
        \hline
        \multirow{3}{*}{\begin{tabular}[c]{@{}c@{}}Context-FID\\ Score\\ (Lower the Better)\end{tabular}} 
        & 2 & 3 & 0.020±.001 & \textbf{0.023±.002} & 0.024±.004 & \textbf{0.006±.000} & \textbf{0.104±.003} & \textbf{0.006±.000} \\
        & 2 & 4 & 0.034±.005 & 0.027±.001 & 0.025±.002 & 0.008±.000 & 0.113±.005 & 0.007±.001 \\
        & 1 & 3 & \textbf{0.010±.001} & 0.041±.004 & \textbf{0.022±.002} & 0.008±.001 & 0.120±.004 & 0.013±.002 \\
        \hline
        \multirow{3}{*}{\begin{tabular}[c]{@{}c@{}}Correlational\\ Score\\ (Lower the Better)\end{tabular}} 
        & 2 & 3 & 0.043±.008 & 0.083±.016 & 0.006±.003 & 0.060±.020 & \textbf{1.073±.005} & \textbf{1.811±.963} \\
        & 2 & 4 & 0.049±.009 & \textbf{0.078±.021} & \textbf{0.005±.004} & \textbf{0.048±.010} & 1.185±.023 & 2.455±.360 \\
        & 1 & 3 & \textbf{0.033±.010} & 0.084±.008 & 0.009±.003 & \textbf{0.048±.010} & 1.287±.043 & 2.877±.698 \\
        \hline
        \multirow{3}{*}{\begin{tabular}[c]{@{}c@{}}DTW-JS\\ Distance\\ (Lower the Better)\end{tabular}} 
        & 2 & 3 & 0.101±.016 & 0.064±.014 & 0.121±.013 & 0.121±.029 & 0.199±.043 & 0.055±.011 \\
        & 2 & 4 & \textbf{0.098±.026} & \textbf{0.060±.016} & \textbf{0.107±.016} & \textbf{0.114±.027} & 0.249±.017 & \textbf{0.048±.012} \\
        & 1 & 3 & 0.101±.015 & 0.070±.018 & 0.117±.011 & 0.136±.039 & \textbf{0.165±.017} & 0.054±.008 \\
        \hline
    \end{tabular}
    }
\end{table}

\begin{table}[t]
    \centering
    \caption{Ablation study results for key WaveletDiff architectural components.}
    \label{tab:ablation}
    \resizebox{\textwidth}{!}{
    \begin{tabular}{c|c|c|c|c|c|c|c}
        \hline
        \textbf{Metrics} & \textbf{Methods} & \textbf{ETTh1} & \textbf{ETTh2} & \textbf{Stocks} & \textbf{Exchange Rate} & \textbf{fMRI} & \textbf{EEG} \\
        \hline
        \multirow{7}{*}{\begin{tabular}[c]{@{}c@{}}Discriminative\\ Score\\ (Lower the Better)\end{tabular}} 
        & WaveletDiff & \textbf{0.005±.005} & \textbf{0.008±.007} & \textbf{0.005±.004} & \textbf{0.004±.001} & 0.087±.077 & \textbf{0.006±.008} \\
        & coefficient prediction & 0.017±.013 & 0.040±.030 & 0.027±.014 & 0.059±.010 & 0.277±.012 & 0.020±.013 \\
        & w/o cross attention & 0.055±.054 & 0.028±.017 & 0.016±.012 & 0.052±.011 & 0.179±.016 & 0.006±.003 \\
        & cosine noise scheduler & 0.024±.014 & 0.021±.019 & 0.094±.013 & 0.012±.005 & 0.112±.037 & 0.500±.000 \\
        & DDIM sampling & 0.135±.094 & 0.017±.005 & 0.010±.003 & 0.006±.004 & 0.494±.006 & 0.024±.017 \\
        & shared transformer & 0.500±.000 & 0.500±.000 & 0.007±.007 & 0.078±.011 & \textbf{0.064±.055} & 0.015±.010 \\
        & \begin{tabular}[c]{@{}c@{}}w/o larger\\ approx.-level capacity\end{tabular} & 0.081±.005 & 0.071±.015 & 0.016±.007 & 0.153±.018 & 0.240±.014 & 0.017±.006 \\
        \hline
        \multirow{7}{*}{\begin{tabular}[c]{@{}c@{}}Predictive\\ Score\\ (Lower the Better)\end{tabular}} 
        & WaveletDiff & 0.119±.002 & \textbf{0.106±.004} & \textbf{0.037±.000} & 0.037±.002 & \textbf{0.100±.000} & \textbf{0.000±.000} \\
        & coefficient prediction & 0.120±.003 & 0.111±.003 & \textbf{0.037±.000} & 0.038±.003 & \textbf{0.100±.000} & \textbf{0.000±.000} \\
        & w/o cross attention & 0.119±.002 & \textbf{0.106±.002} & \textbf{0.037±.000} & 0.037±.001 & \textbf{0.100±.000} & \textbf{0.000±.000} \\
        & cosine noise scheduler & 0.119±.003 & 0.107±.003 & \textbf{0.037±.000} & 0.037±.002 & 0.103±.000 & 0.172±.239 \\
        & DDIM sampling & \textbf{0.118±.005} & \textbf{0.106±.004} & \textbf{0.037±.000} & \textbf{0.036±.002} & 0.101±.000 & \textbf{0.000±.000} \\
        & shared transformer & 0.283±.043 & 0.256±.064 & \textbf{0.037±.000} & 0.037±.001 & \textbf{0.100±.000} & \textbf{0.000±.000} \\
        & \begin{tabular}[c]{@{}c@{}}w/o larger\\ approx.-level capacity\end{tabular} & 0.127±.003 & 0.122±.003 & \textbf{0.037±.000} & 0.037±.002 & \textbf{0.100±.000} & \textbf{0.000±.000} \\
        \hline
        \multirow{7}{*}{\begin{tabular}[c]{@{}c@{}}Context-FID\\ Score\\ (Lower the Better)\end{tabular}} 
        & WaveletDiff & \textbf{0.020±.001} & \textbf{0.023±.002} & 0.018±.002 & \textbf{0.006±.000} & \textbf{0.104±.006} & \textbf{0.006±.000} \\
        & coefficient prediction & 0.027±.003 & 0.059±.004 & 0.053±.007 & 0.085±.008 & 0.131±.010 & 0.009±.001 \\
        & w/o cross attention & 0.118±.006 & 0.044±.004 & 0.039±.006 & 0.042±.004 & 0.170±.008 & 0.008±.001 \\
        & cosine noise scheduler & 0.125±.005 & 0.050±.009 & 0.116±.021 & 0.011±.002 & 0.329±.022 & 44246±3637 \\
        & DDIM sampling & 0.135±.006 & 0.049±.004 & \textbf{0.008±.002} & \textbf{0.006±.000} & 2.394±.043 & 0.009±.001 \\
        & shared transformer & 2177736 & 1210986 & 0.015±.004 & 0.072±.002 & 0.130±.002 & 0.011±.001 \\
        & \begin{tabular}[c]{@{}c@{}}w/o larger\\ approx.-level capacity\end{tabular} & 0.383±.039 & 0.352±.024 & 0.037±.005 & 0.247±.015 & 0.154±.012 & 0.007±.001 \\
        \hline
        \multirow{7}{*}{\begin{tabular}[c]{@{}c@{}}Correlational\\ Score\\ (Lower the Better)\end{tabular}} 
        & WaveletDiff & \textbf{0.043±.008} & 0.083±.016 & \textbf{0.005±.003} & \textbf{0.060±.020} & 1.177±.031 & 1.811±.963 \\
        & coefficient prediction & 0.056±.010 & 0.097±.029 & 0.006±.003 & 0.182±.019 & 2.005±.051 & 2.650±.314 \\
        & w/o cross attention & 0.046±.016 & 0.078±.016 & 0.006±.004 & 0.092±.035 & 1.755±.062 & 1.993±.738 \\
        & cosine noise scheduler & 0.050±.014 & 0.105±.026 & 0.029±.010 & 0.070±.009 & 2.175±.071 & 1.487±.579 \\
        & DDIM sampling & 0.057±.011 & 0.094±.018 & \textbf{0.005±.003} & 0.080±.031 & 2.113±.044 & \textbf{0.844±.150} \\
        & shared transformer & 0.526±.008 & 0.627±.005 & 0.011±.005 & 0.168±.023 & \textbf{1.085±.025} & 3.762±.790 \\
        & \begin{tabular}[c]{@{}c@{}}w/o larger\\ approx.-level capacity\end{tabular} & 0.063±.008 & \textbf{0.073±.012} & 0.010±.002 & 0.150±.019 & 1.852±.058 & 2.290±.929 \\
        \hline
        \multirow{7}{*}{\begin{tabular}[c]{@{}c@{}}DTW-JS\\ Distance\\ (Lower the Better)\end{tabular}} 
        & WaveletDiff & \textbf{0.101±.016} & \textbf{0.064±.014} & \textbf{0.106±.027} & \textbf{0.121±.029} & 0.191±.011 & 0.055±.011 \\
        & coefficient prediction & 0.115±.017 & 0.103±.030 & 0.133±.021 & 0.140±.024 & 0.204±.029 & 0.071±.019 \\
        & w/o cross attention & 0.116±.029 & 0.085±.013 & 0.127±.020 & 0.133±.025 & 0.348±.009 & 0.064±.007 \\
        & cosine noise scheduler & 0.102±.018 & 0.075±.007 & 0.127±.021 & 0.134±.010 & 0.357±.033 & 0.693±.000 \\
        & DDIM sampling & 0.112±.020 & 0.074±.018 & 0.113±.011 & 0.122±.022 & 0.693±.000 & \textbf{0.040±.014} \\
        & shared transformer & 0.693±.000 & 0.693±.000 & 0.119±.021 & 0.160±.016 & \textbf{0.107±.017} & 0.074±.016 \\
        & \begin{tabular}[c]{@{}c@{}}w/o larger\\ approx.-level capacity\end{tabular} & 0.138±.044 & 0.143±.020 & 0.121±.020 & 0.166±.013 & 0.213±.029 & 0.073±.026 \\
        \hline
    \end{tabular}
    }
\end{table}

\begin{table}[htbp]
\centering
\caption{Ablation study regarding energy weights across different sequence lengths.}
\label{tab:ablation_energy_weight}
\resizebox{\textwidth}{!}{%
\begin{tabular}{l|c|c|c|c|c|c|c}
\hline
\textbf{Dataset} & \textbf{Length} & \begin{tabular}[c]{@{}c@{}}\textbf{Energy}\\\textbf{Weight}\end{tabular} & \begin{tabular}[c]{@{}c@{}}\textbf{Discriminative}\\\textbf{Score}\end{tabular} & \begin{tabular}[c]{@{}c@{}}\textbf{Predictive}\\\textbf{Score}\end{tabular} & \begin{tabular}[c]{@{}c@{}}\textbf{Context-FID}\\\textbf{Score}\end{tabular} & \begin{tabular}[c]{@{}c@{}}\textbf{Correlational}\\\textbf{Score}\end{tabular} & \begin{tabular}[c]{@{}c@{}}\textbf{DTW-JS}\\\textbf{Distance}\end{tabular} \\
\hline
\multirow{6}{*}{ETTh1}
& \multirow{2}{*}{32}  & 0   & 0.022±.015 & 0.117±.003 & 0.098±.009 & 0.054±.010 & 0.114±.015 \\
&                      & 0.3 & \textbf{0.016±.004} & \textbf{0.119±.001} & \textbf{0.038±.005} & \textbf{0.050±.004} & \textbf{0.095±.022} \\
\cline{2-8}
& \multirow{2}{*}{64}  & 0   & \textbf{0.027±.017} & \textbf{0.108±.004} & 0.168±.009 & 0.064±.022 & 0.111±.013 \\
&                      & 0.3 & 0.028±.009 & 0.114±.007 & \textbf{0.088±.005} & \textbf{0.054±.009} & \textbf{0.095±.028} \\
\cline{2-8}
& \multirow{2}{*}{128} & 0   & 0.044±.026 & 0.121±.007 & 0.382±.022 & \textbf{0.056±.010} & 0.099±.007 \\
&                      & 0.3 & \textbf{0.034±.037} & \textbf{0.113±.005} & \textbf{0.256±.014} & 0.059±.021 & \textbf{0.105±.035} \\
\hline
\multirow{6}{*}{\shortstack{Exchange\\Rate}}
& \multirow{2}{*}{32}  & 0   & \textbf{0.011±.005} & \textbf{0.035±.002} & \textbf{0.013±.001} & \textbf{0.064±.010} & \textbf{0.108±.037} \\
&                      & 0.3 & 0.032±.013 & 0.036±.001 & 0.057±.006 & 0.088±.025 & 0.126±.016 \\
\cline{2-8}
& \multirow{2}{*}{64}  & 0   & 0.128±.007 & 0.036±.002 & 0.432±.041 & 0.191±.052 & 0.209±.025 \\
&                      & 0.3 & \textbf{0.020±.005} & \textbf{0.035±.001} & \textbf{0.022±.003} & \textbf{0.066±.024} & \textbf{0.132±.015} \\
\cline{2-8}
& \multirow{2}{*}{128} & 0   & 0.166±.017 & 0.036±.002 & 0.812±.146 & 0.217±.016 & 0.184±.008 \\
&                      & 0.3 & \textbf{0.026±.008} & \textbf{0.034±.003} & \textbf{0.052±.003} & \textbf{0.065±.026} & \textbf{0.124±.018} \\
\hline
\end{tabular}%
}
\end{table}

\begin{table}[htbp]
\centering
\caption{WaveletDiff v.s. MSDformer on short sequences (length $24$) generation on iEEG and Sleep-EDF.}
\label{tab:ieeg_sleepedf}
\resizebox{0.5\textwidth}{!}{%
\begin{tabular}{c|c|c|c}
\hline
\textbf{Metric} & \textbf{Methods} & \textbf{iEEG} & \textbf{Sleep-EDF} \\
\hline
\multirow{2}{*}{\begin{tabular}[c]{@{}c@{}}Discriminative\\ Score \end{tabular}}
& WaveletDiff & \textbf{0.009$\pm$.008} & \textbf{0.004$\pm$.004} \\
& MSDformer & 0.160$\pm$.041 & 0.006$\pm$.003 \\
\hline
\multirow{2}{*}{\begin{tabular}[c]{@{}c@{}}Predictive\\ Score \end{tabular}}
& WaveletDiff & \textbf{0.024$\pm$.002} & \textbf{0.064$\pm$.000} \\
& MSDformer & 0.069$\pm$.001 & \textbf{0.064$\pm$.000} \\
\hline
\multirow{2}{*}{\begin{tabular}[c]{@{}c@{}}Context-FID\\ Score \end{tabular}}
& WaveletDiff & \textbf{0.006$\pm$.000} & 0.043$\pm$.002 \\
& MSDformer & 0.826$\pm$.140 & \textbf{0.001$\pm$.000} \\
\hline
\multirow{2}{*}{\begin{tabular}[c]{@{}c@{}}Correlational\\ Score \end{tabular}}
& WaveletDiff & \textbf{0.000$\pm$.000} & \textbf{0.000$\pm$.000} \\
& MSDformer & \textbf{0.000$\pm$.000} & \textbf{0.000$\pm$.000} \\
\hline
\multirow{2}{*}{\begin{tabular}[c]{@{}c@{}}DTW-JS\\ distance \end{tabular}}
& WaveletDiff & \textbf{0.136$\pm$.016} & \textbf{0.091$\pm$.027} \\
& MSDformer & 0.230$\pm$.026 & 0.099$\pm$.019 \\
\hline
\end{tabular}%
}
\end{table}

\subsection{Wavelet Configuration Selection}\label{appendix:wavelet_configuration}
We describe next how the three key wavelet hyperparameters — family, order, and decomposition level — are selected in WaveletDiff. While these can be tuned per dataset, our default heuristics yield strong performance without dataset-specific tuning.
\paragraph{Wavelet Family.} The default wavelet family in WaveletDiff is the Daubechies wavelet family, which achieves consistently strong performance across all datasets (see Table~\ref{tab:wavelet}). The ability to select different wavelet families for different signals is an additional feature of WaveletDiff that can lead to further improvements in generative performance on specific datasets. In our experiments, dataset-specific wavelet selection was performed by choosing the wavelet family with the best validation performance, which requires additional hyperparameter tuning and computational cost. However, the results in Table~\ref{tab:wavelet} demonstrate that such tuning is not really necessary in order to obtain strong performance in general, as the default Daubechies wavelet family already performs competitively across all benchmarks. It is important to notice that rigorous mathematical results are readily available to suggest which wavelet families are most amenable for which types of time series data decomposition/analysis (i.e., stationarity/nonstationarity, smoothness, sparseness etc)~\citep{wojtaszczyk1997mathematical, mallat2008wavelet}. This is nevertheless not directly translatable into amenability for wavelet-domain generation, as seen on the example of fMRI data from the main text.
\paragraph{Wavelet Order.} The wavelet order is selected based on the sequence length to avoid excessive boundary artifacts and ensure a meaningful multilevel decomposition. Specifically, for Daubechies wavelets, we use order $p=2$ for sequence lengths $T\le32$, $p=4$ for $33\le T\le64$, $p=6$ for $65\le T\le128$, and $p=8$ otherwise.
\paragraph{Decomposition Level.} The decomposition level is based on the sequence length, wavelet family, and wavelet order. Following the PyWavelets framework, we compute the maximum decomposition level as $\lfloor \log_2(\frac{T}{F - 1})\rfloor$. We then clip the resulting value to the range $[3,7]$ to avoid excessively shallow or deep decompositions, which would result in too few or too many wavelet-level transformer modules, respectively.

We further evaluate custom selections of wavelet order and decomposition levels (Table~\ref{tab:ablation_order_level}). The results show that moderate changes to these wavelet configurations lead to similar performance, indicating that WaveletDiff is relatively robust to the specific choice of wavelet parameters.

\paragraph{Wavelet Implementation Details.} We use the PyWavelets implementation for all DWT and IDWT operations. Boundary effects are handled using symmetric extension. We follow the standard normalization conventions of the selected wavelet family as implemented in PyWavelets. Since we are using the symmetric extension, the energy regularizer is heuristic rather than exact.

\subsection{Irregular Sampling}
\label{appendix:irregular}
We further evaluate WaveletDiff under the irregular sampling setting, where a subset of time steps is randomly removed from each training window before model training. Following the preprocessing protocol of KoVAE~\citep{naiman2024generative} and GT-GAN~\citep{jeon2022gtgangeneralpurposetime}, we use cubic spline interpolation to impute the missing values and then train WaveletDiff on the interpolated time series. As shown in Table~\ref{tab:irregular_sampling}, WaveletDiff maintains consistently strong performance across all missing rates.

\begin{table}[h]
\centering
\caption{Irregular sampling with different missing rates ($30\%, 50\%, 70\%$).}
\label{tab:irregular_sampling}
\resizebox{\textwidth}{!}{
\begin{tabular}{c|c|c|c|c|c|c}
\hline
\textbf{Dataset}
& \begin{tabular}[c]{@{}c@{}}\textbf{Missing}\\ \textbf{Rate}\end{tabular}
& \begin{tabular}[c]{@{}c@{}}\textbf{Discriminative}\\ \textbf{Score}\end{tabular}
& \begin{tabular}[c]{@{}c@{}}\textbf{Predictive}\\ \textbf{Score}\end{tabular}
& \begin{tabular}[c]{@{}c@{}}\textbf{Context-FID}\\ \textbf{Score}\end{tabular}
& \begin{tabular}[c]{@{}c@{}}\textbf{Correlational}\\ \textbf{Score}\end{tabular}
& \begin{tabular}[c]{@{}c@{}}\textbf{DTW-JS}\\ \textbf{Distance}\end{tabular} \\
\hline
\multirow{4}{*}{ETTh1}
& 0 (Regular) & 0.005±.005 & 0.119±.002 & 0.020±.001 & 0.043±.008 & 0.101±.016 \\
& 0.3 & 0.014±.002 & 0.120±.003 & 0.019±.001 & 0.064±.029 & 0.115±.030 \\
& 0.5 & 0.009±.006 & 0.121±.004 & 0.028±.004 & 0.045±.005 & 0.102±.026 \\
& 0.7 & 0.015±.009 & 0.121±.002 & 0.078±.013 & 0.056±.008 & 0.082±.011 \\
\hline
\multirow{4}{*}{ETTh2}
& 0   & 0.008±.007 & 0.106±.004 & 0.023±.002 & 0.083±.016 & 0.064±.014 \\
& 0.3 & 0.033±.002 & 0.106±.004 & 0.122±.011 & 0.109±.013 & 0.087±.019 \\
& 0.5 & 0.013±.007 & 0.105±.003 & 0.032±.003 & 0.072±.011 & 0.077±.019 \\
& 0.7 & 0.016±.007 & 0.108±.002 & 0.049±.002 & 0.073±.015 & 0.095±.013 \\
\hline
\multirow{4}{*}{Stocks}
& 0   & 0.005±.004 & 0.037±.000 & 0.018±.002 & 0.005±.003 & 0.106±.027 \\
& 0.3 & 0.009±.007 & 0.037±.000 & 0.022±.003 & 0.007±.004 & 0.116±.022 \\
& 0.5 & 0.020±.008 & 0.037±.000 & 0.018±.003 & 0.005±.003 & 0.118±.012 \\
& 0.7 & 0.011±.005 & 0.037±.000 & 0.027±.003 & 0.006±.004 & 0.131±.016 \\
\hline
\multirow{4}{*}{Exchange Rate}
& 0   & 0.004±.001 & 0.037±.002 & 0.006±.000 & 0.060±.020 & 0.121±.029 \\
& 0.3 & 0.011±.005 & 0.035±.001 & 0.010±.001 & 0.089±.039 & 0.119±.014 \\
& 0.5 & 0.019±.008 & 0.036±.002 & 0.030±.002 & 0.094±.036 & 0.129±.026 \\
& 0.7 & 0.010±.007 & 0.036±.001 & 0.010±.001 & 0.067±.019 & 0.136±.015 \\
\hline
\multirow{4}{*}{fMRI}
& 0   & 0.087±.077 & 0.100±.000 & 0.104±.006 & 1.177±.031 & 0.191±.011 \\
& 0.3 & 0.304±.018 & 0.100±.000 & 0.371±.019 & 1.196±.038 & 0.605±.033 \\
& 0.5 & 0.344±.033 & 0.101±.000 & 1.109±.022 & 1.512±.067 & 0.564±.028 \\
& 0.7 & 0.380±.016 & 0.102±.000 & 1.407±.063 & 1.781±.071 & 0.318±.017 \\
\hline
\multirow{4}{*}{EEG}
& 0   & 0.006±.008 & 0.000±.000 & 0.006±.000 & 1.811±.963 & 0.055±.011 \\
& 0.3 & 0.016±.009 & 0.000±.000 & 0.008±.001 & 2.595±.428 & 0.057±.008 \\
& 0.5 & 0.010±.007 & 0.000±.000 & 0.010±.001 & 3.683±1.025 & 0.054±.016 \\
& 0.7 & 0.009±.005 & 0.000±.000 & 0.007±.001 & 4.071±.797 & 0.043±.025 \\
\hline
\end{tabular}
}
\end{table}

\begin{figure}
    \centering
    \includegraphics[width=0.7\linewidth]{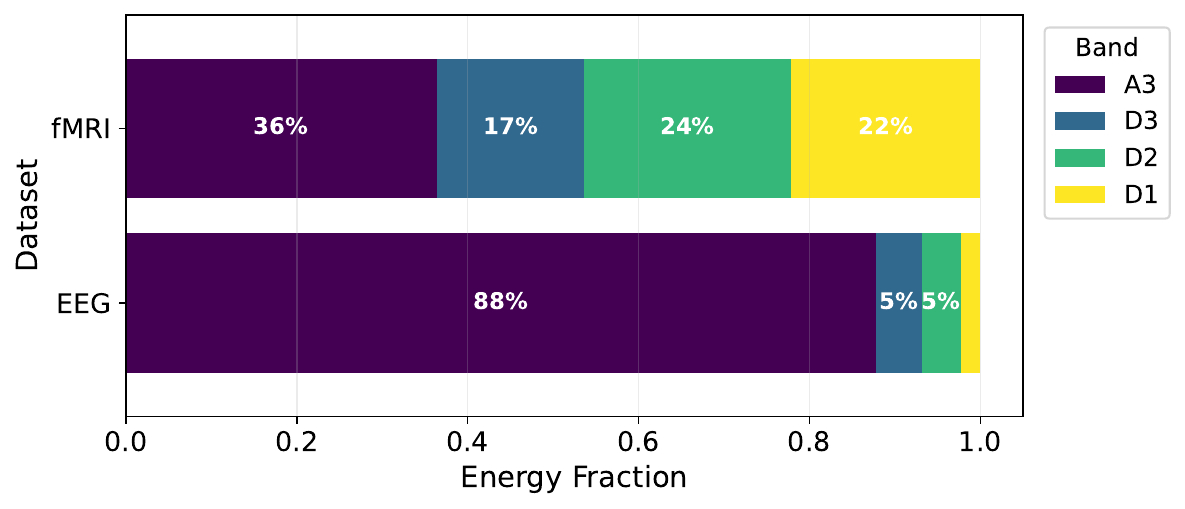}
    \caption{Energy distributions for wavelet coefficients at different levels of fMRI and EEG datasets. This has potential implications on how to select appropriate transformer structure for the levels.}
    \label{fig:energy_fraction_by_band}
\end{figure}

\subsection{Training Configuration Details}\label{appendix:training_configuration}
We provide training configuration information needed for reproducibility. The WaveletDiff model uses an embedding dimension of $256$ with $8$ attention heads across $8$ transformer layers, a time embedding dimension of $128$, and dropout rate of $0.1$. The approximation level transformer uses twice as many embedding dimensions ($512$) and $10$ layers to capture the critically important low-frequency information. 

The diffusion process employs $1000$ timesteps with an exponential noise schedule. The exponential noise schedule is of the form
\begin{equation}
\beta_t = \beta_{start} + (\beta_{end} - \beta_{start}) \cdot \left(1 - e^{-\gamma \cdot t}\right)
\end{equation}
where $\beta_{start} = 0.0001$, $\beta_{end} = 0.02$, $\gamma = 2.0$ is the exponential decay rate, $t \in [0,1]$ is the normalized timestep, and $T = 1000$ is the total number of timesteps. The coefficient-weighted loss strategy assigns an approximation coefficient weight of $2.0$ to emphasize low-frequency components.
We train for $5000$ epochs with batch size of $512$.

For optimization, we use the AdamW optimizer with initial learning rate $2 \times 10^{-4}$ and weight decay $1 \times 10^{-5}$. We employ a one-cycle learning rate schedule~\citep{smith2018superconvergencefasttrainingneural} with cosine annealing strategy. The learning rate follows a two-phase schedule: Linear warm-up for the first $30\%$ of training, then cosine annealing for the remaining $70\%$:
\begin{align}
lr(e) = \begin{cases}
lr_{base} + (lr_{max} - lr_{base}) \times \frac{e}{p \times E}, & \text{if } e \leq p \times E; \\
lr_{final} + (lr_{max} - lr_{final}) \times \left(\frac{1 + \cos\left(\pi \times \frac{e - p \times E}{(1-p) \times E}\right)}{2}\right), & \text{if } e > p \times E.
\end{cases}
\end{align}

where $e$ is the current epoch, $lr_{base} = 4 \times 10^{-5}$, $lr_{max} = 1 \times 10^{-3}$, $lr_{final} = 4 \times10^{-9}$, $p = 0.3$, and $E = 5000$ total epochs.

\subsection{Computational Setup and Training Time Analysis}
\label{appendix:computation}
WaveletDiff experiments were conducted on a single NVIDIA H100 GPU with 80GB memory using PyTorch 2.7.1 with CUDA 11.8, with training performed for 5000 epochs. Our WaveletDiff model contains approximately 63M trainable parameters. FourierDiffusion baseline results were obtained on a Tesla T4 GPU due to hardware constraints, following their original configuration and training settings. We note that the difference in GPU hardware was unavoidable due to High-Performance Computing (HPC) resource availability and the PyTorch Lightning deployment used for WaveletDiff, and was not selected to give any unfair advantage to our model. Tables~\ref{tab:WaveletDiff_training_times} and~\ref{tab:fourier_training_times} present the training times for WaveletDiff and FourierDiffusion across different datasets and sequence lengths, respectively. 
These measurements are provided as implementation details and should not be interpreted as a direct comparison of computational efficiency, since the experiments were conducted on different hardware platforms. Overall, the reported runtimes demonstrate that WaveletDiff can be trained within a practical time budget on a single modern GPU, indicating that compute time is not a bottleneck. In addition, we also provided a comparison of training time and model size between WaveletDiff and MSDformer on the same device (NVIDIA A100) in Table~\ref{tab:short_generation_results}. The results show that WaveletDiff has a slightly smaller model size and shorter training time than the Transformer-based baseline. We also showed the inference time of WaveletDiff in Table~\ref{tab:inference_time}, demonstrating that sampling can be done in seconds.

\begin{table}[htbp]
\centering
\caption{Training times (hours:minutes:seconds) for WaveletDiff across different datasets and sequence lengths on single NVIDIA H100 GPU.}
\label{tab:WaveletDiff_training_times}
\begin{tabular}{c|c|c|c|c|c|c}
\hline
\textbf{sequence length} & \textbf{ETTh1} & \textbf{ETTh2} & \textbf{Stocks} & \textbf{Exchange Rate} & \textbf{fMRI} & \textbf{EEG}\\
\hline
24 & 3:45:54 & 3:36:34 & 1:07:05 & 1:36:59 & 2:40:13 & 3:22:06 \\
32 & 3:45:33 & 3:38:23 & 1:02:49 & 1:46:32 & 2:42:41 & 3:24:07 \\
64 & 3:24:46 & 4:19:04 & 1:12:34 & 2:03:44 & 3:15:34 & 4:13:58 \\
128 & 5:06:03 & 6:27:03 & 1:33:53 & 2:59:33 & 4:32:52 & 6:01:14 \\
\hline
\end{tabular}
\end{table}

\begin{table}[htbp]
\centering
\caption{Training times (hours:minutes:seconds) for FourierDiffusion across different datasets and sequence lengths on single NVIDIA Tesla T4 GPU.}
\label{tab:fourier_training_times}
\begin{tabular}{c|c|c|c|c|c|c}
\hline
\textbf{sequence length} & \textbf{ETTh1} & \textbf{ETTh2} & \textbf{Stocks} & \textbf{Exchange Rate} & \textbf{fMRI} & \textbf{EEG}\\
\hline
24 & 1:09:51 & 1:19:01 & 1:57:23 & 2:21:39 & 2:21:38 & 2:39:03 \\
32 & 1:38:56 & 1:39:13 & 2:10:41 & 2:30:54 & 2:27:28 & 3:47:09 \\
64 & 4:49:37 & 4:49:39 & 3:14:13 & 2:31:45 & 3:58:26 & 5:21:55 \\
128 & 10:14:05 & 10:14:04 & 6:35:18 & 3:46:43 & 6:29:26 & 7:16:49 \\
\hline
\end{tabular}
\end{table}

\begin{table}[htbp]
\centering
\caption{Inference time (seconds per 1,000 samples) for WaveletDiff across different datasets on single NVIDIA H100 GPU.}
\begin{tabular}{lcccccc}
\toprule
 & ETTh1 & ETTh2 & Stocks & Exchange Rate & fMRI & EEG \\
\midrule
Time (sec) & 35 & 25 & 27 & 27 & 33 & 23 \\
\bottomrule
\end{tabular}
\label{tab:inference_time}
\end{table}

\section{Visualization}
\subsection{T-SNE and Data Distribution on Short Sequence Generation}\label{appendix:tsne_pdf_short}
\begin{figure}[t]
    \centering
    \resizebox{0.8\textwidth}{!}{
    \begin{subfigure}[b]{0.3\textwidth}
        \includegraphics[width=\textwidth]{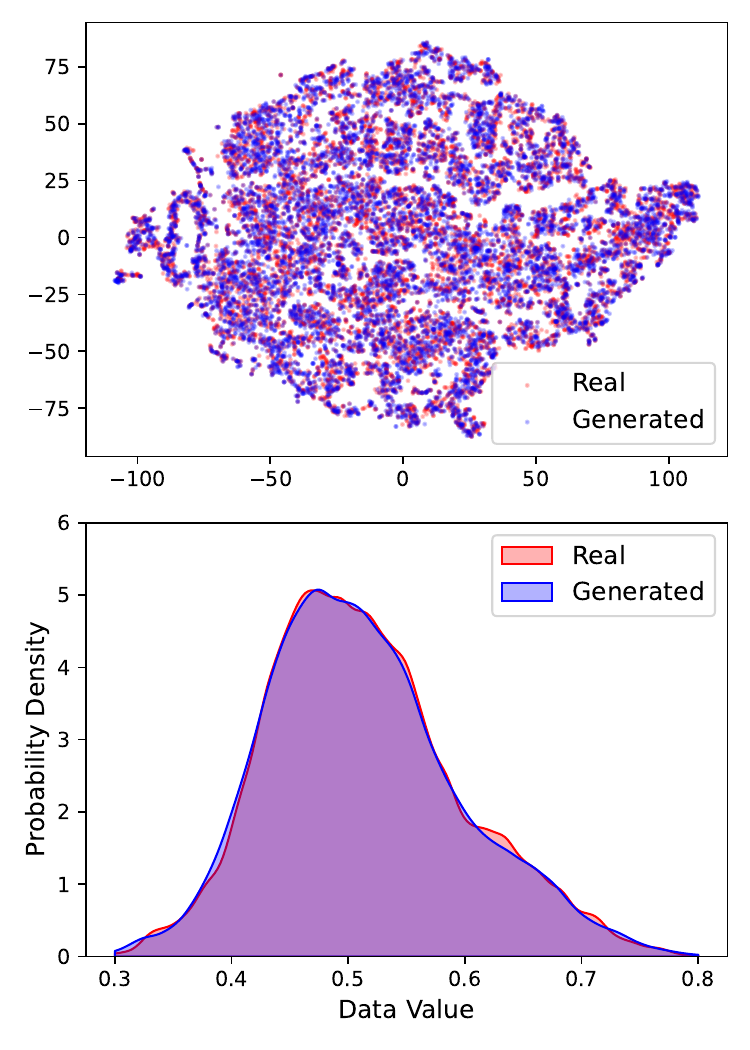}
        \caption{WaveletDiff}
    \end{subfigure}
    \hfill
    \begin{subfigure}[b]{0.3\textwidth}
        \includegraphics[width=\textwidth]{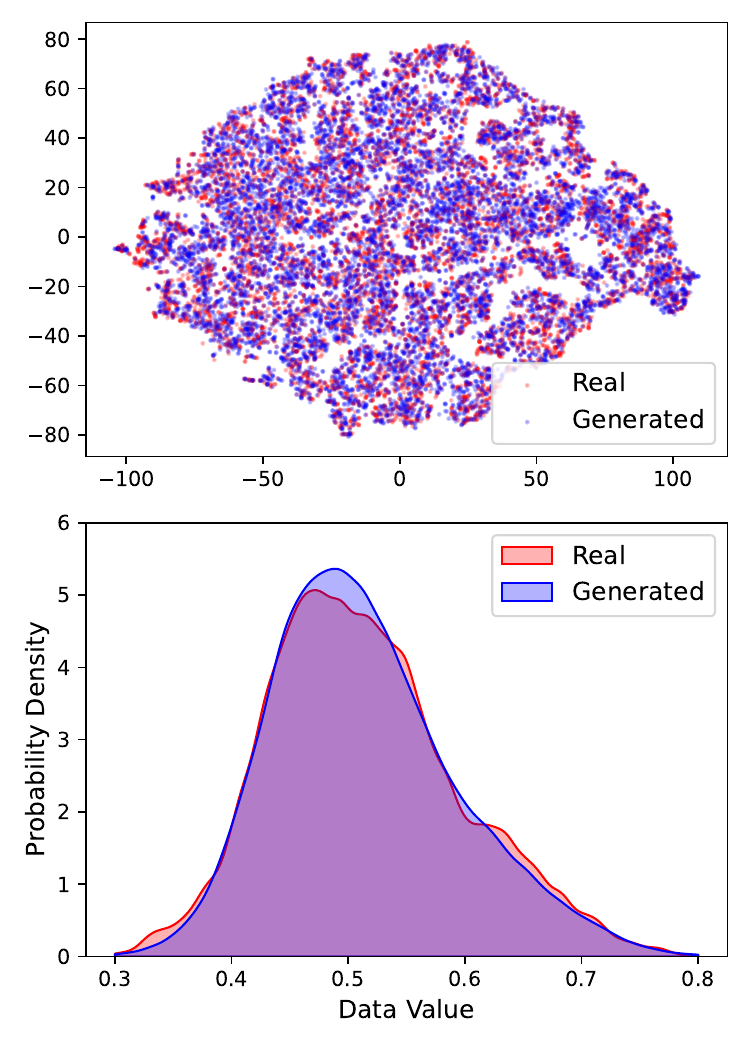}
        \caption{FourierDiffusion}
    \end{subfigure}
    \hfill
    \begin{subfigure}[b]{0.3\textwidth}
        \includegraphics[width=\textwidth]{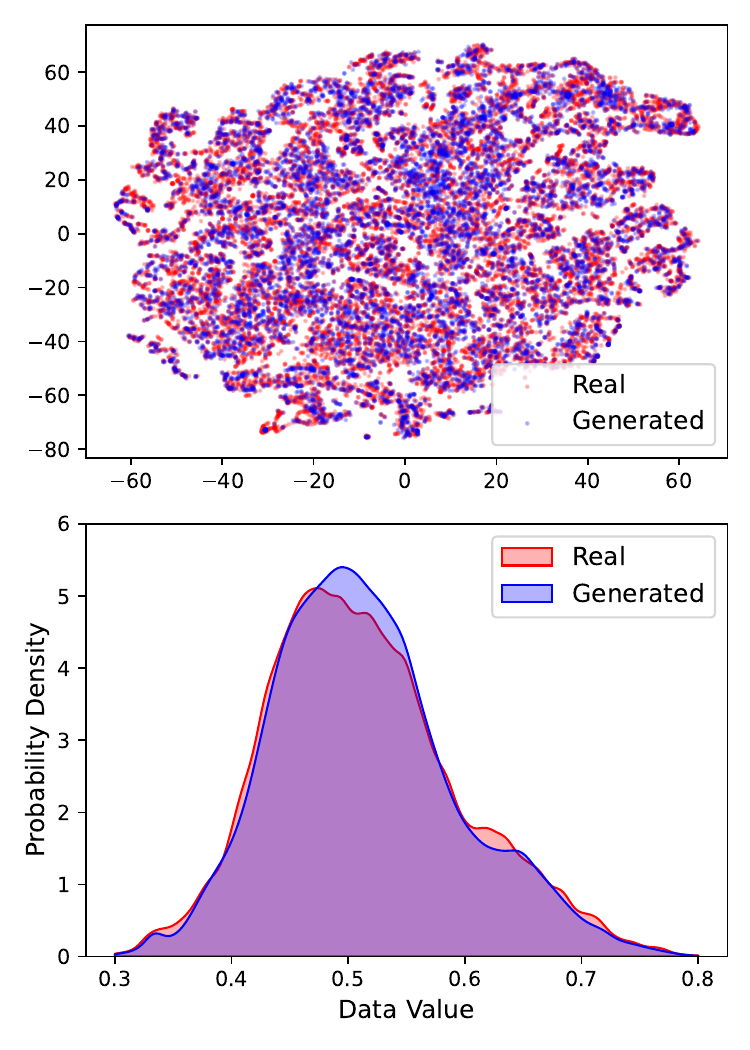}
        \caption{Diffusion-TS}
    \end{subfigure}
    }
    \caption{t-SNE visualization and probability distributions of training/synthetic data for ETTh1.}
    \label{fig:etth1_tsne_pdf}
\end{figure}

\begin{figure}[htbp]
    \centering
    \resizebox{0.8\textwidth}{!}{%
    \begin{subfigure}[b]{0.3\textwidth}
        \includegraphics[width=\textwidth]{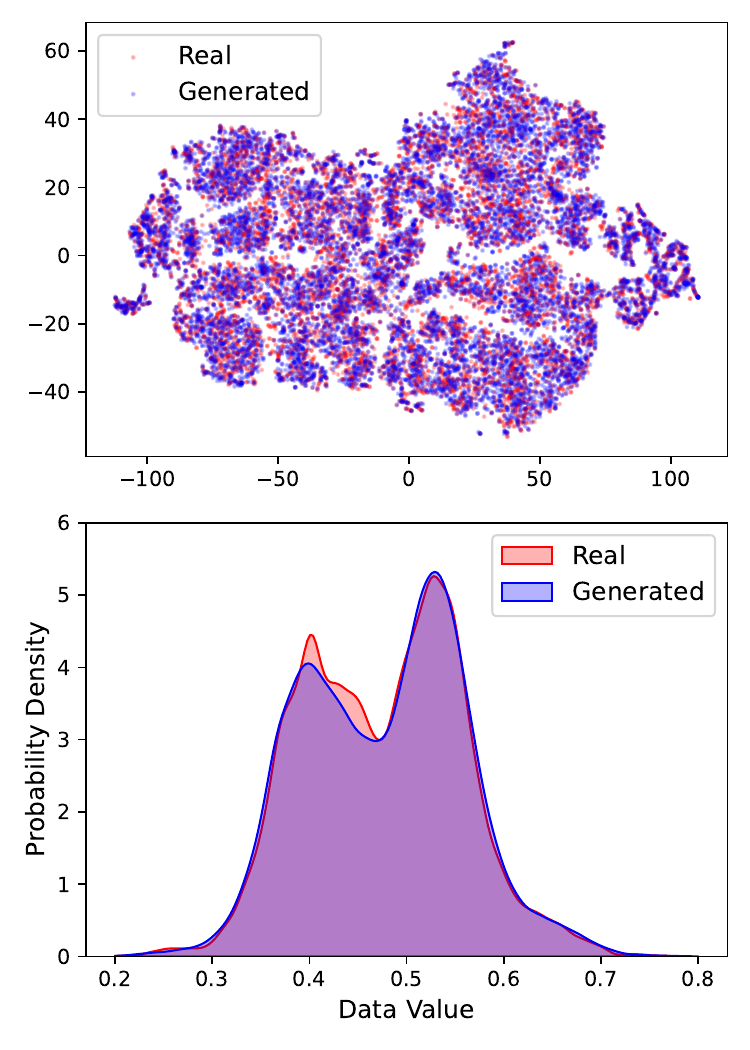}
        \caption{WaveletDiff}
    \end{subfigure}
    \hfill
    \begin{subfigure}[b]{0.3\textwidth}
        \includegraphics[width=\textwidth]{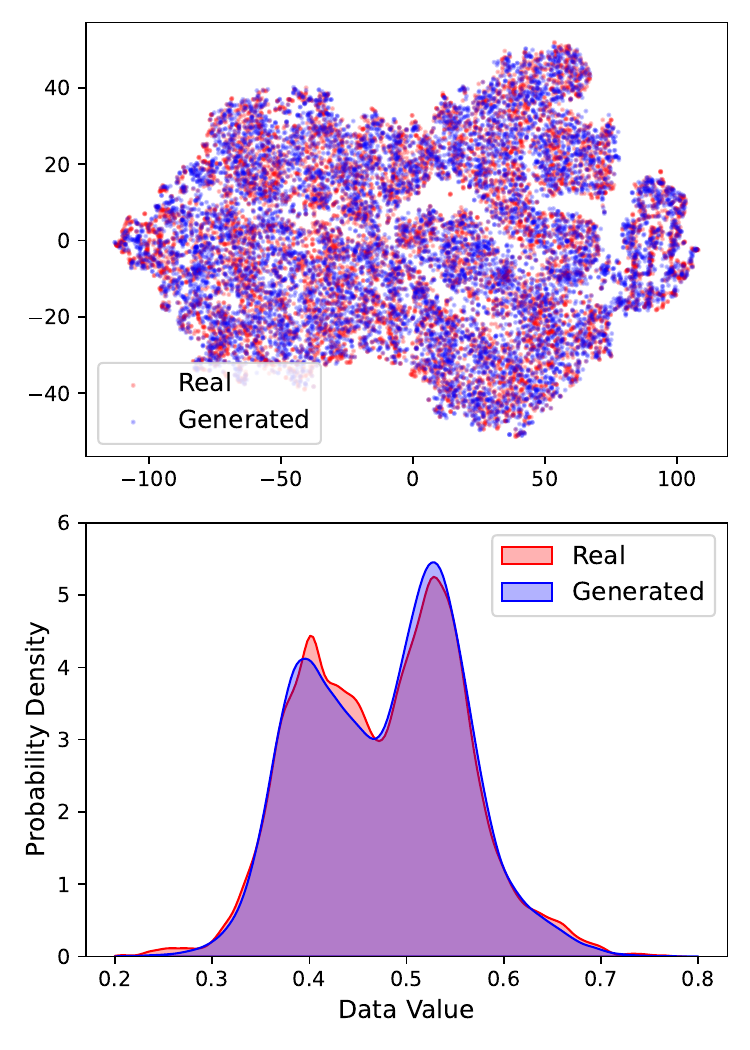}
        \caption{FourierDiffusion}
    \end{subfigure}
    \hfill
    \begin{subfigure}[b]{0.3\textwidth}
        \includegraphics[width=\textwidth]{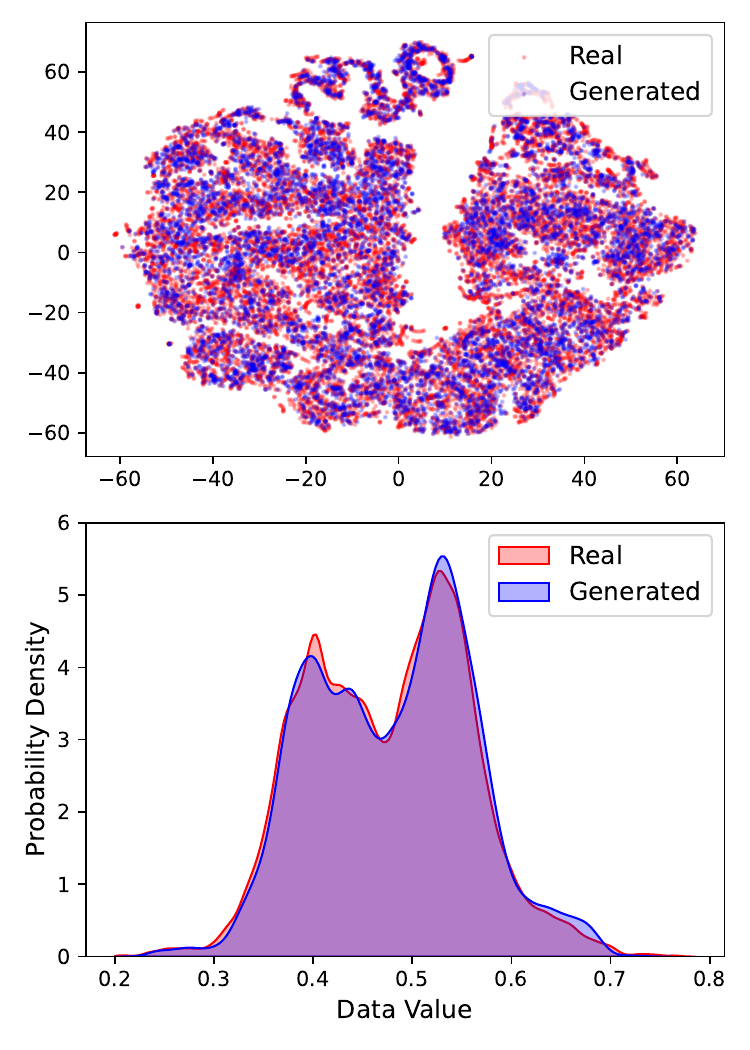}
        \caption{Diffusion-TS}
    \end{subfigure}
    }
    \caption{t-SNE visualization and probability distribution of data values on ETTh2 dataset.}
    \label{fig:etth2_tsne_pdf}
\end{figure}

\begin{figure}[htbp]
    \centering
    \resizebox{0.8\textwidth}{!}{
    \begin{subfigure}[b]{0.3\textwidth}
        \includegraphics[width=\textwidth]{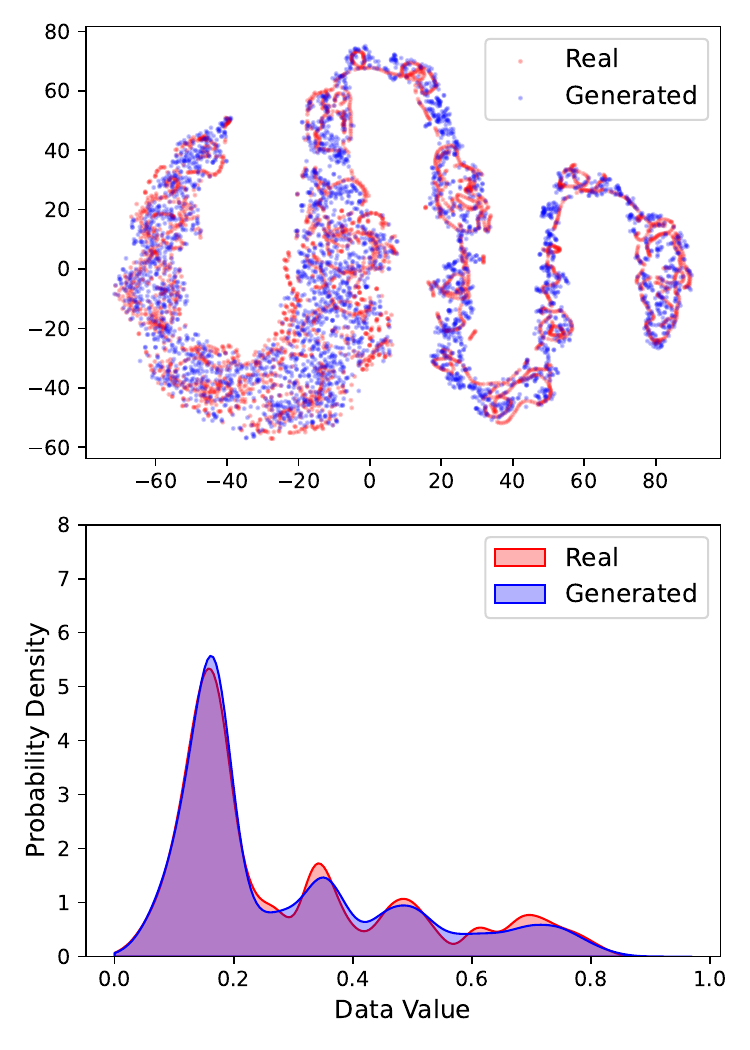}
        \caption{WaveletDiff}
    \end{subfigure}
    \hfill
    \begin{subfigure}[b]{0.3\textwidth}
        \includegraphics[width=\textwidth]{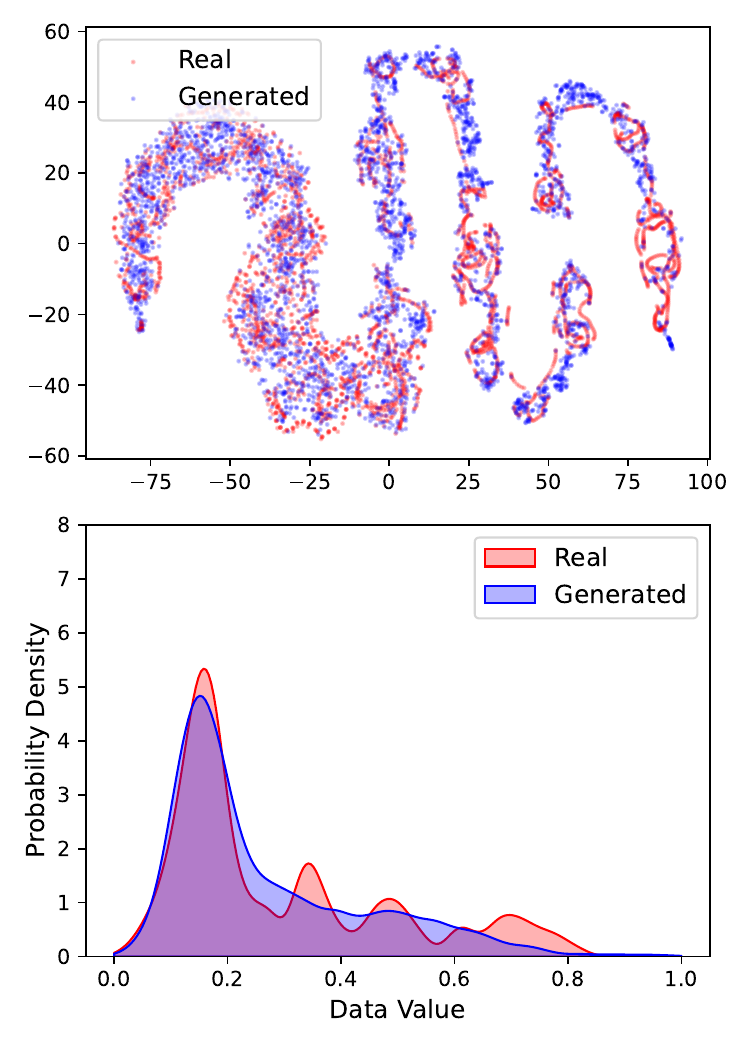}
        \caption{FourierDiffusion}
    \end{subfigure}
    \hfill
    \begin{subfigure}[b]{0.3\textwidth}
        \includegraphics[width=\textwidth]{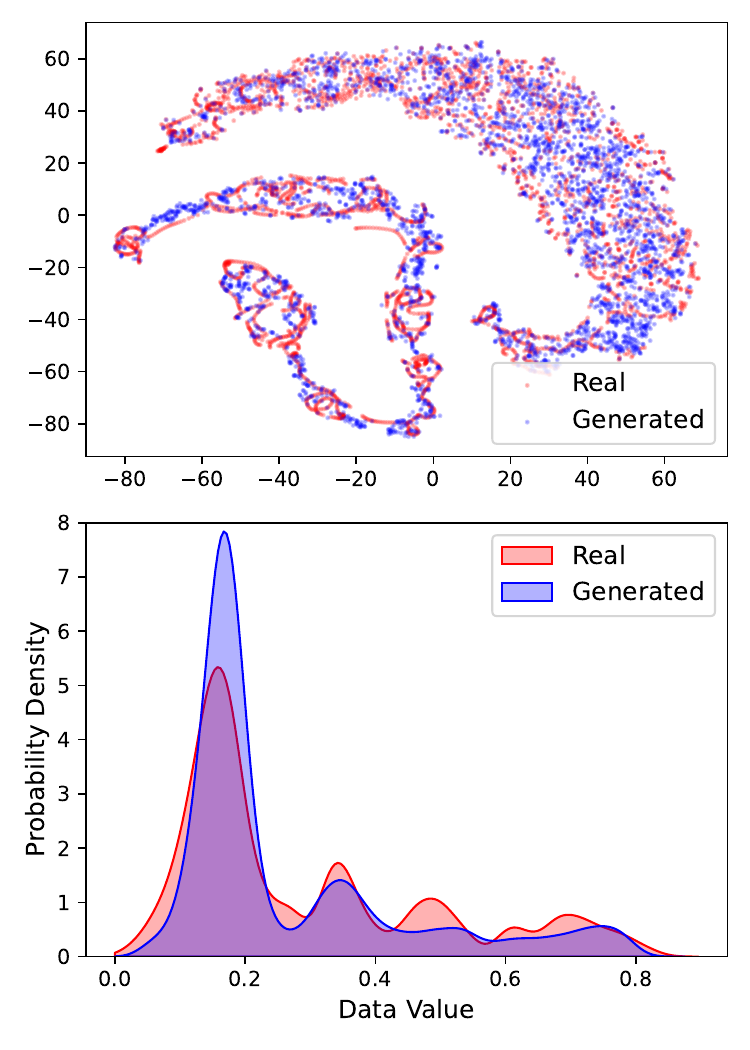}
        \caption{Diffusion-TS}
    \end{subfigure}
    }
    \caption{t-SNE visualization and probability distributions of training/synthetic data for Stocks.}
    \label{fig:stocks_tsne_pdf}
\end{figure}

\begin{figure}[htbp]
    \centering
    \resizebox{0.8\textwidth}{!}{%
    \begin{subfigure}[b]{0.3\textwidth}
        \includegraphics[width=\textwidth]{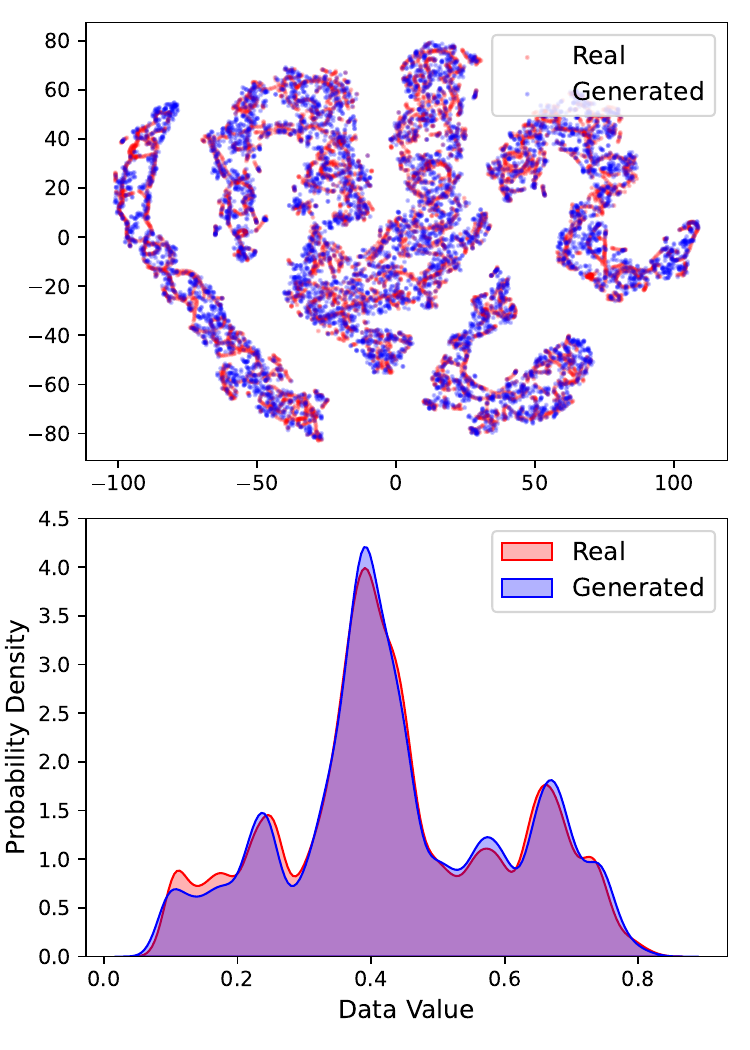}
        \caption{WaveletDiff}
    \end{subfigure}
    \hfill
    \begin{subfigure}[b]{0.3\textwidth}
        \includegraphics[width=\textwidth]{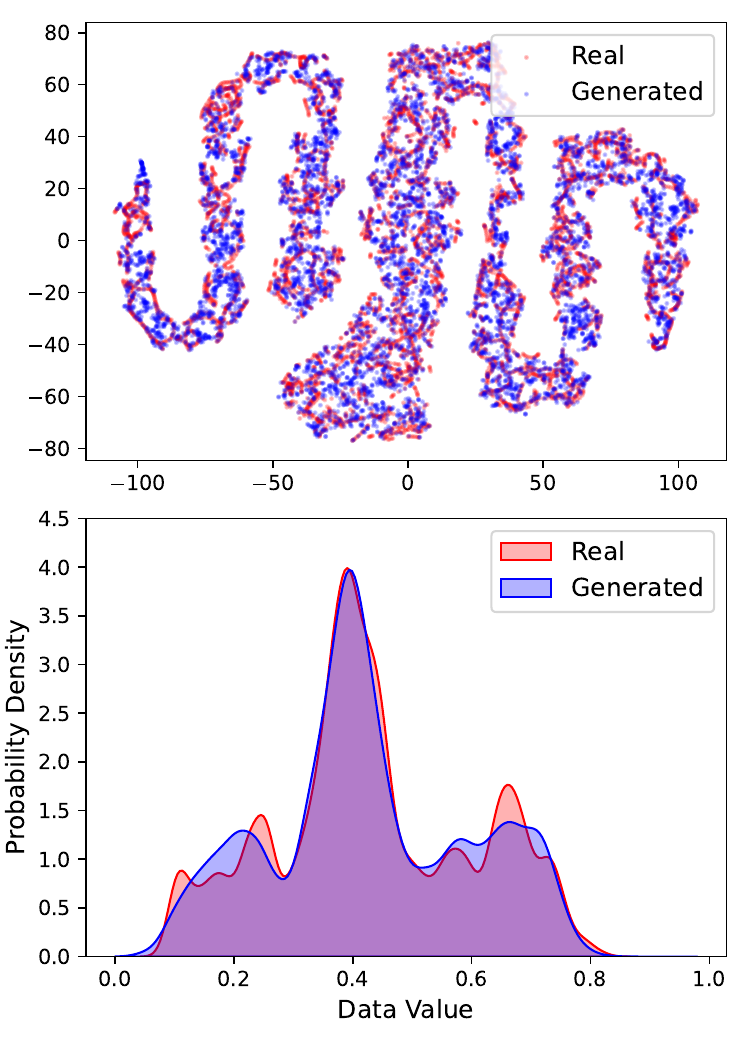}
        \caption{FourierDiffusion}
    \end{subfigure}
    \hfill
    \begin{subfigure}[b]{0.3\textwidth}
        \includegraphics[width=\textwidth]{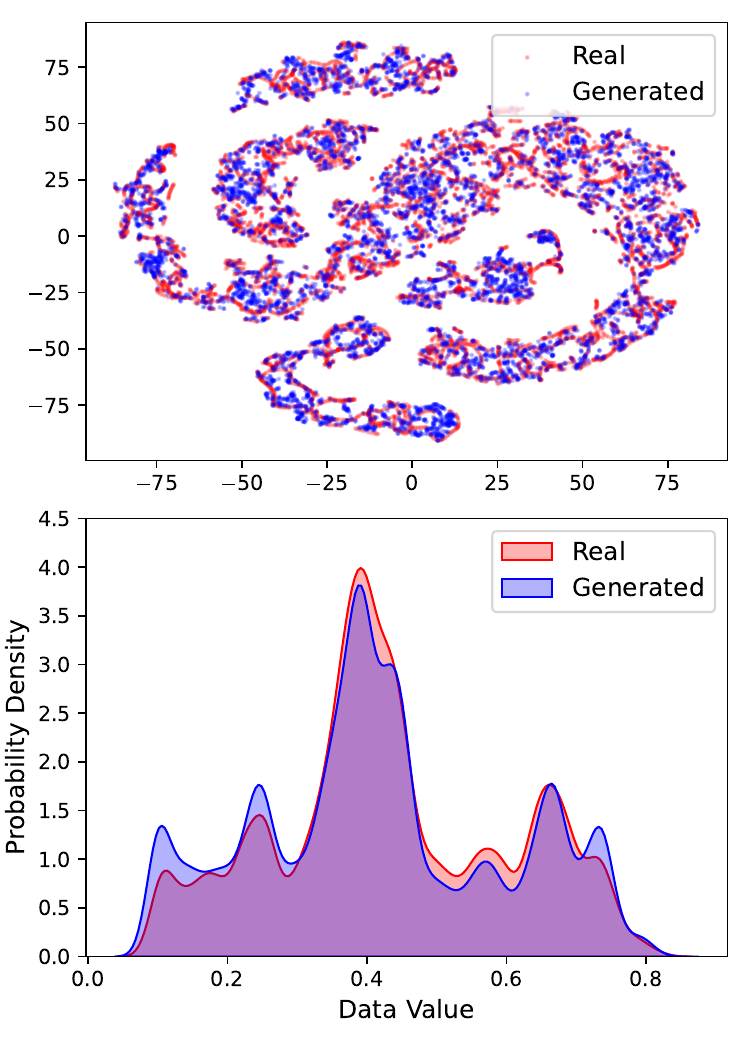}
        \caption{Diffusion-TS}
    \end{subfigure}
    }
    \caption{t-SNE visualization and probability distribution of data values on Exchange Rate dataset.}
    \label{fig:exchange_rate_tsne_pdf}
\end{figure}

\begin{figure}[htbp]
    \centering
    \resizebox{0.8\textwidth}{!}{%
    \begin{subfigure}[b]{0.3\textwidth}
        \includegraphics[width=\textwidth]{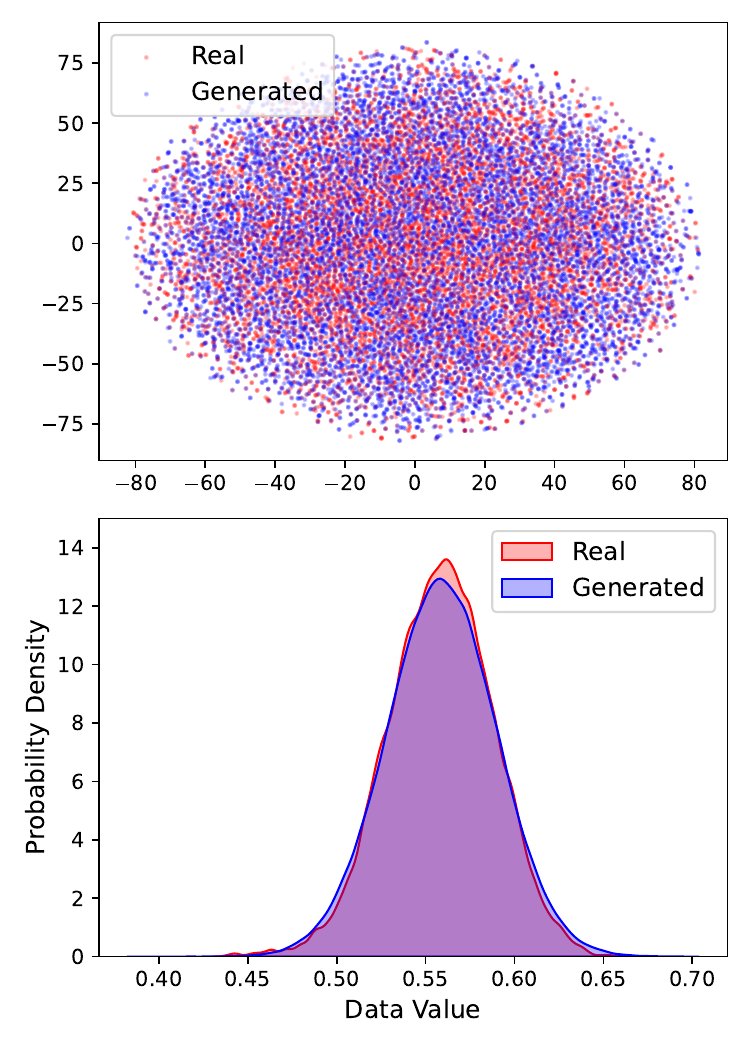}
        \caption{WaveletDiff}
    \end{subfigure}
    \hfill
    \begin{subfigure}[b]{0.3\textwidth}
        \includegraphics[width=\textwidth]{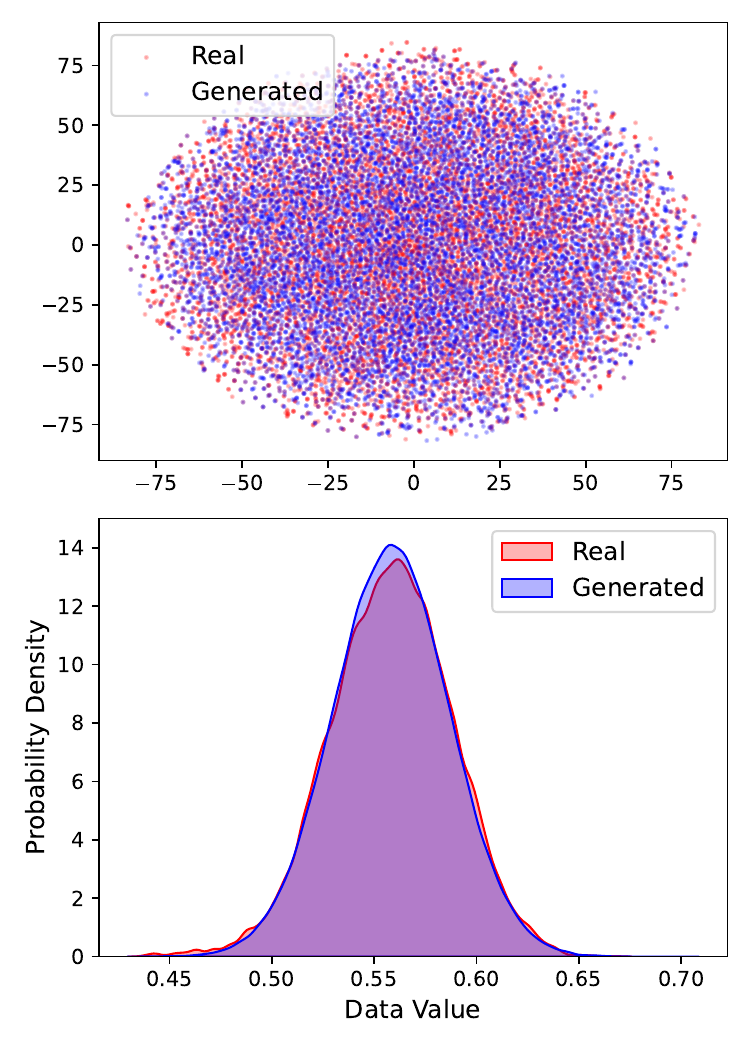}
        \caption{FourierDiffusion}
    \end{subfigure}
    \hfill
    \begin{subfigure}[b]{0.3\textwidth}
        \includegraphics[width=\textwidth]{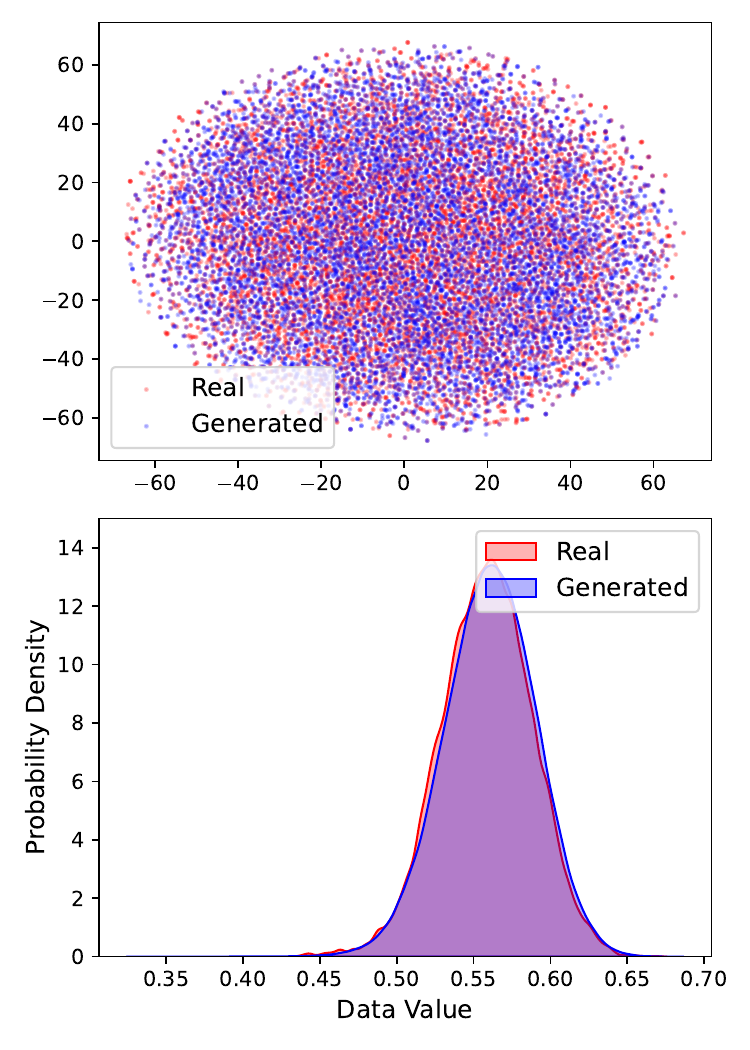}
        \caption{Diffusion-TS}
    \end{subfigure}
    }
    \caption{t-SNE visualization and probability distribution of data values on fMRI dataset.}
    \label{fig:fmri_tsne_pdf}
\end{figure}

\begin{figure}[htbp]
    \centering
    \resizebox{0.8\textwidth}{!}{%
    \begin{subfigure}[b]{0.3\textwidth}
        \includegraphics[width=\textwidth]{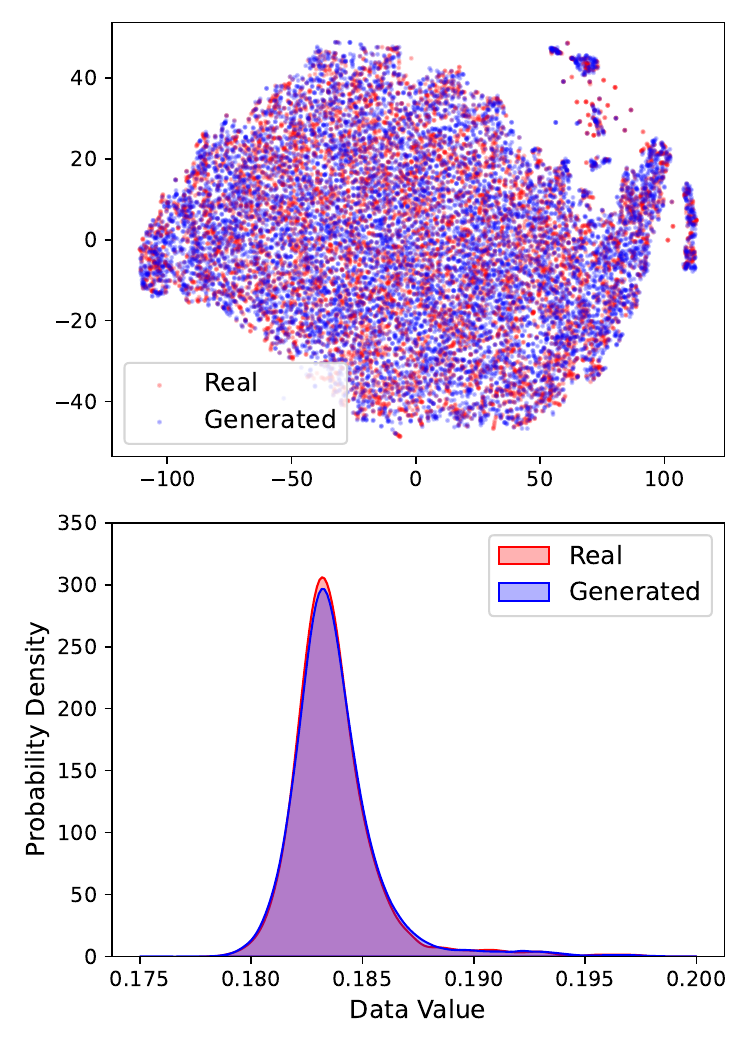}
        \caption{WaveletDiff}
    \end{subfigure}
    \hfill
    \begin{subfigure}[b]{0.3\textwidth}
        \includegraphics[width=\textwidth]{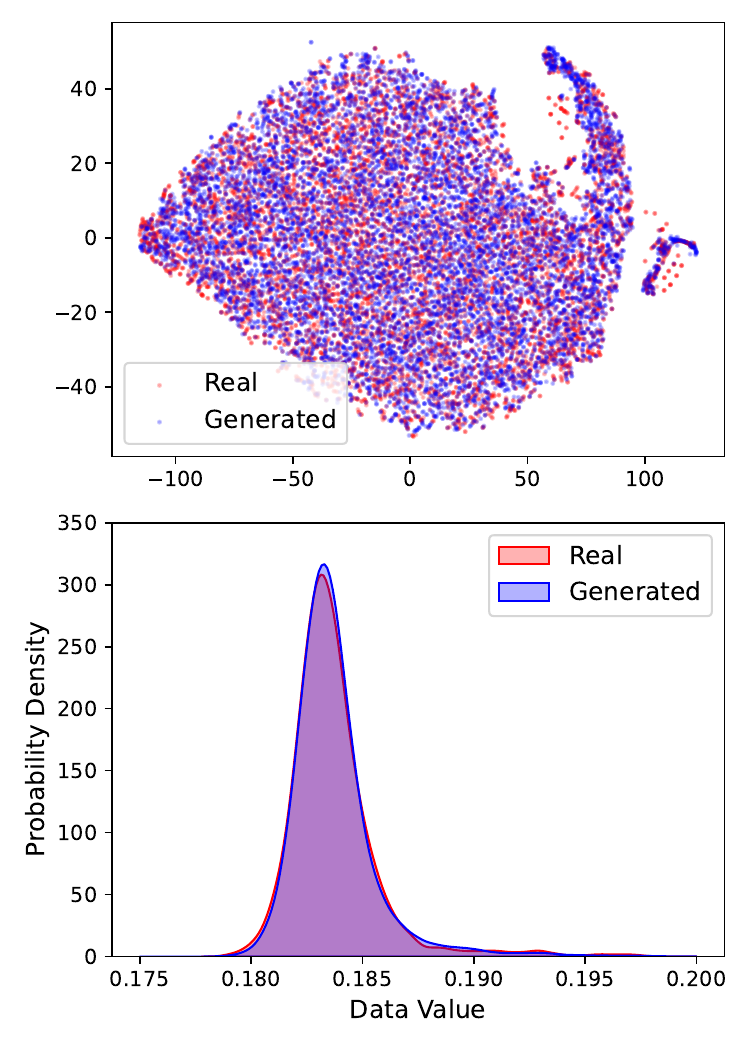}
        \caption{FourierDiffusion}
    \end{subfigure}
    \hfill
    \begin{subfigure}[b]{0.3\textwidth}
        \includegraphics[width=\textwidth]{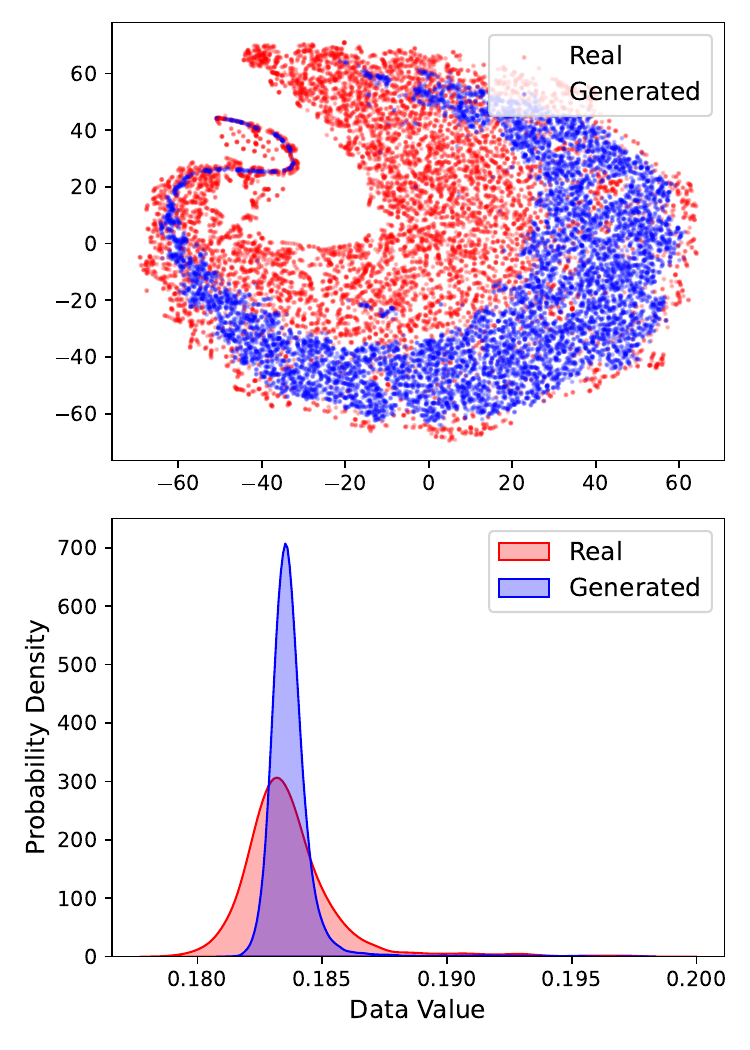}
        \caption{Diffusion-TS}
    \end{subfigure}
    }
    \caption{t-SNE visualization and probability distribution of data values on EEG dataset.}
    \label{fig:eeg_tsne_pdf}
\end{figure}

We present the t-SNE and data distribution visualization of short sequence generation on ETTh1, ETTh2, Stocks, Exchange Rate, fMRI, and EEG dataset in Figure~\ref{fig:etth1_tsne_pdf},~\ref{fig:etth2_tsne_pdf},~\ref{fig:stocks_tsne_pdf},~\ref{fig:exchange_rate_tsne_pdf},~\ref{fig:fmri_tsne_pdf}, and~\ref{fig:eeg_tsne_pdf}, respectively. The results demonstrate that WaveletDiff consistently outperforms diffusion baseline methods in capturing the underlying data distributions across all datasets.

\section{Diffusion Model Reproducibility Analysis}
\label{appendix:reproducibility}
Inspired by recent work on reproducibility in diffusion models for images~\citep{pmlr-v235-zhang24cn, li2024understanding, kadkhodaie2024generalization}, we examine whether this phenomenon extends to time series generation. Specifically, we train pairs of models with slightly different architectures or configurations, then generate samples from the same fixed Gaussian noise input using deterministic DDIM sampling. For each pair of generated sequences $(x_1, x_2)$, we compute their similarity using dynamic time warping (DTW) distance. To quantify reproducibility, we follow~\citep{pmlr-v235-zhang24cn} and define the RP score as
\begin{equation}
    \text{RP score} \coloneq \mathbb{P}\!\left(\text{DTW}(x_1, x_2) < \overline{\text{DTW}}_{\text{rand}}\right),
\end{equation}
where $\overline{\text{DTW}}_{\text{rand}}$ denotes the average DTW distance between randomly chosen sequence pairs generated by the two models. Thus, the RP score measures the probability that two models produce more similar samples from the same noise than would be expected by chance. An RP score greater than $0.5$ indicates reproducibility.

Unlike image generation, which commonly uses U-Net architectures for comparison, time series generation employs diverse architectures. We examine the RP score across different model variations for both WaveletDiff and FourierDiffusion. Table~\ref{tab:rp_score} demonstrates that reproducibility exists for time series data regardless of the representation domain (wavelet, time, or Fourier). To the best of our knowledge, we are the first to examine the reproducibility phenomenon specifically for time series generation.

\begin{table}[htbp]
    \caption{Reproducibility scores for different model variations demonstrate that time series diffusion models exhibit reproducibility across architectural changes and representation domains.}
    \centering
    \begin{tabular}{c|c|c|c}
        \hline
        Datasets & Model 1 & Model 2 & RP score \\
        \hline
        \multirow{4}{*}{Stocks} 
            & \multirow{2}{*}{WaveletDiff} 
            & WaveletDiff w/o cross-attention 
                & 1.0\\
            & & WaveletDiff + cosine noise scheduler 
                & 0.66 \\
            \cline{2-4}
            & \multirow{2}{*}{FourierDiffusion} 
            & FourierDiffusion on time domain 
                & 0.665 \\
            & & FourierDiffusion using LSTM score model
                & 0.805 \\
        \hline
        \multirow{4}{*}{Exchange Rate} 
            & \multirow{2}{*}{WaveletDiff} 
            & WaveletDiff w/o cross-attention 
                & 0.999 \\
            & & WaveletDiff + cosine noise scheduler 
                & 0.619 \\
            \cline{2-4}
            & \multirow{2}{*}{FourierDiffusion} 
            & FourierDiffusion on time domain 
                & 0.610 \\
            & & FourierDiffusion using LSTM score model
                & 0.945 \\
        \hline
    \end{tabular}
    \label{tab:rp_score}
\end{table}

\subsection{Impact of Architectural Variations}
\begin{figure}[htbp]
    \centering
    \includegraphics[width=0.85\linewidth]{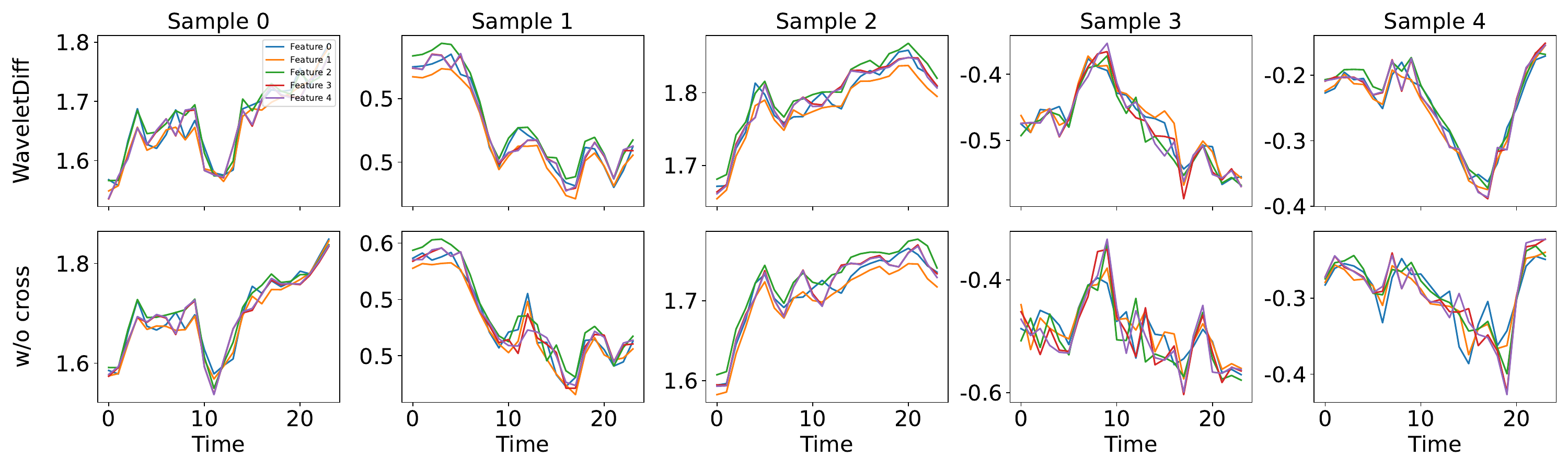}
    \caption{Reproducibility comparison on Stocks dataset using identical initial noise (volume feature excluded for clarity).}
    \label{fig:stocks_nocross_reproducibility}
\end{figure}

\begin{figure}[htbp]
    \centering
    \includegraphics[width=0.85\linewidth]{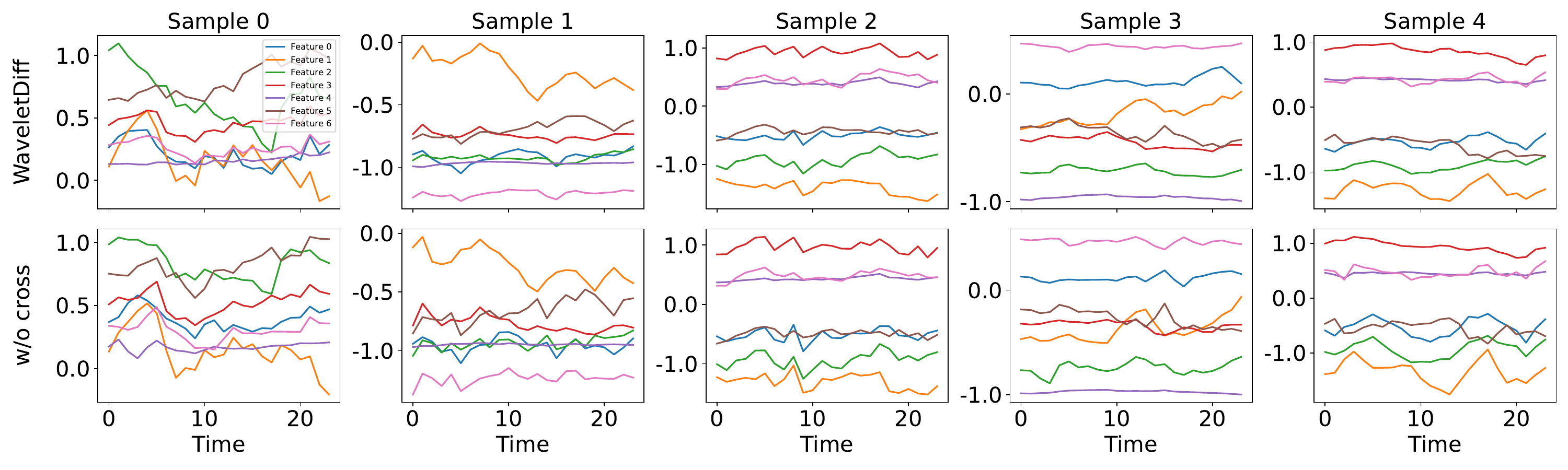}
    \caption{Reproducibility comparison on Exchange Rate dataset using identical initial noise.}
    \label{fig:exchange_rate_nocross_reproducibility}
\end{figure}

We evaluate how architectural modifications affect reproducibility by removing the cross-attention module from WaveletDiff. Figures~\ref{fig:stocks_nocross_reproducibility} and~\ref{fig:exchange_rate_nocross_reproducibility} show that the model maintains strong reproducibility despite this significant architectural change, with generated sequences from identical noise exhibiting nearly identical patterns.

\subsection{Impact of Mother Wavelet Selection}
\begin{figure}[htbp]
    \centering
    \includegraphics[width=0.85\linewidth]{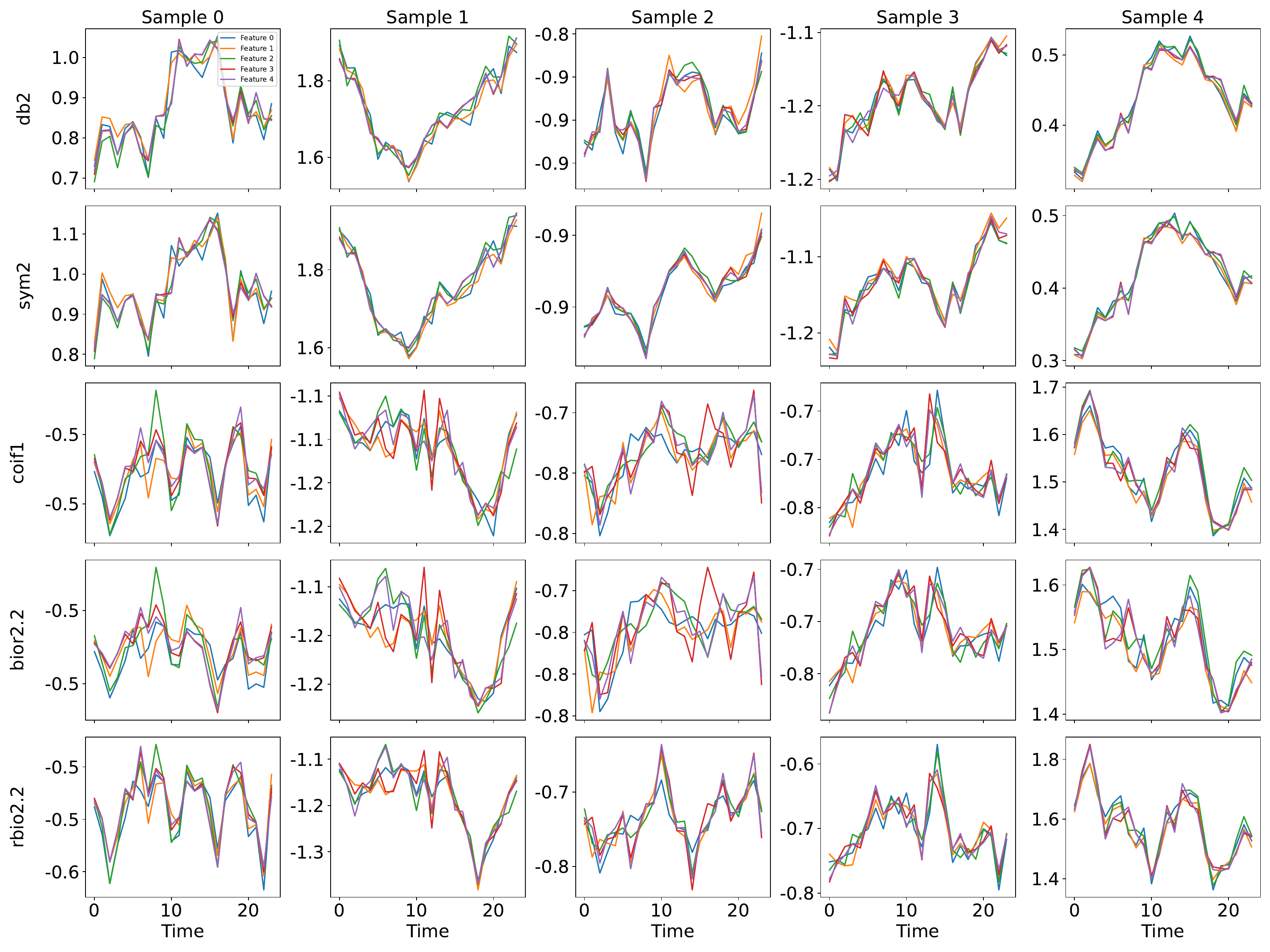}
    \caption{Wavelet family comparison on Stocks dataset demonstrating varying reproducibility across different mother wavelets.}
    \label{fig:stocks_wavelet_reproducibility}
\end{figure}

\begin{figure}[htbp]
    \centering
    \includegraphics[width=0.85\linewidth]{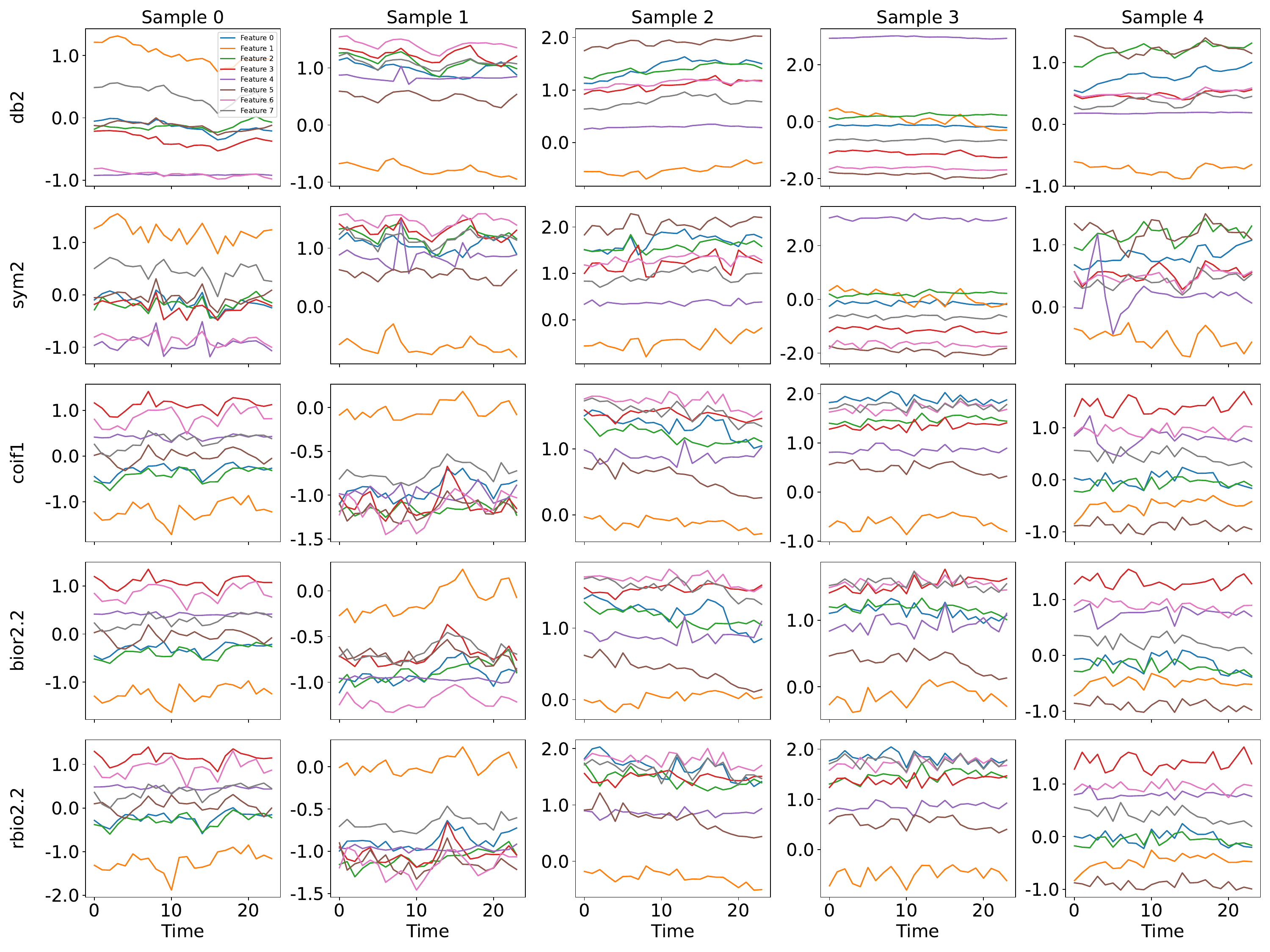}
    \caption{Wavelet family comparison on Exchange Rate dataset demonstrating varying reproducibility across mother wavelets.}
    \label{fig:exchange_rate_wavelet_reproducibility}
\end{figure}

We further investigate how different mother wavelet choices affect reproducibility. As shown in Figures~\ref{fig:stocks_wavelet_reproducibility} and~\ref{fig:exchange_rate_wavelet_reproducibility}, Daubechies and Symlets wavelets demonstrate high reproducibility, while Coiflets, Biorthogonal, and Reverse Biorthogonal wavelets also exhibit good consistency. This indicates that wavelet choice significantly affects the learned distribution, with similar wavelet families (e.g., orthogonal wavelets like Daubechies and Symlets) producing more comparable results than dissimilar families.

\end{document}